\documentclass[lettersize,journal]{IEEEtran}
\usepackage{array}
\usepackage{textcomp}
\usepackage{stfloats}
\usepackage{url}
\usepackage{verbatim}
\def\BibTeX{{\rm B\kern-.05em{\sc i\kern-.025em b}\kern-.08em
    T\kern-.1667em\lower.7ex\hbox{E}\kern-.125emX}}
\usepackage{balance}

\usepackage[english]{babel}
\usepackage{booktabs}

\usepackage{ifpdf}

\usepackage{cite} 
\usepackage{hyperref}
\usepackage[framemethod=default]{mdframed}

\ifCLASSINFOpdf
\usepackage[pdftex]{graphicx}
\graphicspath{{./figures/}}
\else
\usepackage[dvips]{graphicx}
\graphicspath{./figures/}
\fi
\usepackage{color}

\usepackage{pgf, tikz, pgfplots}
\usetikzlibrary{shapes, arrows, automata}
\usetikzlibrary{calc,hobby,decorations}

\usetikzlibrary{shapes.geometric}
\usetikzlibrary{fit}
\usetikzlibrary{decorations.pathmorphing, decorations.pathreplacing,angles,quotes}
\usetikzlibrary{positioning, arrows.meta, bending}

\usepackage[cmex10]{amsmath}
\usepackage{amsfonts, amssymb, amsthm}
\usepackage{mathrsfs}
\usepackage{tcolorbox}

\usepackage{algorithm} 
\usepackage{algorithmic}
\usepackage{enumerate}
\usepackage{multirow}
\usepackage{rotating}
\usepackage{subcaption}
\usepackage{needspace}

\input{mySymbol.sty}

\definecolor{penndarkestblue}{cmyk}{1,0.74,0,0.77}
\definecolor{penndarkerblue}{cmyk}{1,0.74,0,0.70}
\definecolor{pennblue}{cmyk}{0.99,0.66,0,0.57} 
\definecolor{pennlighterblue}{cmyk}{0.98,0.44,0,0.35}
\definecolor{pennlightestblue}{cmyk}{0.38,0.17,0,0.17} 

\definecolor{penndarkestred}{cmyk}{0,1,0.89,0.66}
\definecolor{penndarkerred}{cmyk}{0,1,0.88,0.55}
\definecolor{pennred}{cmyk}{0,1,0.83,0.42} 
\definecolor{pennlighterred}{cmyk}{0,1,0.6,0.24}
\definecolor{pennlightestred}{cmyk}{0,0.43,0.26,0.12} 

\definecolor{penndarkestgreen}{cmyk}{1,0,1,0.68}
\definecolor{penndarkergreen}{cmyk}{1,0,1,0.57}
\definecolor{penngreen}{cmyk}{1,0,1,0.44} 
\definecolor{pennlightergreen}{cmyk}{1,0,1,0.25}
\definecolor{pennlightestgreen}{cmyk}{0.43,0,0.43,0.13}

\definecolor{penndarkestorange}{cmyk}{0,0.65,1,0.49}
\definecolor{penndarkerorange}{cmyk}{0,0.65,1,0.33}
\definecolor{pennorange}{cmyk}{0,0.54,1,0.24} 
\definecolor{pennlighterorange}{cmyk}{0,0.32,1,0.13}
\definecolor{pennlightestorange}{cmyk}{0,0.15,0.46,0.06}

\definecolor{penndarkestpurple}{cmyk}{0,1,0.11,0.86}
\definecolor{penndarkerpurple}{cmyk}{0,1,0.13,0.82}
\definecolor{pennpurple}{cmyk}{0,1,0.11,0.71} 
\definecolor{pennlighterpurple}{cmyk}{0,1,0.05,0.46}
\definecolor{pennlightestpurple}{cmyk}{0,0.35,0.02,0.23}

\definecolor{pennyellow}{cmyk}{0,0.20,1,0.05} 
\definecolor{pennlightgray1}{cmyk}{0,0,0,0.05}
\definecolor{pennlightgray3}{cmyk}{0.01,0.01,0,0.18}
\definecolor{pennmediumgray1}{cmyk}{0.04,0.03,0,0.31}
\definecolor{pennmediumgray4}{cmyk}{0.08,0.06,0,0.54}
\definecolor{penndarkgray2}{cmyk}{0.09,0.07,0,0.71}
\definecolor{penndarkgray4}{cmyk}{0.1,0.1,0,0.92}

\def\SO3{\mathrm{SO(3)}}

\newtheorem{assumption}{\hspace{0pt}\bf Assumption \hspace{-0.15cm}}

\newtheorem{theorem}{\hspace{0pt}\bf Theorem}

\newtheorem{remark}{\hspace{0pt}\bf Remark}

\newtheorem{definition}{\hspace{0pt}\bf Definition}

\begin{document}
\title{Co-Optimizing Reconfigurable Environments and Policies for Decentralized Multi-Agent Navigation}

\author{Zhan Gao, ~\IEEEmembership{IEEE Student~Member},  Guang Yang, ~\IEEEmembership{IEEE Member} and Amanda Prorok, ~\IEEEmembership{IEEE Senior Member}
\thanks{The authors are with the Department of Computer Science and Technology, University of Cambridge, UK (Email: $\{$zg292, asp45$\}$@cam.ac.uk, gyang101@bu.edu). This work is supported by ERC Project 949940 (gAIa).}}

\markboth{}%
{Co-Optimizing Reconfigurable Environments and Policies for Decentralized Multi-Agent Navigation}

\maketitle

\begin{abstract}
This work views the multi-agent system and its surrounding environment as a co-evolving system, where the behavior of one affects the other. The goal is to take both agent actions and environment configurations as decision variables, and optimize these two components in a coordinated manner to improve some measure of interest. Towards this end, we consider the problem of decentralized multi-agent navigation in a cluttered environment, where we assume that the layout of the environment is reconfigurable. By introducing two sub-objectives---multi-agent navigation and environment optimization---we propose an \textit{agent-environment co-optimization} problem and develop a \textit{coordinated algorithm} that alternates between these sub-objectives to search for an optimal synthesis of agent actions and environment configurations; ultimately, improving the navigation performance. Due to the challenge of explicitly modeling the relation between the agents, the environment and their performance therein, we leverage policy gradient to formulate a model-free learning mechanism within the coordinated framework. A formal convergence analysis shows that our coordinated algorithm tracks the local minimum solution of an associated time-varying non-convex optimization problem. Experiments corroborate theoretical findings and show the benefits of co-optimization. Interestingly, the results also indicate that optimized environments can offer structural guidance to de-conflict agents in motion.
\end{abstract}


\section{Introduction}

Multi-agent systems comprise multiple decision-making agents that interact in a shared environment to achieve common or individual goals. They present an attractive solution to spatially distributed tasks, wherein efficient and safe motion planning is one of the central problems. Classically, the field of multi-agent motion planning does not consider the surrounding environment as a decision variable, instead acquiring environmental information through perceptual techniques and modeling it as an unmodifiable constraint~\cite{van2008reciprocal, desaraju2012decentralized1, standley2011complete, wu2020multi, everett2018motion, gao2023online}. However, such spatial constraints may result in dead-locks, live-locks and prioritization conflicts even for state-of-the-art algorithms \cite{mani2010search, ruderman_Uncovering_2018}, highlighting the environment impact on agents' performance. Yet, despite the entanglement of these two components, coordinated optimization of multi-agent policies and environment configurations has received little attention thus far. 

With advances in programmable and responsive buildings, reconfigurable and automated environments are emerging as a new trend~\cite{wang2010new, bier2014robotic, bellusci2020multi, custodio2020flexible}. For example, the locations of shelves in warehouses can be adjusted with rack pulleys or sliding track systems; the layout of manufacturing equipment in factories can be re-designed or re-arranged within pre-designated regions; and the positions of dining tables in restaurants can be adapted to better suit food delivery paths. In parallel to developing agent policies, this means that we have the opportunity to re-configure the spatial layout of the environment as a means to improve some objective performance measure. The latter is especially appealing when agents are expected to solve repetitive tasks in structured settings (like warehouses, factories, restaurants, etc.). While it is possible to hand-design environment configurations, such a procedure is inefficient and often sub-optimal \cite{cap_Prioritized_2015}.

Recent works have suggested that significant benefits can be reaped by either adapting the environment to agent policies or tuning agent policies based on the environment \cite{zhang2009general, keren2015goal, vcap2015complete, kulkarni2020designing, gao2022environment, gao2023constrained}. These insights motivate the coupling of the environment configuration design with the agents' policy development. The goal of this work is to consider the obstacle layout of the environment as a decision variable, as well as the navigation policy of agents, and propose a system-level co-optimization problem that simultaneously optimizes two components; ultimately, \textit{establishing a symbiotic agent-environment system} that maximizes the navigation performance of the multi-agent system. Applications include warehouse logistics (e.g., finding the optimal shelf positions and robot policies for cargo transportation \cite{wurman2008coordinating}), search and rescue in collapsed sites (e.g., clearing passageways and deploying rescue strategies 
for trapped victims \cite{karma2015use}), city planning (e.g., designing the optimal 
one-way routes and autonomous vehicle strategies for traffic efficiency~\cite{drezner1997selecting}), and digital entertainment (e.g., building gaming scenes that elicit the best possible behaviors in animated video game characters\cite{johnson2001computer}). 

\begin{figure}[!t]
\centering
    \resizebox{0.435 \textwidth}{!}{\input{Overview}}

\caption{Agent-environment co-optimization is an optimization process that alternates between a multi-agent navigation policy (parameterized by $\theta_a$) and an environment generative model (parameterized by $\theta_o$), searching for an optimal pair $(\theta_a^*, \theta_o^*)$ maximizing the system performance. In the above example, it optimizes the navigation policy with reinforcement learning (e.g., blue trajectories become yellow ones) and the environment configuration with unsupervised learning (e.g., grey obstacles become green ones), in an alternating manner.}
\label{fig:Framework}\vspace{-6mm}
\end{figure}

In pursuit of this goal, we propose a \textit{coordinated optimization} algorithm that alternates between the synthesis of navigation behavior and environment design. Due to the challenge of explicitly modeling the relation between agents, environment and navigation performance, we pose the problem in a model-free learning framework and introduce two sub-objectives of multi-agent navigation and environment optimization, respectively. We parameterize the navigation policy of agents with a graph neural network (GNN) for decentralized implementation and use a generative model of the environment, parameterized by a deep neural network (DNN). The two models are updated with respect to two sub-objectives in an alternating manner, pursuing an optimal pair of navigation and generative parameters that maximizes the multi-agent system performance -- see Fig. \ref{fig:Framework} for an overview of the agent-environment coordinated optimization framework. In more detail, our contributions are: 

\begin{itemize}
	\item We propose the problem of agent-environment co-optimization, which considers both environment configurations and multi-agent policies as decision variables in a system-level optimization framework. The goal is to search for an optimal pair that jointly maximizes the 
    navigation performance. 
	\item We develop a coordinated method that optimizes the navigation policy of the agents with reinforcement learning and a generative model of the environment configuration with unsupervised learning, based on two sub-objectives, in an alternating manner. We formulate a model-free learning mechanism within the coordinated framework, which overcomes the challenge of modeling the explicit relation between agents, environment and performance.    
	\item We provide a formal convergence analysis for the proposed coordinated method by detailing its relation with an associated non-convex time-varying optimization problem and an ordinary differential equation. 
	\item We perform simulations as well as real-world experiments to evaluate the proposed method. The results corroborate theoretical findings and provide novel insights regarding the relationship between the environment and the multi-agent system. That is, we show how appropriately designed environments are able to guide multi-agent navigation through targeted spatial de-confliction.
\end{itemize}

\smallskip
\noindent \textbf{Related work.} The idea of co-optimization (or co-design) in robotics can be traced back to work on \textit{embodied cognitive science}~\cite{clark1998being}, which emphasizes the importance of the robot body and its interaction with the environment for achieving robust behavior. Motivated by this idea, research in the domain of robot co-design aims to find an optimized controller, sensing and morphology, such that they together produce the desired performance~\cite{tanaka2015sdp,tatikonda2004control,tzoumas2018sensing}. A few seminal works tackled this problem with an evolutionary approach that co-optimizes objectives~\cite{lipson2000automatic,cheney2018scalable1}. More recently, \cite{schaff_Jointly_2019} employed reinforcement learning to learn both the policy and morphology design parameters in an alternating manner, where a single shared policy controls the distribution of different robot designs. Within the multi-agent domain, the concept of co-optimization has mainly been leveraged to allocate physical resources to agent tasks. The work in~\cite{gabel2008joint} proposed a joint equilibrium policy search method that encourages agents to cooperate to reach states maximizing a global reward, while \cite{zhang2016co} developed a multi-level search approach that co-optimizes agent placements with task assignment and scheduling. In a similar vein, \cite{jain2017cooperative} used online multi-agent reinforcement learning to co-optimize cores, caches and on-chip networks. The works in \cite{ali2018motion, jaleel2016decentralized} considered the problem of co-optimizing mobility and communication among agents, and designed schedule policies that minimize the total energy. While broadly similar in their ideas, none of these works consider co-optimizing the \textit{environment} and \textit{navigation behaviors} to improve the system performance of multiple agents (i.e., robots). 

The key challenge of co-optimization stands in the fact that the explicit relation between agents, environment and system performance is generally not known (and hard to model), compounding the difficulty of solving the joint problem. The agent-environment relationship has been previously explored. The works in \cite{bennewitz2002finding, jager2001decentralized} identified the existence of congestion and deadlocks in undesirable environments and developed trajectory planning methods for agents to escape from these potential deadlocks, while \cite{vcap2015complete} introduced the concept of ``well-formed'' environment where the navigation tasks of all agents can be carried out successfully without collision. Wu et al. \cite{wu2020multi} showed that the environment shape can result in distinct path prospects for different agents and proposed to leverage this information for coordinating agents' motion. Additionally, \cite{gur2021adversarial} focused on adversarial environments and developed resilient navigation algorithms for agents in these environments. 
However, none of the aforementioned works consider \textit{both} the environment as well as the agents' policies as decision variables in a system-level optimization problem (with the aim of improving navigation performance). 

More in the vein of our work, Hauser \cite{hauser2013minimum, hauser2014minimum} attempted to remove obstacles from the environment to improve the agent's navigation performance, while focusing on a single agent scenario. The work in \cite{bellusci2020multi} extended a related idea to the multi-agent setting, to search over all possible environment configurations to find the best solution for agents. However, the technique is limited to discrete settings, and only considers obstacle removal (in contrast to general re-configuration). The problem of multi-agent pick-up and delivery also explores the relationship between agents and environment \cite{liu2019task, yamauchi2021path, yamauchi2022standby}. However, it considers environment changes as pre-defined tasks and assigns agents to complete these tasks -- constituting a different problem that is defined only in a discrete setting. 

\section{Problem Formulation}\label{sec:Problem}

Let $\ccalE$ be a $2$-D environment with $m$ obstacles $\ccalO \!=\! \{O_j\}_{j=1}^m$. The obstacles are dispersed within the obstacle region $\Delta$ according to specific environment configurations. Consider a multi-agent system with $n$ agents $\ccalA \!=\! \{A_i\}_{i=1}^n$ distributed in $\ccalE$. The agents are initialized randomly in the starting region $\ccalS$ and deploy a navigation policy $\pi_a$ to move towards the goal region $\ccalD$. We are interested in \textit{decentralized} multi-agent navigation, where agents have access to their own states and exchange local observations with neighbors in the communication range. Our performance metrics include \textit{traveled distance}, \textit{average speed} and \textit{collision avoidance}, which depend not only on the agents' policy $\pi_a$ but also the surrounding environment $\ccalE$. For example, we may facilitate the navigation by either developing an efficient policy, or designing a ``well-formed'' environment with an appropriate obstacle layout, or both. 

This motivates a view of agents $\ccalA$ and their environment $\ccalE$ as a co-evolving system. The problem of \emph{agent-environment co-optimization} then considers both components as decision variables for two objectives: \textit{(i)} finding the optimal policy $\pi^*_a$ for agents to reach goal positions without collision \textit{(ii)} finding the optimal obstacle layout for the environment $\ccalE^*$. The goal is to achieve these objectives simultaneously, i.e., \textit{improving the navigation policy while, towards the same end, optimizing the obstacle layout}. This explores the coupling between the agents' policy and the environment configuration, to find an optimal pair $(\pi_a^*, \ccalE^*)$ that maximizes the performance. In the following, we first introduce individual problem components and then formulate the co-optimization problem.

\subsection{Multi-Agent Navigation}\label{subsec:navigation}

We consider the agents $\ccalA$ initialized at starting positions $\bbS = [\bbs_1,...,\bbs_n] \in \ccalS$ and tasked towards goal positions $\bbD = [\bbd_1,...,\bbd_n] \in \ccalD$. Let $\bbX_a = [\bbx_{a1},...,\bbx_{an}]$ be the agent states and $\bbU_a = [\bbu_{a1},...,\bbu_{an}]$ the agent actions. For example, $\bbX_a$ are the positions or velocities and $\bbU_a$ are the accelerations. Let $\pi_a(\bbU_a | \bbX_a)$ be the navigation policy that generates actions $\bbU_a$ to steer the agents from $\bbS$ towards $\bbD$, which is a distribution of $\bbU_a$ conditioned on states $\bbX_a$. We model the navigation process as a \textit{sequential} decision-making problem, where agents take actions based on their instantaneous states (e.g., positions, velocities and goals) across time steps $t$ and do not require knowledge of the environment at run-time. 

In this work, we consider the decentralized navigation problem where states are local observations (available to individual agents) and the full state is not observable \cite{jimenez2018decentralized, sivanathan2020decentralized, tolstaya2020learning, ji2021decentralized, gao2023online}. The decentralized setting is of practical importance because: (i) There may not exist a central processing unit to compute actions of all agents or each agent may not be able to obtain global information; (ii) it may be challenging to maintain a centralized condition throughout time; and (iii) the centralized setting introduces a single point of failure. 

The navigation performance of the multi-agent system is determined by the navigation task, the moving capacity of the agents (e.g., maximal velocity) and the physical restrictions of the environment (e.g., obstacle blocking). In this context, we use an objective function $f(\bbS, \bbD, \pi_a, \ccalE)$ that depends on the initial positions $\bbS$, the goal positions $\bbD$, the navigation policy $\pi_a$, and the environment $\ccalE$. The goal of multi-agent navigation is to find the optimal $\pi_a^*$ that maximizes $f(\bbS,\bbD,\pi_a^*,\ccalE)$ given the navigation task $(\bbS, \bbD)$ and the environment $\ccalE$.

\subsection{Environment Optimization}\label{subsec:optimization}
The obstacle layout $\ccalO$ determines the environment configuration $\ccalE$ for multi-agent navigation. Let $\pi_o(\ccalO|\bbS,\bbD)$ be the generative model that configures the obstacle layout based on the navigation task, i.e., agents' initial and goal positions, which is a distribution of $\ccalO$ conditioned on $(\bbS, \bbD)$ of the agents $\ccalA$. The outputs of the generative model are re-configurable parameters of the obstacle layout, e.g., the obstacle position and the obstacle size, which are the degrees of freedom for environment optimization. 

Since the environment $\ccalE$ is determined by the obstacle layout and the latter is determined by the generative model $\pi_o$, we can re-write the objective function as $f(\bbS,\bbD,\pi_a,\pi_o)$. The goal of environment optimization is to find the optimal $\pi^*_o$ that generates the environment $\ccalE^*$ to maximize $f(\bbS,\bbD,\pi_a,\pi_o^*)$ given the navigation task $(\bbS, \bbD)$ and the agents' policy $\pi_a$. 

\subsection{Agent-Environment Co-Optimization}\label{subsec:AgentEnvironment}
We use definitions from individual sub-problems (Sec. \ref{subsec:navigation} and \ref{subsec:optimization}) to formulate the problem of agent-environment co-optimization, which searches for an optimal pair of the navigation policy and the generative model $(\pi_a^*, \pi_o^*)$ that maximizes the navigation performance of the multi-agent system, i.e., the objective $f(\bbS, \bbD, \pi_a, \pi_o)$, in a joint manner. Specifically, we define the agent-environment co-optimization problem as
\begin{align}\label{eq:agentEnvironmentCoProblem}
	&\argmax_{\pi_a, \pi_o}~ f(\bbS, \bbD, \pi_a, \pi_o)\\
	&~~~~\text{s.t.}~~~ \pi_{a}(\bbU_{a}|\bbX_a) \in \ccalU_{a},~\pi_o(\ccalO|\bbS,\bbD) \in \ccalP_{o},\nonumber
\end{align}
where $\ccalU_a$ is the action space of the agents and $\ccalP_o$ is the feasible set of the obstacle layout. The preceding problem is difficult due to the following challenges:

\smallskip
\noindent \textbf{(i)} The objective function $f(\bbS, \bbD, \pi_a, \pi_o)$ depends on the multi-agent system and the environment. It is, however, impractical to model this relationship analytically, hence, precluding the application of model-based methods. For the same reason, heuristic methods will perform poorly. 
	
\noindent \textbf{(ii)} The navigation policy $\pi_a(\bbU_a | \bbX_a)$ generates sequential agent actions based on instantaneous states across time steps, while the generative model $\pi_o(\ccalO|\bbS,\bbD)$ generates a one-shot (time-invariant) environment configuration based on the navigation task \textit{before} agents start to move. The navigation policy needs unrolling to be evaluated for any generated environment. Hence, $\pi_a$ and $\pi_o$ have different inner-working mechanisms and cannot be optimized in the same way, which complicates the joint optimization. 
	
\noindent \textbf{(iii)} The navigation policy $\pi_a(\bbU_a | \bbX_a)$ and the generative model $\pi_o(\ccalO|\bbS,\bbD)$ are arbitrary mappings from $\bbX_a$ and $(\bbS,\bbD)$ to $\bbU_a$ and $\ccalO$, which can take any function forms and thus are infinitely dimensional. 
	
\noindent \textbf{(iv)} The action space of the agents $\ccalU_a$ can be discrete or continuous and the feasible set $\ccalP_o$ can be convex or non-convex, leading to complex constraints on the feasible solution.

\smallskip
Due to the above challenges, we propose to solve problem \eqref{eq:agentEnvironmentCoProblem} with a model-free learning-based approach, in which we alternate between two optimization sub-objectives. Specifically, we parameterize the navigation policy $\pi_a(\bbU_a | \bbX_a)$ by a graph neural network with parameters $\bbtheta_a$, and the generative model $\pi_o(\ccalO|\bbS,\bbD)$ by a deep neural network with parameters $\bbtheta_o$. We update $\bbtheta_a$ with actor-critic reinforcement learning (RL), and $\bbtheta_o$ with a policy gradient ascent (unsupervised learning), in an alternating manner, to find an optimal pair $(\bbtheta_a^*, \bbtheta_o^*)$ that maximizes the navigation performance. The proposed approach overcomes challenge \textbf{(i)} with its model-free implementation, challenge \textbf{(ii)} with its alternating optimization, challenge \textbf{(iii)} with its finite parameterization, and challenge \textbf{(iv)} with its appropriate selection of policy / generative distributions.

\section{Coordinated Optimization Methodology}\label{sec:Method}

We begin by re-formulating problem \ref{eq:agentEnvironmentCoProblem} as an optimization problem with two sub-objectives:
\begin{align}\label{eq:agentEnvironmentCoProblem1}
	&\argmax_{\pi_a}~ f(\bbS, \bbD, \pi_a, \ccalE)~ \text{s.t.}~\pi_{a}(\bbU_{a}|\bbX_a) \in \ccalU_{a}, \\
	\label{eq:agentEnvironmentCoProblem2}
	&\argmax_{\pi_o}~ f(\bbS, \bbD, \pi_a, \pi_o)~\text{s.t.}~\pi_o(\ccalO|\bbS,\bbD) \in \ccalP_{o}.
\end{align} 
Sub-objective~\eqref{eq:agentEnvironmentCoProblem1} optimizes the navigation policy given the environment $\ccalE$, while sub-objective \eqref{eq:agentEnvironmentCoProblem2} optimizes the generative model given the navigation policy of the agents $\pi_a$.

\subsection{Navigation Policy}

We are interested in synthesizing a decentralized navigation policy $\pi_a$ that generates sequential agent actions as a function of local observations. 
To do so, we define a Markov decision process wherein, at time $t$, agents observe states $\bbX_a^{(t)} = [\bbx_{a1}^{(t)},..., \bbx_{an}^{(t)}]$ and take actions $\bbU_a^{(t)} = [\bbu_{a1}^{(t)},..., \bbu_{an}^{(t)}]$. Each agent $A_i$ only has access to its own state $\bbx_{ai}^{(t)}$ (e.g., position and velocity), and communicates with its neighbors to collect $\bbX_{a \ccalN_i}^{(t)} = \{\bbx_{aj}^{(t)}\}_{j \in \ccalN_i}$, where $\ccalN_i$ is the neighbor set within communication range. The navigation policy $\pi_a$ generates the action $\bbu_{ai}^{(t)}$ based on the local states $\{\bbx_{ai}^{(t)}, \bbX_{a \ccalN_i}^{(t)}\}$ for $i=1,...,n$, and these actions $\bbU_a^{(t)}$ control agents to transit from $\bbX_a^{(t)}$ to $\bbX_a^{(t+1)}$ based on the probability function $P_a(\bbX_a^{(t+1)}|\bbX_a^{(t)},\bbU_a^{(t)}, \ccalE)$, which is a distribution of states conditioned on the previous states, actions and environment. Let $r_{ai}(\bbX_a^{(t)},\!\bbU_a^{(t)},\ccalE)$ be the reward of agent $A_i$, representing the instantaneous navigation performance at time $t$, which depends on agent states $\bbX_a^{(t)}$, agent actions $\bbU_a^{(t)}$ and environment configurations $\ccalE$ (e.g., obstacle positions and sizes). The reward of the multi-agent system is the performance averaged over $n$ agents, i.e., $r_a^{(t)} \!\!=\!\! \sum_{i=1}^n r_{ai}(\bbX_a^{(t)}\!,\!\bbU_a^{(t)}\!,\! \ccalE) / n$. With the discount factor $\gamma$ that accounts for the future rewards, the expected discounted reward is 
\begin{align}\label{eq:expectedReward}
	R_a(\bbS, \bbD, \pi_a, \ccalE) = \mathbb{E} \Big[\sum_{t = 0}^\infty \gamma^{t} r_{a}^{(t)}\Big] 
\end{align}
where the initial states $\bbX_a^{(0)}$ are determined by the initial and goal positions $(\bbS, \bbD)$, and $\mathbb{E}[\cdot]$ is w.r.t. the navigation policy $\pi_a$ and the state transition probability $P_a$. By parameterizing $\pi_a$ with an information processing architecture $\bbPhi_a(\bbX_a, \bbtheta_a)$ of parameters $\bbtheta_a$, we can represent the expected discounted reward as $R_a(\bbS, \bbD, \bbtheta_a, \ccalE)$ and the goal is to find the optimal $\bbtheta^*_a$ that maximize $R_a(\bbS, \bbD, \bbtheta_a, \ccalE)$. The expected discounted reward in \eqref{eq:expectedReward} is equivalent to the sub-objective in~\eqref{eq:agentEnvironmentCoProblem1} and we can re-formulate the problem in the RL domain as
\begin{align}\label{eq:maxProblemRL}
	\argmax_{\bbtheta_a}~ R_a(\bbS, \bbD, \bbtheta_a, \ccalE)~ \text{s.t.}~\bbPhi_a(\bbX_a, \bbtheta_a) \in \ccalU_{a}.
\end{align}

\subsection{Generative Model}

We use a generative model to design the environment. Different from the navigation policy $\pi_a$ that synthesizes sequential time-varying agent actions, the generative model $\pi_o$ generates a static environment (i.e., obstacle layout) based on agents' initial and goal positions in one shot before agents start to move. This allows to re-formulate \eqref{eq:agentEnvironmentCoProblem2} as a stochastic optimization problem. To do so, we assume the initial positions $\bbS$, goal positions $\bbD$ and \textit{parameterized} navigation policy $\bbPhi_a(\bbX_a, \bbtheta_a)$ given, and thus, can measure the navigation performance with the expected discounted reward $R_a(\bbS, \bbD, \bbtheta_a, \ccalE)$ [cf. \eqref{eq:maxProblemRL}] in any environment $\ccalE$ generated by  the generative model $\pi_o$. By parameterizing $\pi_o$ with an information processing architecture $\bbPhi_o(\bbS, \bbD, \bbtheta_o)$ of parameters $\bbtheta_o$, we can represent the expected discounted reward as $R_a(\bbS, \bbD, \bbtheta_a, \bbtheta_o)$ because $\bbPhi_o(\bbS, \bbD, \bbtheta_o)$ determines the obstacle layout of the environment $\ccalE$, and use it as the objective function. The goal is to find the optimal $\bbtheta_o^*$ that maximizes $R_a(\bbS, \bbD, \bbtheta_a, \bbtheta_o)$. The stochastic optimization problem is, hence, formulated in an unsupervised manner as 
\begin{align}\label{eq:minProblemRL}
	\argmax_{\bbtheta_o}~ \mathbb{E}\big[R_a(\bbS, \bbD, \bbtheta_a, \bbtheta_o)\big]~ \text{s.t.}~\bbPhi_o(\bbS, \bbD, \bbtheta_o) \in \ccalP_{o},
\end{align}
where the expectation $\mathbb{E}[\cdot]$ is w.r.t. the generative model $\bbPhi_o(\bbS, \bbD, \bbtheta_o)$ and the navigation policy $\bbPhi_a(\bbX_a, \bbtheta_a)$.

\begin{remark}
    A potential question could be why we do not use reinforcement learning for the sub-problem of environment optimization, as well. The rationale is that in this sub-problem, the states are the navigation task and multi-agent policies given by the system, whereas the action is the obstacle layout produced by the generative model. Since the obstacle layout affects only the navigation performance but not the navigation task or multi-agent policies, the action does not affect the states and there is no transition probability between successive ``time'' steps. That is, historical information does not influence the current states and there exists no Markov decision process. Therefore, the sub-problem of environment optimization does not fit well into the framework of reinforcement learning. As states at each ``time'' step are independent (not correlated), it is better modeled as a stochastic optimization problem in an unsupervised manner, where the stochasticity is w.r.t. policy distributions of the generative model and multi-agent policies.
\end{remark}

\subsection{Coordinated Optimization}

With these preliminaries, we propose to solve the agent-environment co-optimization problem by coupling the two sub-objectives in an alternating manner. In particular, \eqref{eq:maxProblemRL} formulates multi-agent navigation as a reinforcement learning problem, while \eqref{eq:minProblemRL} formulates environment optimization as an unsupervised learning problem. The former optimizes the navigation parameters $\bbtheta_a$ given the environment $\ccalE$, while the latter optimizes the generative parameters $\bbtheta_o$ given the navigation policy $\bbPhi_a(\bbX_a, \bbtheta_a)$. We coordinate the optimization of these sub-objectives to search for an optimal pair $(\bbtheta_a^*, \bbtheta_o^*)$ that maximizes the navigation performance. Specifically, we update parameters iteratively, where each iteration consists of two phases: a multi-agent navigation phase and an environment optimization phase. Details of two phases are as follows. 

\smallskip
\noindent \emph{Synthesizing the navigation policy:} At iteration $k$, let $\bbtheta_a^{(k)}$ be the navigation parameters and $\bbtheta_o^{(k)}$ the generative parameters. Given the initial and goal positions $\bbS$ and $\bbD$, the generative model $\bbPhi_o(\bbS, \bbD, \bbtheta_o^{(k)})$ generates the obstacle layout $\ccalO^{(k)}$ and determines the environment $\ccalE^{(k)}$. At time $t$ with the agent states $\bbX^{(k,t)}_a$, the navigation policy $\bbPhi_a(\bbX^{(k,t)}_a, \bbtheta_a^{(k)})$ generates actions $\bbU_a^{(k, t)}$ that move the agents towards $\bbD$ with local neighborhood information. These actions lead to the new states $\bbX_a^{(k, t+1)}$ based on the transition probability $P_a(\bbX_a^{(k, t+1)}|\bbX_a^{(k, t)},\bbU_a^{(k, t)},\ccalE^{(k)})$ and receive a reward $r_a^{(k, t)}$.

We follow the actor-critic mechanism and make use of a value function to estimate the expected discounted rewards starting from the agent states $\bbX^{(k, t)}$ and $\bbX^{(k, t+1)}$ with the navigation parameters $\bbtheta^{(k)}_a$ in the environment $\ccalE^{(k)}$ [cf. \eqref{eq:expectedReward}], i.e., $V(\bbX_a^{(k, t)}, \bbtheta^{(k)}_a, \bbomega^{(k)}, \ccalE^{(k)})$ and $V(\bbX_a^{(k, t+1)}, \bbtheta^{(k)}_a, \bbomega^{(k)}, \ccalE^{(k)})$, where $\bbomega^{(k)}$ are the parameters of the value function at iteration $k$. The estimation error can be computed as
\begin{align}\label{eq:valueFunction}
	\delta = r_a^{(k, t)} &+ \gamma V(\bbX_a^{(k, t)}, \bbtheta^{(k)}_a, \bbomega^{(k)}, \ccalE^{(k)}) \\
	&- V(\bbX_a^{(k, t+1)}, \bbtheta^{(k)}_a, \bbomega^{(k)}, \ccalE^{(k)}), \nonumber
\end{align}
which is used to update $\bbomega^{(k)}$ of the value function. Then, the navigation parameters $\bbtheta^{(k)}_a$ are updated with a policy gradient 
\begin{align}\label{eq:policyGradient}
	&\bbtheta^{(k)}_{a, 1} = \bbtheta^{(k)}_a + \Delta \alpha \nabla_{\bbtheta} \log \pi_a(\bbU_a^{(k)} | \bbX_a^{(k)}) \delta, 
\end{align}
where $\Delta \alpha$ is the step-size and $\pi_a(\bbU_a^{(k)} | \bbX_a^{(k)})$ is the policy distribution specified by the navigation parameters $\bbtheta^{(k)}_a$. The update in \eqref{eq:policyGradient} is model-free because it does not require the theoretical model (i.e., analytic expression) of the objective function $R_a(\bbX^{(k, t)}, \bbtheta^{(k)}_a, \ccalE^{(k)})$, but only the error value $\delta$ and the gradient of the policy distribution. 
The error value $\delta$ can be obtained through \eqref{eq:valueFunction}, and the policy distribution is known by design choice. By performing the above update recursively $\rho_a$ times, we get $\bbtheta^{(k)}_{a,1}, \bbtheta^{(k)}_{a,2},\ldots,\bbtheta^{(k)}_{a,\rho_a}$ for some $\rho_a \in \mathbb{Z}_+$. After $\rho_a$ updates, we transition to the environment optimization phase, with navigation parameters updated by $\bbtheta^{(k+1)}_a:= \bbtheta^{(k)}_{a,\rho_a}$.

\smallskip
\noindent \emph{Learning to generate environments:} With the (updated) navigation parameters $\bbtheta^{(k+1)}_a$ and the initial and goal positions $\bbS$ and $\bbD$, we can measure the multi-agent navigation performance and leverage the latter to update the generative parameters $\bbtheta_o^{(k)}$. Specifically, the generative model $\bbPhi_o(\bbS, \bbD, \bbtheta_o^{(k)})$ generates the obstacle layout $\ccalO^{(k)}$ based on agents' initial and goal positions $(\bbS, \bbD)$ to determine the environment $\ccalE^{(k)}$. In this environment, we run the updated navigation policy $\bbPhi_a(\bbX^{(k)}_a, \bbtheta_a^{(k+1)})$ to move agents from $\bbS$ to $\bbD$. Doing so results in the objective value $R_a^{(k)}:=R_a(\bbS, \bbD, \bbtheta^{(k+1)}_a, \bbtheta_o^{(k)})$ representing the navigation performance of $\bbPhi_a(\bbX^{(k)}_a, \bbtheta_a^{(k+1)})$ in $\ccalE^{(k)}$, and allows us to update $\bbtheta_o^{(k)}$ with a gradient ascent 
\begin{align}\label{eq:optimizationParametersUpdate}
	&\bbtheta_{o,1}^{(k)} = \bbtheta_o^{(k)} + \Delta \beta \ \nabla_{\bbtheta_o} \mathbb{E}\big[R_a^{(k)} \big],
\end{align}
where $\Delta \beta$ is the step-size, and $\mathbb{E}[\cdot]$ is w.r.t. $\bbPhi_o(\bbS, \bbD, \bbtheta_o^{(k)})$ and $\bbPhi_a(\bbX^{(k)}_a, \bbtheta_a^{(k+1)})$. However, \eqref{eq:optimizationParametersUpdate} requires computing the gradient $\nabla_{\bbtheta_o} \mathbb{E}\big[R_a^{(k)} \big]$, which, in turn, requires a closed-form differentiable model of the objective function $R_a^{(k)}$. The latter is not available as demonstrated in challenge (i) of Sec. \ref{subsec:AgentEnvironment}. 

\begin{algorithm}[tb]
	\caption{Coordinated Optimization} 
	\label{alg:AEAlgorithm}
	\begin{algorithmic}
		\STATE {\bfseries Input:} initial agent positions $\bbS$, goal positions $\bbD$, navigation parameters $\bbtheta_a^{(0)}$, generative parameters $\bbtheta_o^{(0)}$ 
		\FOR{$k=0, 1, 2, \ldots$}
		\STATE \emph{(i) Synthesizing the multi-agent navigation policy:}
		\STATE 1. Generate the obstacle layout $\ccalO^{(k)}$ with generative model $\bbPhi_o(\bbS,\! \bbD,\! \bbtheta_o^{(k)}\!)$ and determine \!the\! environment $\ccalE^{(\!k\!)}$
		\STATE 2. Follow the actor-critic mechanism to update navigation parameters $\bbtheta_a^{(k)}$ with policy gradient $\rho_a$ times [cf. \eqref{eq:policyGradient}]
		\STATE 3. Define new navigation parameters $\bbtheta^{(k+1)}_a= \bbtheta^{(k)}_{a,\rho_a}$
		\STATE \emph{(ii) Learning to generate the environment:}
		\STATE 4. Compute the navigation reward $R_a(\bbS, \bbD, \bbtheta^{(k+1)}_a, \ccalE^{(k)})$ with new navigation parameters $\bbtheta^{(k+1)}_a$
		\STATE 5. Compute the policy gradient $\nabla_{\bbtheta_o} \mathbb{E}\big[R_a^{(k)} \big]$ [cf. \eqref{eq:policGradient}]
		\STATE 6. Update generative parameters $\bbtheta_o^{(k)}$ with gradient ascent $\rho_o$ times and define new parameters $\bbtheta^{(k+1)}_o= \bbtheta^{(k)}_{o,\rho_o}$
		\ENDFOR
	\end{algorithmic}
\end{algorithm}

We overcome this issue by combing the gradient ascent with the policy gradient. In particular, the policy gradient provides a stochastic and model-free approximation for the gradient of policy functions by exploiting the likelihood ratio property \cite{sutton1999policy}. This allows us to estimate the gradient in \eqref{eq:optimizationParametersUpdate} as
\begin{align}\label{eq:policGradient}
	&\nabla_{\bbtheta_o} \mathbb{E}\big[R_a^{(k)} \big] = \mathbb{E}\big[ R_a^{(k)} \nabla_{\bbtheta_o}\! \log \pi_o(\ccalO^{(k)}|\bbS,\bbD)\big], 
\end{align}
where $\pi_o(\ccalO^{(k)}|\bbS,\bbD)$ is the probability distribution of the obstacle layout specified by the generative parameters $\bbtheta^{(k)}_o$, referred to as the generative distribution. It translates the computation of the function gradient into a multiplication of two terms: (i) the function value $R_a^{(k)}$ and (ii) the gradient of the generative distribution $\nabla_{\bbtheta_o}\! \log \pi_o(\bbO^{(k)}|\bbS,\bbD)$. The former is available with the observed objective value, and the latter can be computed because the generative distribution is known by design choice. Therefore, \eqref{eq:policGradient} is model-free, i.e., it can be implemented without any knowledge about the analytic expression of the objective function. The expectation $\mathbb{E}[\cdot]$ can be approximated by unrolling the generative model and the navigation policy $\ccalT$ times and taking the average. We perform this update $\rho_o \!\in\! \mathbb{Z}_+$ times to get $\bbtheta^{(k)}_{o,1}, \bbtheta^{(k)}_{o,2},...,\bbtheta^{(k)}_{o,\rho_o}$, and the updated environment generative parameters are $\bbtheta^{(k+1)}_o:= \bbtheta^{(k)}_{o,\rho_o}$. Algo. \ref{alg:AEAlgorithm} gives an overview of our coordinated method, which can be conducted offline. 

In summary, the coordinated optimization method \textit{couples} multi-agent navigation with environment generation. It optimizes the agents' navigation policy with RL for long-term reward maximization and the generative model with unsupervised learning for one-shot stochastic optimization, representing a new way of combining two machine learning paradigms. The resulting navigation policy and generative model adapt to each other, jointly maximizing the performance. Fig. \ref{fig:methodology} summarizes this approach. Note that an end-to-end learning without coordination (either RL or unsupervised learning alone) is ill-suited for agent-environment co-optimization because the two models have different inner-working mechanisms, i.e., the navigation policy optimizes sequential actions for a long-term reward [cf. \eqref{eq:expectedReward}] but the generative model generates a one-shot environment for an instantaneous objective value [cf. \eqref{eq:minProblemRL}]. Moreover, our coordinated scheme is more interpretable and provides a wider range of customization options. 

\begin{figure*}[!t]
\centering
    \resizebox{0.65 \textwidth}{!}{\input{Methodology}}

\caption{General framework of the agent-environment coordinated optimization method, which alternates between two phases. Given the environment configuration $\ccalE$, the multi-agent navigation phase updates the navigation policy $\bbPhi_a(\bbtheta_a)$ through reinforcement learning. 
Given the navigation policy $\bbPhi_a(\bbtheta_a)$, the environment optimization phase updates the generative model $\bbPhi_o(\bbtheta_o)$ with policy gradient ascent based on an unsupervised stochastic optimization problem \eqref{eq:minProblemRL}. The coordinated optimization method conducts these two phases in an alternating manner, searching for an optimal pair of the environment configuration and the agents' policy.}
\label{fig:methodology}\vspace{-6mm}
\end{figure*}

\section{Convergence Analysis}

In this section, we analyze the convergence of the proposed method and provide an interpretation for the coordinated optimization scheme by exploring its relation with ordinary differential equations. Specifically, solving the agent-environment co-optimization problem \eqref{eq:agentEnvironmentCoProblem} with our method is equivalent to solving a time-varying non-convex optimization problem
\begin{align}\label{eq:nonlinearOptimization}
	&\min_{\bbtheta_o}~~ g(\bbS, \bbD, \bbtheta_a^{(k)}, \bbtheta_o)~~\text{s.t.}~~~ \bbPhi_o(\bbS,\bbD,\bbtheta_o) \in \ccalP_{o} 
\end{align}
with policy gradient ascent of step-size $\Delta \beta$, where $g(\bbS, \bbD, \bbtheta_a^{(k)}, \bbtheta_o) = -f(\bbS, \bbD, \bbtheta_a^{(k)}, \bbtheta_o)$ is the inverse objective function, the navigation policy $\pi_a$ and the generative model $\pi_o$ are represented by their parameters $\bbtheta_a^{(k)}$ and $\bbtheta_o$, and $\bbtheta_a^{(k)}$ are updated by the actor-critic mechanism with step-size $\Delta \alpha$ [cf. \eqref{eq:policyGradient}]. Problem \eqref{eq:nonlinearOptimization} considers the co-optimization problem from the perspective of environment optimization and regards the navigation policy as time-varying parameters that modify its optimization objective across iterations $k$. The problem is typically non-convex because of the unknown function $g$ and the nonlinear parameterization $\bbPhi_o$. 

The time-varying problem \eqref{eq:nonlinearOptimization} changes with the navigation parameters $\bbtheta_a$, which motivates the definition of a ``time'' variable $\alpha$ to capture the inherent variation. In particular, $\bbtheta_a$ is updated by the actor-critic mechanism with policy gradient [cf. \eqref{eq:policyGradient}]. By assuming the policy gradient is bounded, the variation of $\bbtheta_a$ is determined by the step-size $\Delta \alpha$. In this context, we can represent $\bbtheta_a^{(k)}$ as a function of the initial $\bbtheta_a^{(0)}$, the step-size $\Delta \alpha$ and the number of iterations $k$, i.e., $\bbtheta_a^{(k)} = \Theta\big(\bbtheta_a^{(0)}, \alpha^{(k)} \big)$ where $\alpha^{(k)} = k \Delta \alpha$ is the step length. Define a continuous ``time'' variable $\alpha$, which is instantiated as $\alpha^{(k)}$ at iteration $k$. By combining this definition with problem \eqref{eq:nonlinearOptimization}, we can formulate a continuous time-varying problem as
\begin{align}\label{eq:nonlinearOptimization2}
	&\min_{\bbtheta_o}~~~ g(\bbS,\bbD, \alpha,\bbtheta_o)~~\text{s.t.}~~~ \bbPhi_o(\bbS,\bbD,\bbtheta_o) \in \ccalP_{o},
\end{align}
where $\alpha \ge 0$ denotes the scale of ``time'' and $\bbtheta_o$ are the decision variables. Problem \eqref{eq:nonlinearOptimization2} is the continuous limit of \eqref{eq:nonlinearOptimization}, where the former reduces to the latter if particularizing $\alpha$ to discrete values $\{\alpha^{(k)}\}_k$ corresponding to different iterations. Because of the non-convexity, we consider the first-order stationary condition as the metric for convergence analysis and define the local minimum trajectory of problem \eqref{eq:nonlinearOptimization2}. 
\begin{definition}[Local minimum trajectory]\label{def:minimumTrajectory}
	\emph{A $p$-dimensional continuous trajectory $\ccalT(\alpha):[0,+\infty] \to \mathbb{R}^p$ is a local minimum trajectory of the non-convex time-varying optimization problem \eqref{eq:nonlinearOptimization2} if each point of $\ccalT(\alpha)$ is a local minimum of problem \eqref{eq:nonlinearOptimization2} at ``time'' $\alpha \in [0, +\infty)$.}
\end{definition} 
The local minimum trajectory $\ccalT(\alpha)$ tracks the sequence of local solutions for the continuous non-convex time-varying optimization problem \eqref{eq:nonlinearOptimization2}.\footnote{The local minimum trajectory is a sequence of local minimum solutions that lie in the model parameter space, not the robot physical space.} If freezing the ``time'' variable $\alpha$ at a particular value, problem \eqref{eq:nonlinearOptimization2} becomes a classic (time-invariant) non-convex optimization problem and the local minimum trajectory becomes a stationary local solution. 
\begin{assumption}\label{as1}
	\emph{The inverse objective function $g(\bbS,\bbD, \alpha,\bbtheta_o)$ of problem \eqref{eq:nonlinearOptimization2} is differentiable w.r.t. $\alpha$ and $\bbtheta_o$, and the gradient is Lipschitz continuous w.r.t. $\bbtheta_o$, i.e.,
	\begin{align}
		\|\nabla_{\bbtheta_o} g(\bbS,\!\bbD,\! \alpha,\!\bbtheta_{o,1}) \!-\!\! \nabla_{\bbtheta_o}g(\bbS,\!\bbD,\!\alpha,\!\bbtheta_{o,2})\| \!\!\le\! C_L \|\bbtheta_{o,1} \!-\! \bbtheta_{o,2}\|\nonumber
	\end{align}
	for any $\bbtheta_{o,1}, \bbtheta_{o,2} \in \mathbb{R}^p$, where $\|\cdot\|$ is the vector norm, $C_L$ is the Lipschitz constant, and $p$ is the dimension of $\bbtheta_o$.}
\end{assumption}
\begin{assumption}\label{as2}
	\emph{The policy gradient in \eqref{eq:policGradient} estimates the true gradient in \eqref{eq:optimizationParametersUpdate} within an error neighborhood of size $\varepsilon$.} 
\end{assumption}
Assumption \ref{as1} indicates that the gradient of the inverse objective function $\nabla_{\bbtheta_o} g(\bbS,\bbD, \alpha,\bbtheta_o)$ does not change faster than linear w.r.t. the generative parameters $\bbtheta_o$, which is a standard assumption in optimization theory \cite{boyd2004convex}. Here, the Lipschitz continuity is assumed w.r.t. the objective function rather than the designed reward. The objective function is an analytic expression characterizing the relationship between the environment, agents and performance, which is challenging to model due to their unclear relationship, while the designed reward approximates the value of the objective function and leverages the latter value to estimate the gradient of the objective function for optimization. Assumption \ref{as2} implies that the policy gradient approximates the true gradient within an error neighborhood $\varepsilon$. The value of $\varepsilon$ depends on the performance of the policy gradient, which has exhibited success in a wide array of reinforcement learning problems \cite{sutton1999policy, grondman2012survey}. 

\begin{figure}
\centering
\begin{subfigure}{0.45\columnwidth}
	\includegraphics[width=1.0\linewidth,height = 0.9\linewidth]{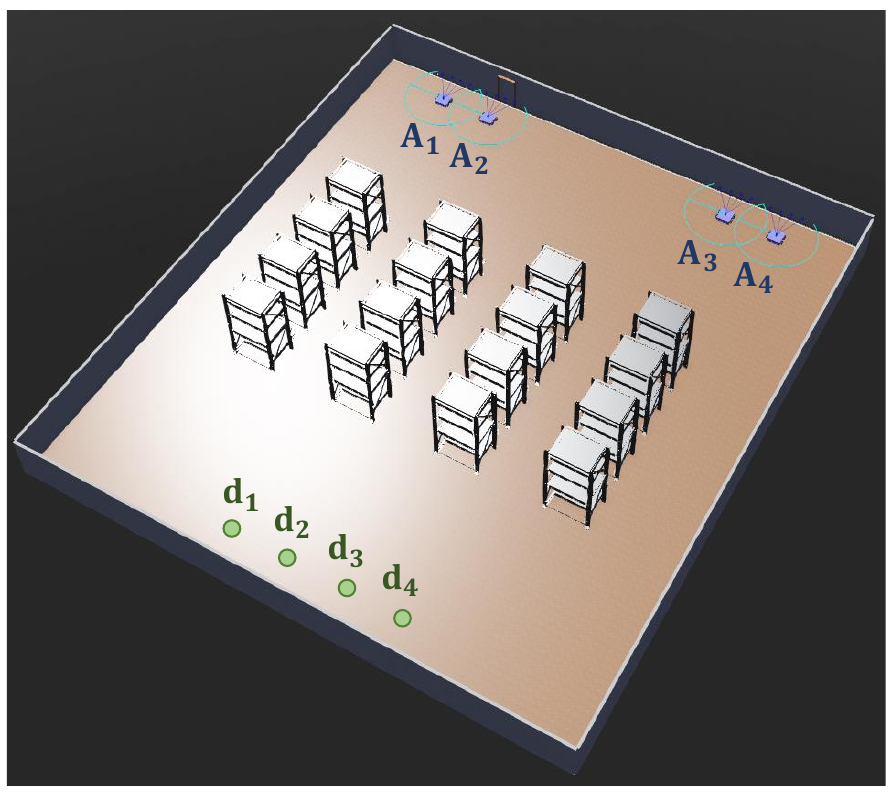}%
	\caption{Side view}%
	\label{subfig:environmentside}%
\end{subfigure}
\begin{subfigure}{0.45\columnwidth}
	\includegraphics[width=1.0\linewidth, height = 0.9\linewidth]{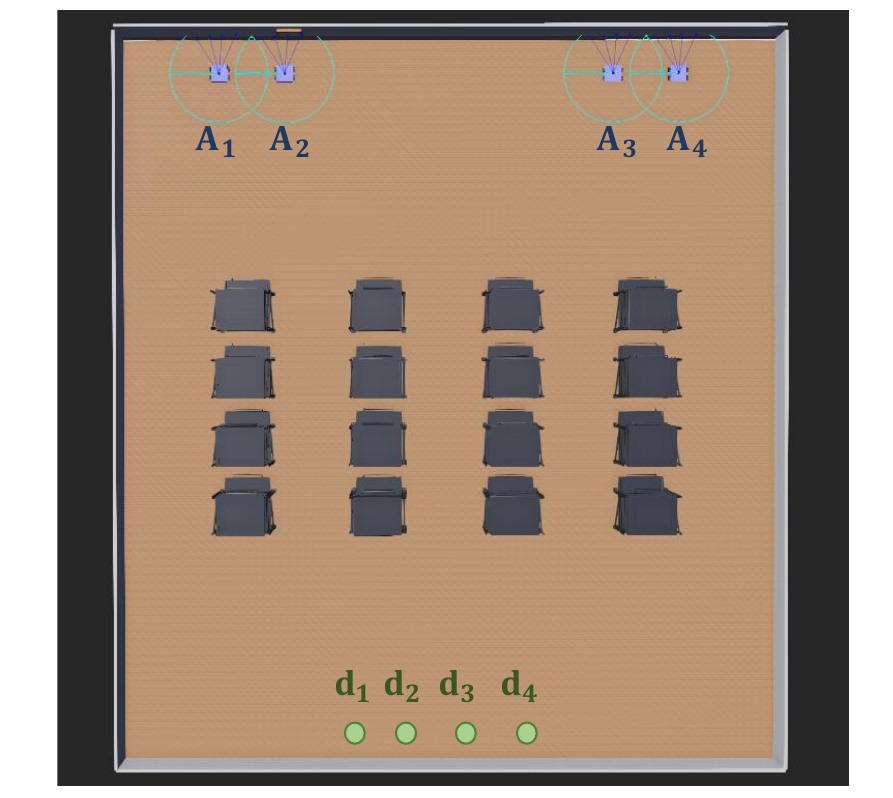}%
	\caption{Top-down view}%
	\label{subfig:environmenttop}
\end{subfigure}
\caption{Environment scenario with $4$ agents and $4$ obstacles in a warehouse setting. The agents $\{A_i\}_{i=1}^4$ represent robots, the obstacles represent rectangular shelves of size $1 \times 4$, and the green circles $\{\bbd_i\}_{i=1}^4$ represent the goal positions. (a) Side view of the environment. (b) Top-down view of the environment.}\label{fig:step1_environment}\vspace{-6mm}
\end{figure}

In real-world applications, it is neither practical nor realistic to have solutions that abruptly change over ``time''. It is standard to impose a soft constraint that penalizes the inverse objective function with the deviation of its solution from the one obtained at the previous ``time'' \cite{Ding2021}. The resulting  discrete time-varying optimization problem with proximal regularization (except for the initial problem) becomes
\begin{align}\label{eq:nonlinearOptimization3}
	&\min_{\bbtheta_o}~ g(\bbS,\bbD, \alpha^{(0)},\bbtheta_o)\\
	\label{eq:nonlinearOptimization4}&\min_{\bbtheta_o}~ g(\bbS,\!\bbD,\! \alpha^{(k)},\!\bbtheta_o) \!+\! \frac{\eta \| \bbtheta_o \!-\! \bbtheta_o^{(k\!-\!1)*}\|}{2\Delta \alpha},\text{for}~k\!=\!1,2,...
\end{align}
where $\bbtheta_o^{(k-1)*}$ is a local minimum of problem \eqref{eq:nonlinearOptimization4} at iteration $k\!-\!1$ and $\eta$ is a regularization parameter that weighs the regularization term to the inverse objective function. The first term in \eqref{eq:nonlinearOptimization4} is the inverse objective, while the second term in \eqref{eq:nonlinearOptimization4} denotes the deviation of the decision variable $\bbtheta_o$ at current iteration $k$ from the local minimum at previous iteration $k-1$. By taking the derivative and using the first-order stationary condition, the local minimum of \eqref{eq:nonlinearOptimization4} at iteration $k$ satisfies
\begin{align}\label{eq:discreteODE}
	\nabla_{\bbtheta_o} g(\bbS,\bbD, \alpha^{(k)},\bbtheta_o^{(k)*}) \!+\! \eta \frac{\bbtheta_o^{(k)*} \!-\! \bbtheta_o^{(k-1)*}}{\Delta \alpha} \!=\! 0,
\end{align}
where the first term represents the inverse objective gradient and the second term represents the change of the local minimum at two successive iterations. When the step-size approaches zero $\Delta \alpha \to 0$, the continuous limit of \eqref{eq:discreteODE} becomes the ordinary differential equation (ODE) as
\begin{align}\label{eq:ODE}
	\eta \bbtheta_o'(\alpha) \!=\! -\! \nabla_{\bbtheta_o} g(\bbS,\bbD, \alpha,\bbtheta_o(\alpha)),~\bbtheta_o(0) \!=\! \bbtheta_o^{(0)*},
\end{align}
where $\bbtheta_o(\alpha)$ is a dynamic function of $\alpha$ and $\bbtheta'_o(\alpha)$ is the derivative of $\bbtheta_o(\alpha)$ w.r.t. $\alpha$. The solution of the ODE \eqref{eq:ODE} is a local minimum trajectory of the continuous time-varying optimization problem with proximal regularization [cf. \eqref{eq:nonlinearOptimization3}-\eqref{eq:nonlinearOptimization4} with $\Delta \alpha \to 0$]. When $\eta = 0$, the ODE \eqref{eq:ODE} reduces to the first-order stationary condition of the time-varying optimization problem \eqref{eq:nonlinearOptimization2}. Given the initial point $\bbtheta_o(0)$, we assume there exists a local minimum trajectory $\ccalT(\alpha)$ [Def. \ref{def:minimumTrajectory}] satisfying that $\bbtheta_o(\alpha) - \ccalT(\alpha)$ lies in a compact set for $\alpha \ge 0$. With these preliminaries, we formally show the convergence of the proposed coordinated optimization method. 

\begin{theorem}\label{thm:Convergence}
	Consider the non-convex time-varying optimization problem \eqref{eq:nonlinearOptimization2} satisfying Assumption \ref{as1} w.r.t. $C_L$ and the policy gradient in \eqref{eq:optimizationParametersUpdate} satisfying Assumption \ref{as2} w.r.t. $\varepsilon$. Let $\bbtheta_o(\alpha)$ be a solution of the ODE \eqref{eq:ODE} with the regularization parameter $\eta$ and $\big\{\bbtheta_o^{(k)}\big\}_k$ be the discrete sequence of the generative parameters updated by the proposed method, where the step-size of the policy gradient ascent is $\Delta \beta = \Delta \alpha / \eta$ [cf. \eqref{eq:optimizationParametersUpdate}]. Let also the initial $\bbtheta_o^{(0)} = \bbtheta_o(0)$ be a local minimum of $g(\bbS,\bbD, 0,\bbtheta_o)$. Then, for any $\eps > 0$ and ``time'' horizon $T$, there exists a step-size $\Delta \alpha$ such that
	\begin{align}
		\|\bbtheta_o^{(k)} - \bbtheta_o(\alpha^{(k)})\| \le \eps + C \varepsilon 
	\end{align}
	for all iterations $k = 1, \ldots, \lfloor T/\Delta \alpha \rfloor$ within the horizon, where $C$ is a constant depending on $C_L$, $\eta$ and $T$ [cf. \eqref{proof:thm1eq14}].
\end{theorem}

\begin{proof}
	See Appendix \ref{proof:Theorem1}. 
\end{proof}

\begin{figure*}%
	\centering
	\begin{subfigure}{0.45\columnwidth}
		\includegraphics[width=1.2\linewidth,height = 0.9\linewidth]{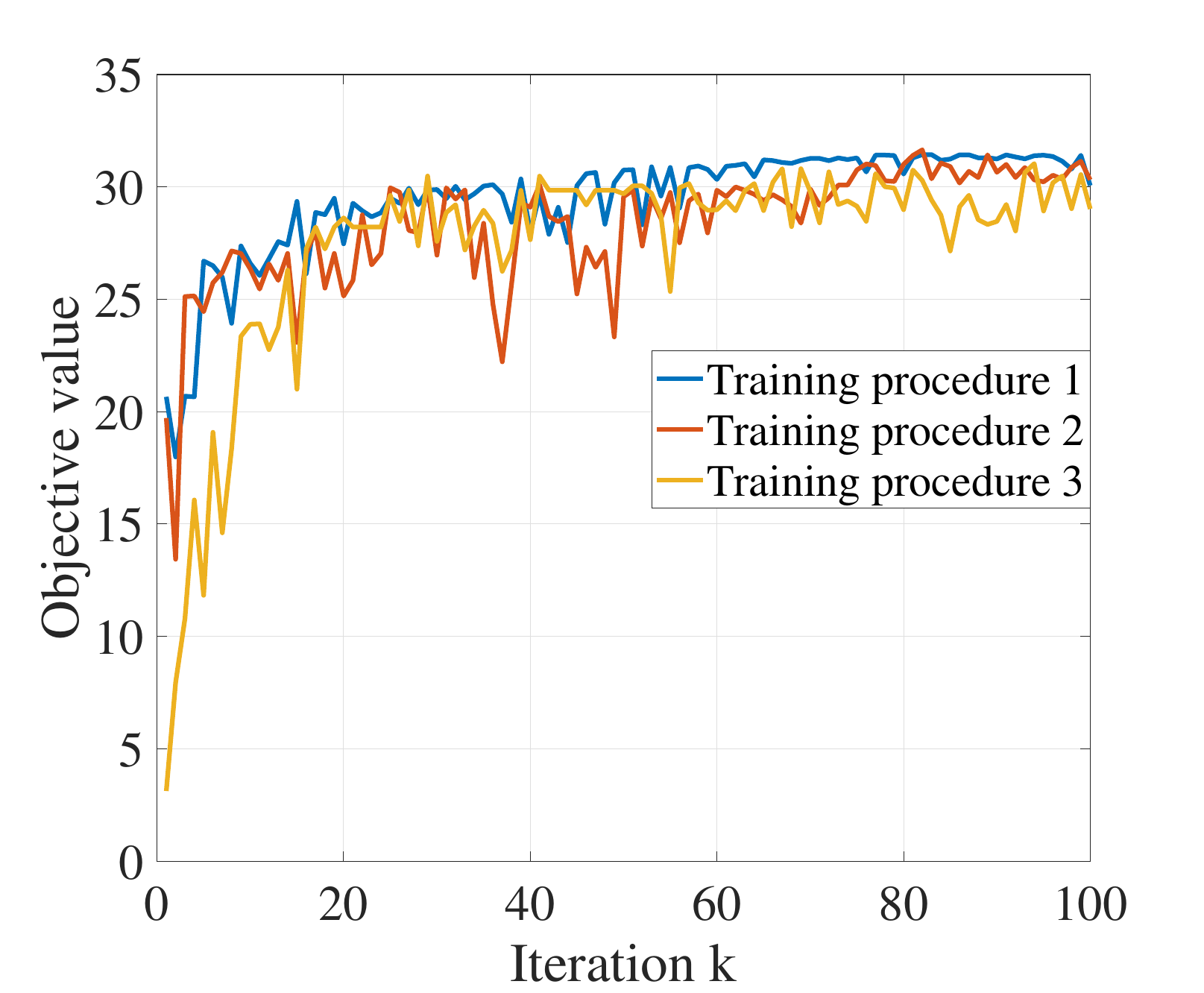}%
		\caption{}%
		\label{subfig:Proofconvergence}%
	\end{subfigure}\hfill\hfill%
	\begin{subfigure}{0.45\columnwidth}
		\includegraphics[width=1.2\linewidth, height = 0.9\linewidth]{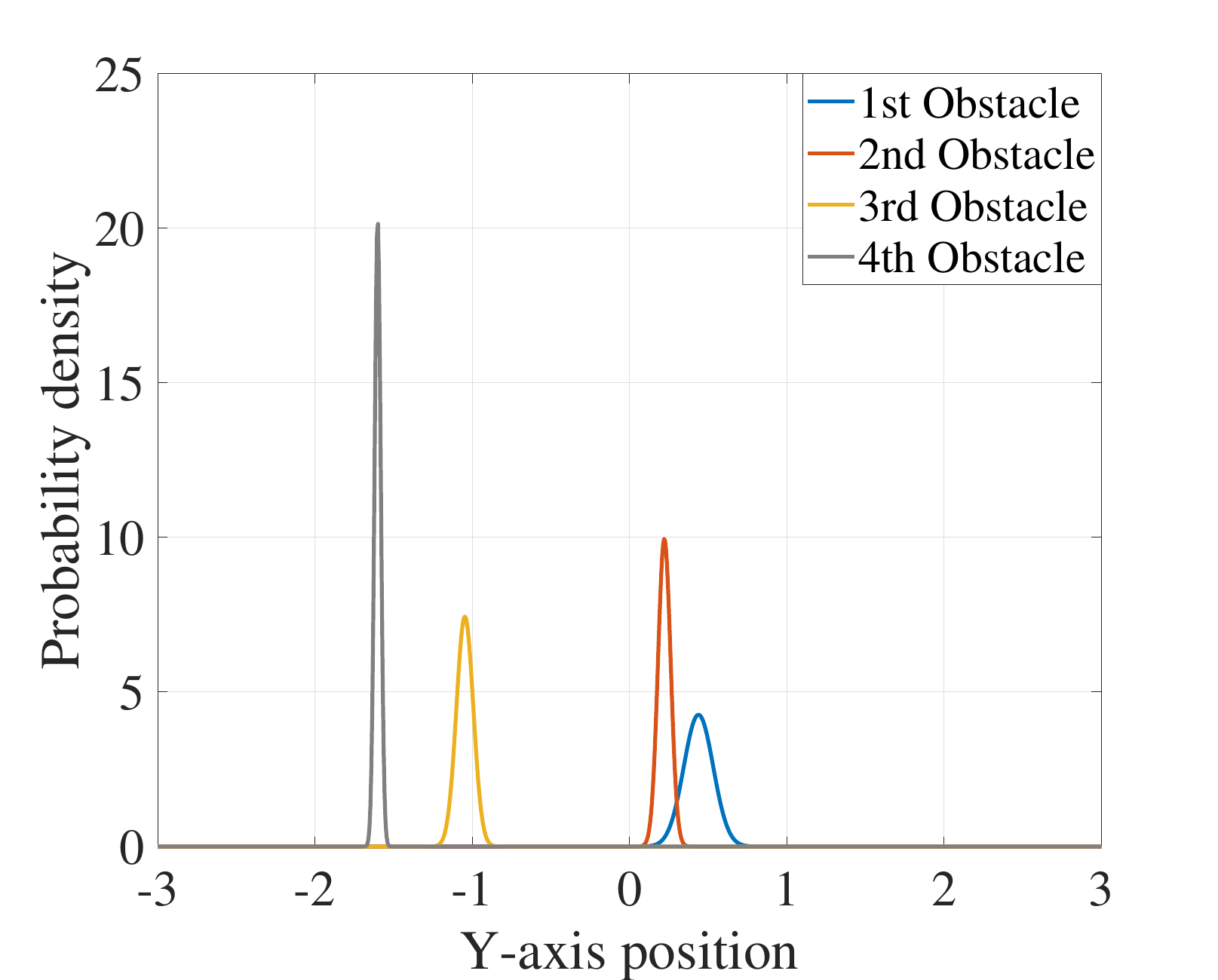}%
		\caption{}%
		\label{subfig:Proofdistribution}
	\end{subfigure}\hfill\hfill%
	\begin{subfigure}{0.45\columnwidth}
		\includegraphics[width=1\linewidth,height = 0.9\linewidth]{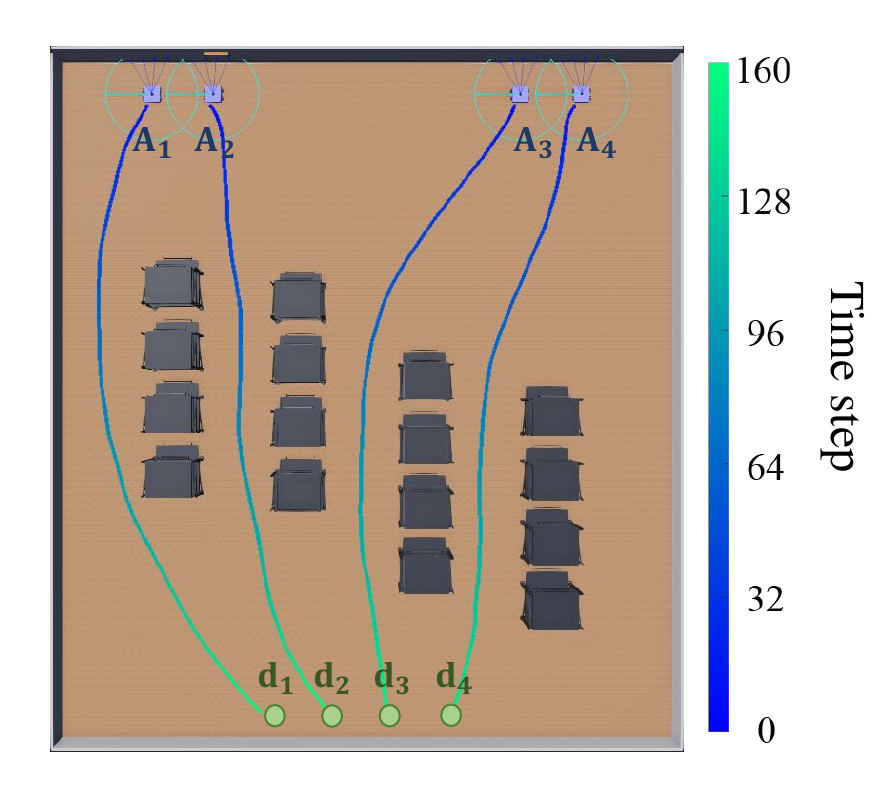}%
		\caption{}%
		\label{subfig:Prooftrajectories}%
	\end{subfigure}
	\caption{Proof of concept for agent-environment coordinated optimization. (a) Training procedure of coordinated optimization method. (b) Generative distributions for $4$ obstacles. (c) Obstacle layout produced by the generative model and agent trajectories controlled by the navigation policy. Blue-to-green lines are trajectories of agents and the color bar represents the time scale.}\label{fig:step1}\vspace{-4mm}
\end{figure*}

Theorem \ref{thm:Convergence} states that the proposed co-optimization method converges to the local minimum trajectory of a time-varying non-convex optimization problem, i.e., the solution of the ODE \eqref{eq:ODE}, up to a limiting error neighborhood. The latter consists of two additive terms: \textit{i)} the first term $\eps$ is the desired error, which can be arbitrarily small; \textit{ii)} the second term is proportional to the accuracy of the policy gradient $\varepsilon$, which becomes null as the policy gradient approaches the true gradient. The update step-sizes of the navigation policy $\Delta \alpha$ and the generative model $\Delta \beta = \Delta \alpha / \eta$ depend on the Lipschitz constant $C_L$, the regularization parameter $\eta$, the desired error $\eps$ and the ``time'' horizon $T$ [cf. \eqref{proof:thm1eq14}]. This indicates that a more accurate solution, i.e., a smaller $\eps$, requires a smaller step-size $\Delta \alpha$, which may slow down the training procedure. We note that the total number of iterations $\lfloor T/\Delta \alpha \rfloor$ is inversely proportional to the step size $\Delta \alpha$, which increases as $\Delta \alpha$ decreases and keeps the ``time'' horizon $T$ unchanged; hence, Theorem 1 holds uniformly in any given ``time'' horizon $\alpha^{(k)} \in [0, T]$. This result characterizes the convergence behavior of the proposed method and interprets its inherent working mechanism, i.e., performing the proposed method is equivalent to tracking the local minimum trajectory of an associated time-varying non-convex optimization problem.

\begin{figure*}%
\centering
\begin{subfigure}{0.45\columnwidth}
	\includegraphics[width=1\linewidth,height = 0.9\linewidth]{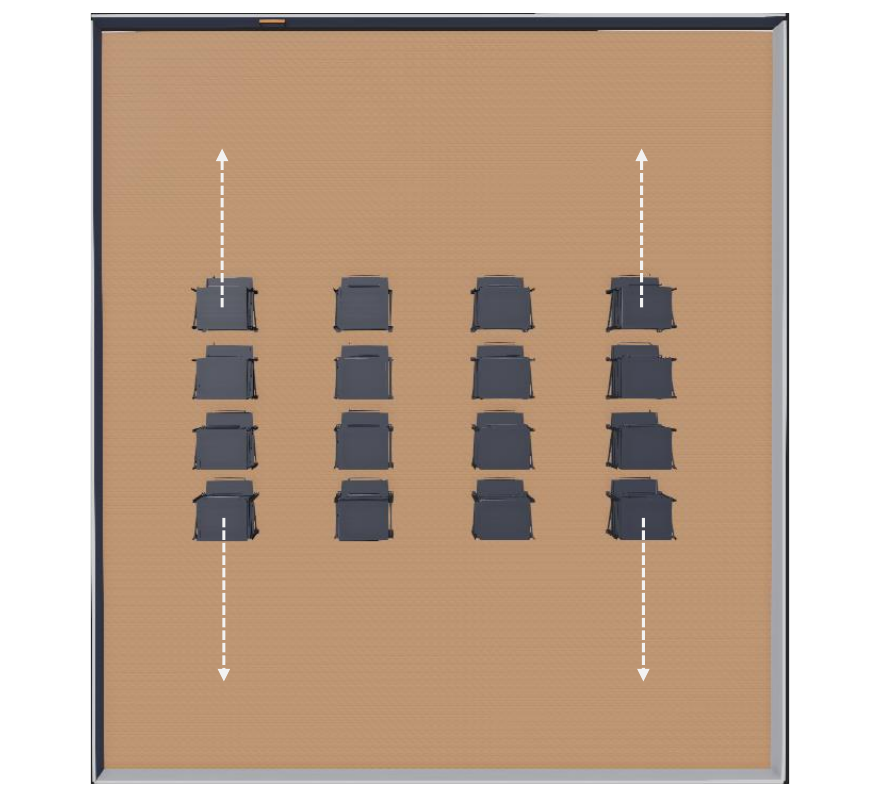}%
	\caption{Environment I}%
	\label{subfig:environment1}%
\end{subfigure}\hfill\hfill%
\begin{subfigure}{0.45\columnwidth}
	\includegraphics[width=1.\linewidth, height = 0.9\linewidth]{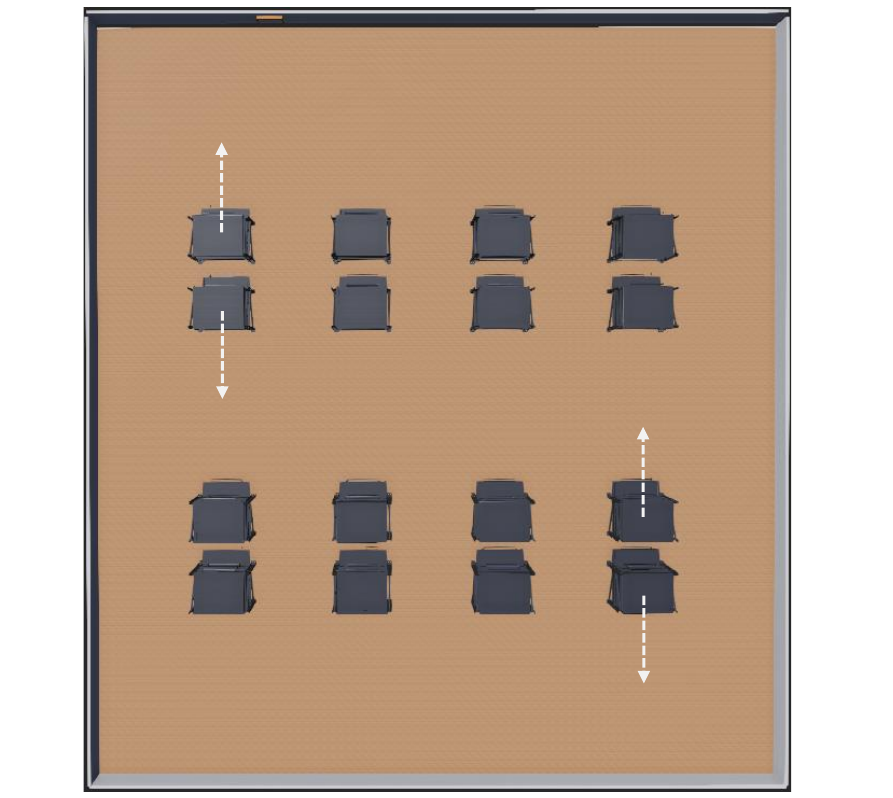}%
	\caption{Environment II}%
	\label{subfig:environment2}
\end{subfigure}\hfill\hfill%
\begin{subfigure}{0.45\columnwidth}
	\includegraphics[width=1.\linewidth,height = 0.9\linewidth]{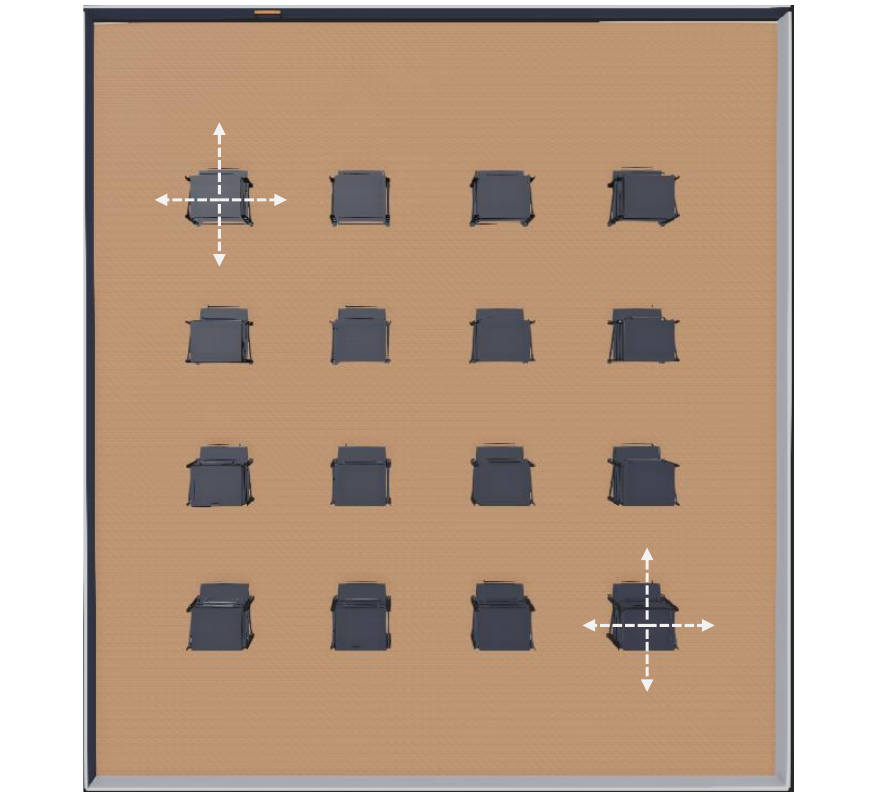}%
	\caption{Environment III}%
	\label{subfig:environment3}%
\end{subfigure}
\caption{Three environments in Section \ref{subsec:performance}. (a) Environment I with $4$ obstacles of size $1 \times 4$. (b) Environment II with $8$ obstacles of size $1 \times 2$. (c) Environment III with $16$ obstacles of size $1 \times 1$.}\label{fig:step21}\vspace{-6mm}
\end{figure*}

\section{Experiments}\label{sec:experiments}

In this section, we evaluate the coordinated optimization method. First, we conduct a proof of concept to demonstrate its effectiveness. Then, we show how our approach improves the navigation performance in a warehouse setting. Lastly, we characterize the role of the environment in the multi-agent system through our method and show that an appropriate obstacle layout provides ``positive'' guidance for agents, beyond ``negative'' obstruction as in conventional beliefs. This finding has not been explored thus far, highlighting the importance of building a symbiotic agent-environment co-evolved system. All experiments are performed on a computer with an Intel Core i7-11700 CPU and a GeForce RTX 3060 GPU.\footnote{Parallel computation with more CPUs and GPUs can speed up our method, depending on specific time requirements in practical applications.}

\subsection{Implementation Details}

The agents and obstacles are modeled by their positions $\{\bbp_{ai}\}_{i=1}^n$, $\{\bbp_{oj}\}_{j=1}^m$ and velocities $\{\bbv_{ai}\}_{i=1}^n$. For multi-agent navigation, each agent observes its own state (e.g., position and velocity), communicates with neighboring agents, and generates the desired velocity with received neighborhood information. We integrate an acceleration-constrained mechanism for position changes and an episode ends if all agents reach destinations or episode times out. The maximal acceleration is $1m/s^2$, the maximal velocity is $1.5m/s$, the communication radius is $2m$, the maximal time is $500$ steps and each time step is $0.05s$. The goal is to make agents reach destinations quickly while avoiding collision. We define the reward function as
\begin{align}
	r_{ai}^{(t)} \!\!=\!\! \Big(\!\frac{\bbp_{ai}^{(t)}\!-\!\bbd_i}{\|\bbp_{ai}^{(t)}\!-\!\bbd_i\|_2} \cdot \frac{\bbv_{ai}^{(t)}}{\|\bbv_{ai}^{(t)}\|_2} \!\Big) \|\bbv_{ai}^{(t)}\|_2 \!+\! C_p^{(t)}~\text{at time step}~t. \nonumber
\end{align}
The first term rewards fast movement to the goal position at the desired orientation, and the second term represents the collision penalty with $C_p^{(t)}\!=\!-C_p$ if there exists a collision and $C_p^{(t)}=0$ otherwise, where $C_p$ is a positive constant. We parameterize the navigation policy with a single-layer GNN, where message aggregation and feature update functions are multi-layer perceptrons. We train the policy using PPO \cite{schulman2017proximal}.

The environment generative model produces $m$ generative distributions, each controlling the configuration (e.g., position and size) of one obstacle. We consider generative distributions as truncated Gaussian distributions -- see Appendix \ref{sec:NNs}, where the dimension, lower and upper bounds depend on specific environment settings. For example in Fig. \ref{subfig:environmentside}, each obstacle is a shelf in the warehouse. The obstacle position on the $x$-axis is fixed and that on the $y$-axis can be adjusted in $[-4, 4]$ by sliding along a linear track. The generative distribution is a $1$-D truncated Gaussian distribution with a lower bound $-4$ and an upper bound $4$. We parameterize the generative model with a $3$-layer DNN of $(32, 64, 32)$ units and 
ReLU nonlinearity. 

\begin{figure*}
	\centering
	\begin{subfigure}{0.575\columnwidth}
		\includegraphics[width=1\linewidth,height = 0.75\linewidth]{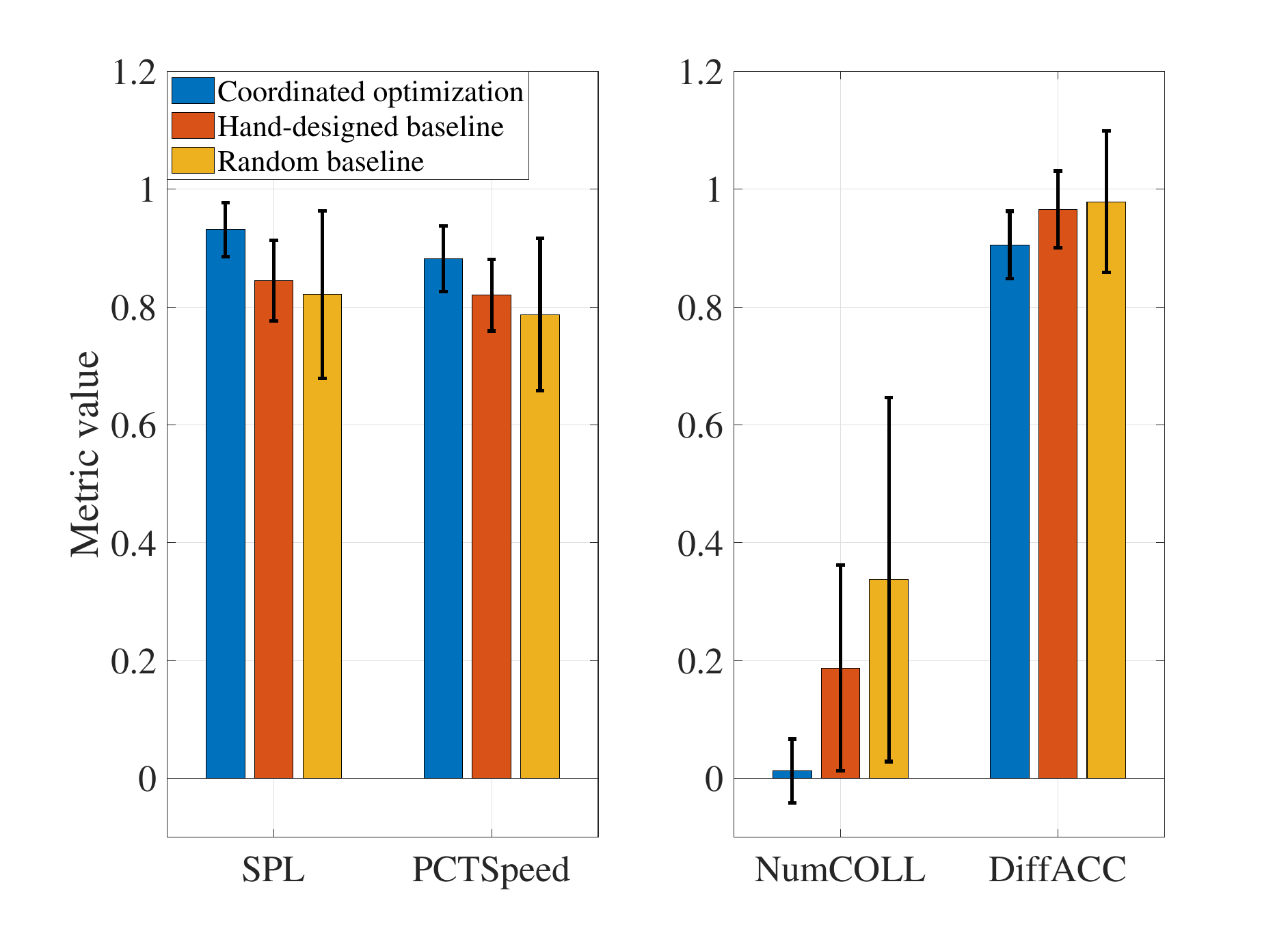}
		\caption{}%
		\label{subfig:Performance4Obstacles}%
	\end{subfigure}\hfill\hfill%
	\begin{subfigure}{0.575\columnwidth}
		\includegraphics[width=1\linewidth, height = 0.75\linewidth]{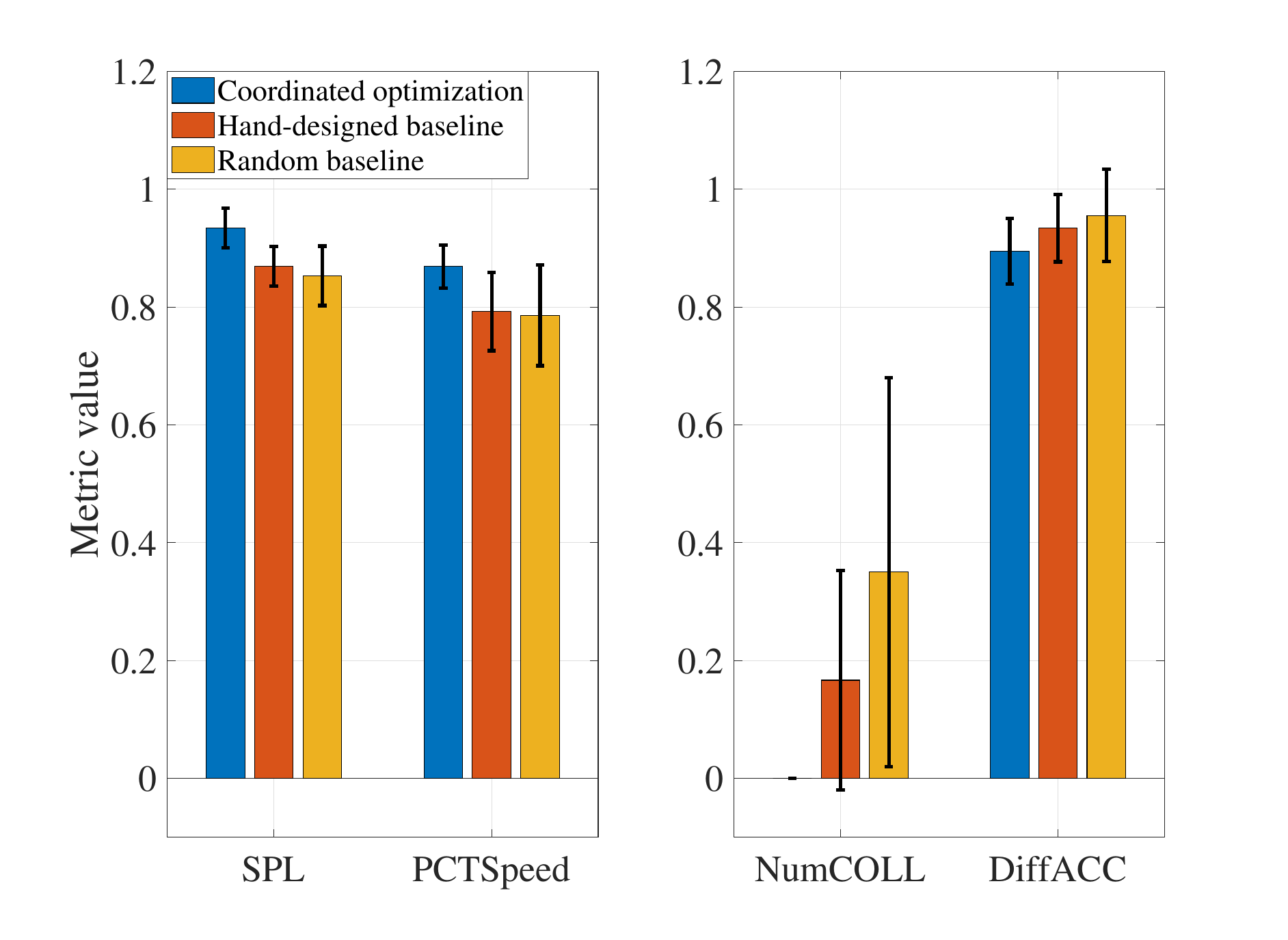}
		\caption{}%
		\label{subfig:Performance8Obstacles}
	\end{subfigure}\hfill\hfill%
	\begin{subfigure}{0.575\columnwidth}
		\includegraphics[width=1\linewidth,height = 0.75\linewidth]{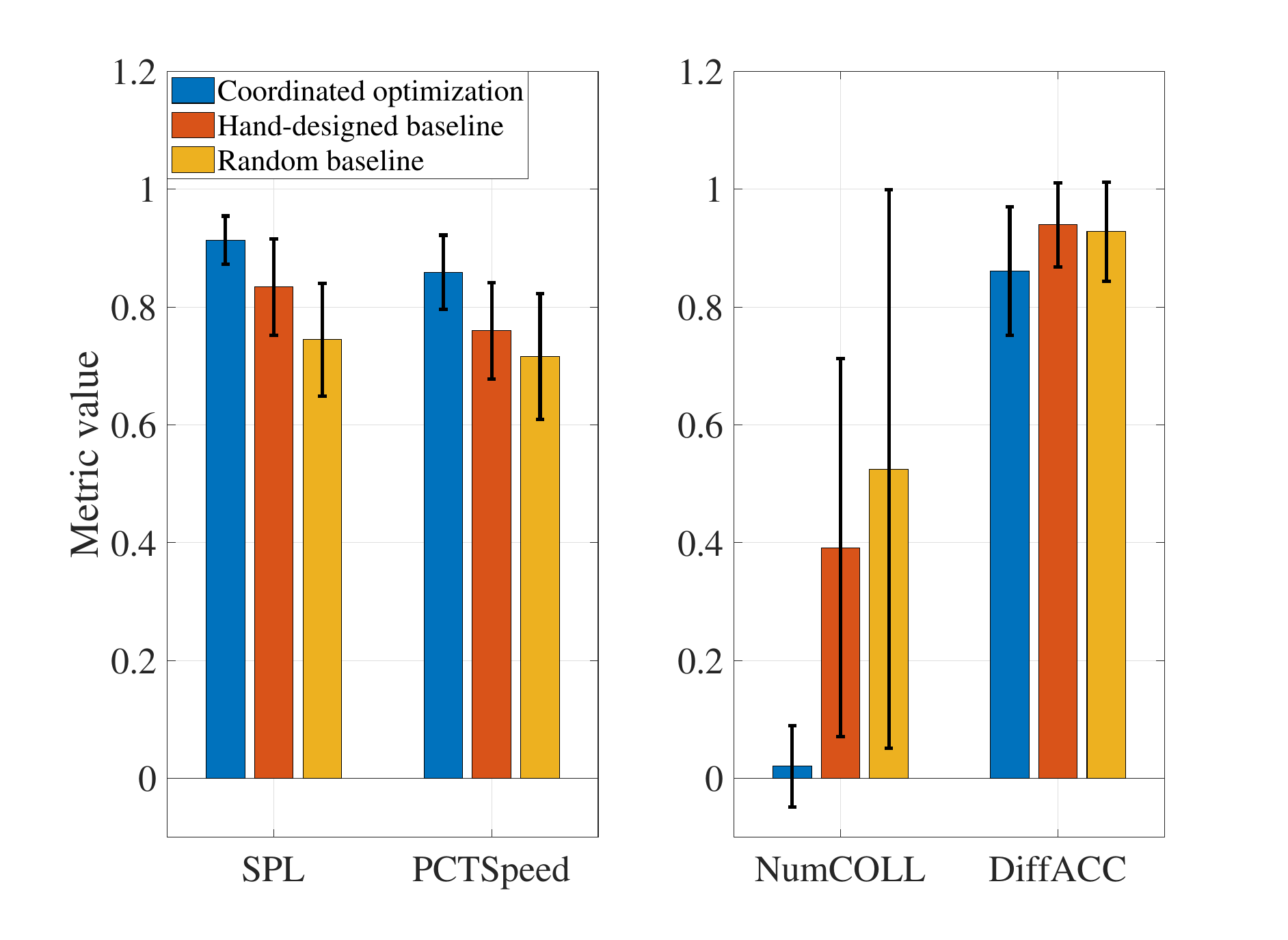}
		\caption{}%
		\label{subfig:Performance16Obstacles}%
	\end{subfigure}
	\caption{Performance with expectation and standard deviation of agent-environment coordinated optimization method compared with two baselines in three environments. A higher value of SPL or PCTSpeed represents a better performance, while a lower value of NumCOLL or DiffACC represents better safety or more comfort. (a) Environment I. (b) Environment II. (c) Environment III.}\label{fig:step2-1}\vspace{-5mm}
\end{figure*}

\begin{figure*}
	\centering
	\begin{subfigure}{0.475\columnwidth}
		\includegraphics[width=0.975\linewidth,height = 0.9\linewidth]{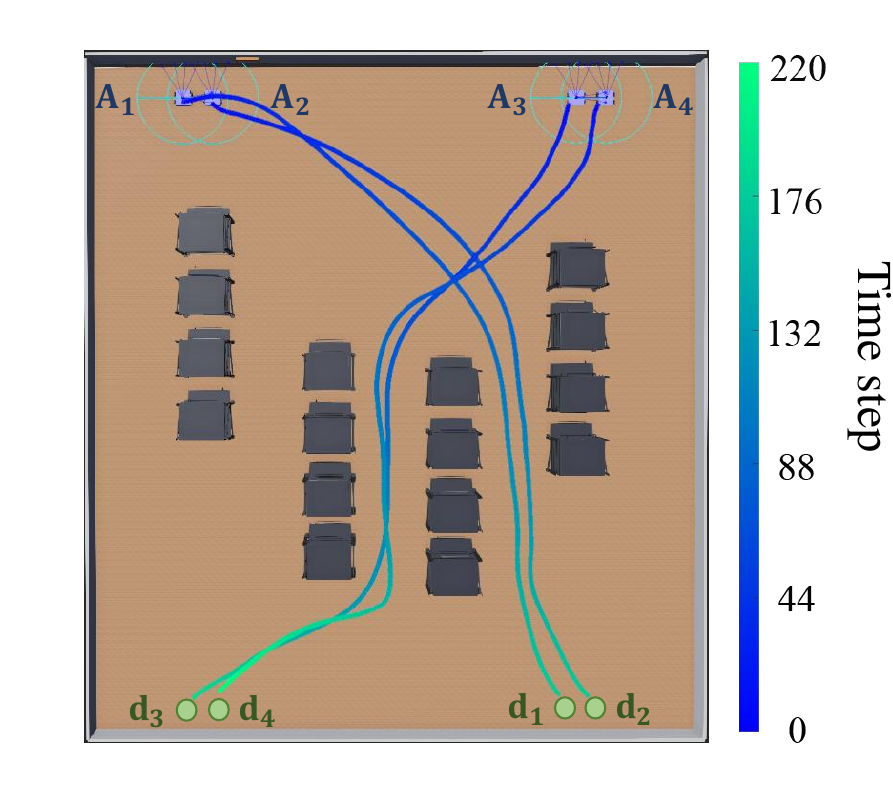}
		\caption{Optimized Environment I}%
		\label{subfig:exampleEnvironmentI}%
	\end{subfigure}\hfill\hfill%
	\begin{subfigure}{0.475\columnwidth}
		\includegraphics[width=0.975\linewidth, height = 0.9\linewidth]{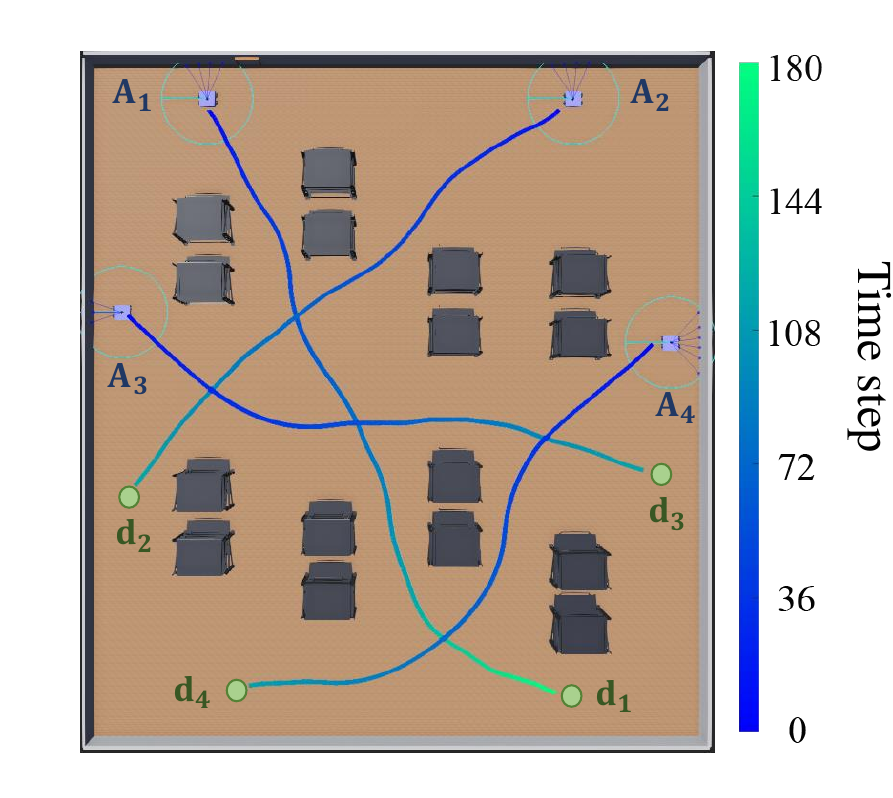}
		\caption{Optimized Environment II}%
		\label{subfig:distribution1}
	\end{subfigure}\hfill\hfill%
	\begin{subfigure}{0.475\columnwidth}
		\includegraphics[width=0.975\linewidth,height = 0.9\linewidth]{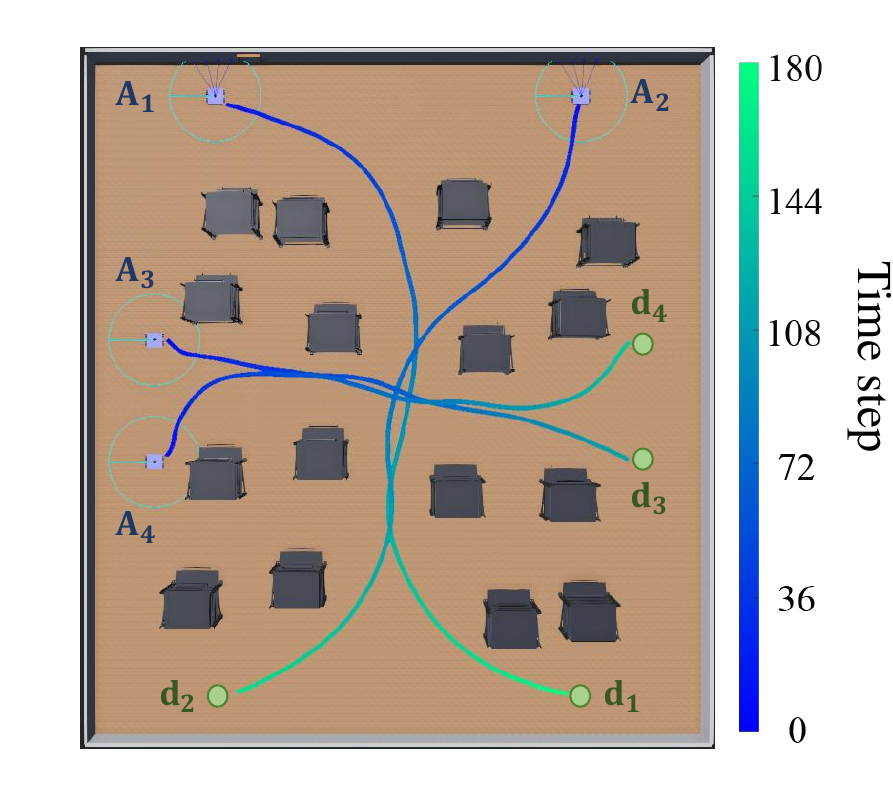}
		\caption{Optimized Environment III}%
		\label{subfig:step1trajectories}%
	\end{subfigure}
	\caption{Examples of agent-environment co-optimization in three environments. Blue robots $\{A_i\}_{i=1}^4$ represent agents at initial positions, green circles $\{\bbd_i\}_{i=1}^4$ represent goal positions, and black boxes represent obstacles (shelves). Colored lines from blue to green are agent trajectories and the color bar is the time scale. (a) Optimized Environment I. (b) Optimized Environment II. (c) Optimized Environment III.}\label{fig:step2-3}\vspace{-6mm}
\end{figure*}

We employ Webots as our 3D simulator, which contains realistic physical properties to simulate real-world conditions. Specifically, we design robots with joints, motors, communication modules and omnidirectional wheels to consider realistic physical constraints. We follow \cite{WebotsContact} to design materials of the floor and wheels, and tune the Coulomb friction coefficient to define contact properties for wheel movements. Moreover, we configure motor constraints, such as maximum angular velocity and torque for each wheel, to mimic real-world motor behaviors. All objects in Webots, including robot bodies, wheels, sensors and obstacles, have inertia, and the collision detection is based on the mesh geometry imposed on entities within the simulator. Each robot has a receiver and an emitter to broadcast and receive the state from its neighboring robots via communication. The low-level control is handled by model predictive control, which converts the navigation policy output to each wheel's speed using inverse kinematics. Therefore, our simulations account for realistic physical constraints instead of considering robots as simple point masses. 

\subsection{Proof of Concept}

First, we conduct a proof of concept with $4$ agents and $4$ obstacles. We consider an environment of size $10 \times 12$ in the warehouse setting as depicted in Fig. \ref{fig:step1_environment}. The agents are robots of size $0.4$, and the obstacles are rectangular shelves of size $1 \times 4$. The obstacle positions on the $x$-axis are fixed, yet are configurable on the $y$-axis in the range of $[-4, 4]$ to improve the navigation performance. 

Fig. \ref{subfig:Proofconvergence} shows that the objective value increases with the number of iterations and approaches a stationary condition, which corroborates the convergence analysis in Theorem \ref{thm:Convergence}. Fig. \ref{subfig:Proofdistribution} plots the generative distributions for four obstacles, respectively, which demonstrate how the generative model places the obstacles in the environment. First, the standard deviation of the distribution is small, i.e., the probability density concentrates in a small region. This indicates that there is little variation in the environment generation (hence, the resulting performance). Second, the generative model produces an irregular obstacle layout that differs from the hand-designed one, which implies that structure irregularity is capable of providing navigation help for the multi-agent system. Fig. \ref{subfig:Prooftrajectories} shows the agent trajectories controlled by our navigation policy in the generated environment, where agents move smoothly from initial positions to their goals without collision.

\subsection{Performance Evaluation}\label{subsec:performance}

We further evaluate our approach in three different cluttered environments that resemble a warehouse setting: 

\smallskip
\noindent \textbf{Environment I} is equipped with $4$ obstacles of size $1 \times 4$ -- see Fig. \ref{subfig:environment1}. The obstacle positions on the $x$-axis are fixed, and those on the $y$-axis can be re-configured in the range $[-4, 4]$.  
 
\noindent \textbf{Environment II} is equipped with $8$ obstacles of size $1 \times 2$ -- see Fig. \ref{subfig:environment2}. The obstacle positions on the $x$-axis are fixed, and those on the $y$-axis can be re-configured in $[-4, 0]$ or $[0, 4]$. 
 
\noindent \textbf{Environment III} is equipped with $16$ obstacles of size $1$$\times$$1$ -- see Fig. \ref{subfig:environment3}. The obstacle positions can be re-configured along both $x$- and $y$-axes in a neighborhood of size $2 \times 2$. 

\noindent We systematically consider Environments I-III with different degrees of flexibility w.r.t. obstacle positions, corresponding to different practical constraints. For example, Environment I re-arranges obstacles along the y-axis assuming there exist sliding tracks on the y-axis, while Environment III re-arranges obstacles along both x- and y-axes assuming they are flexible in any direction.

\begin{figure*}
	\centering
	\begin{subfigure}{0.55\columnwidth}
		\includegraphics[width=1\linewidth,height = 0.75\linewidth]{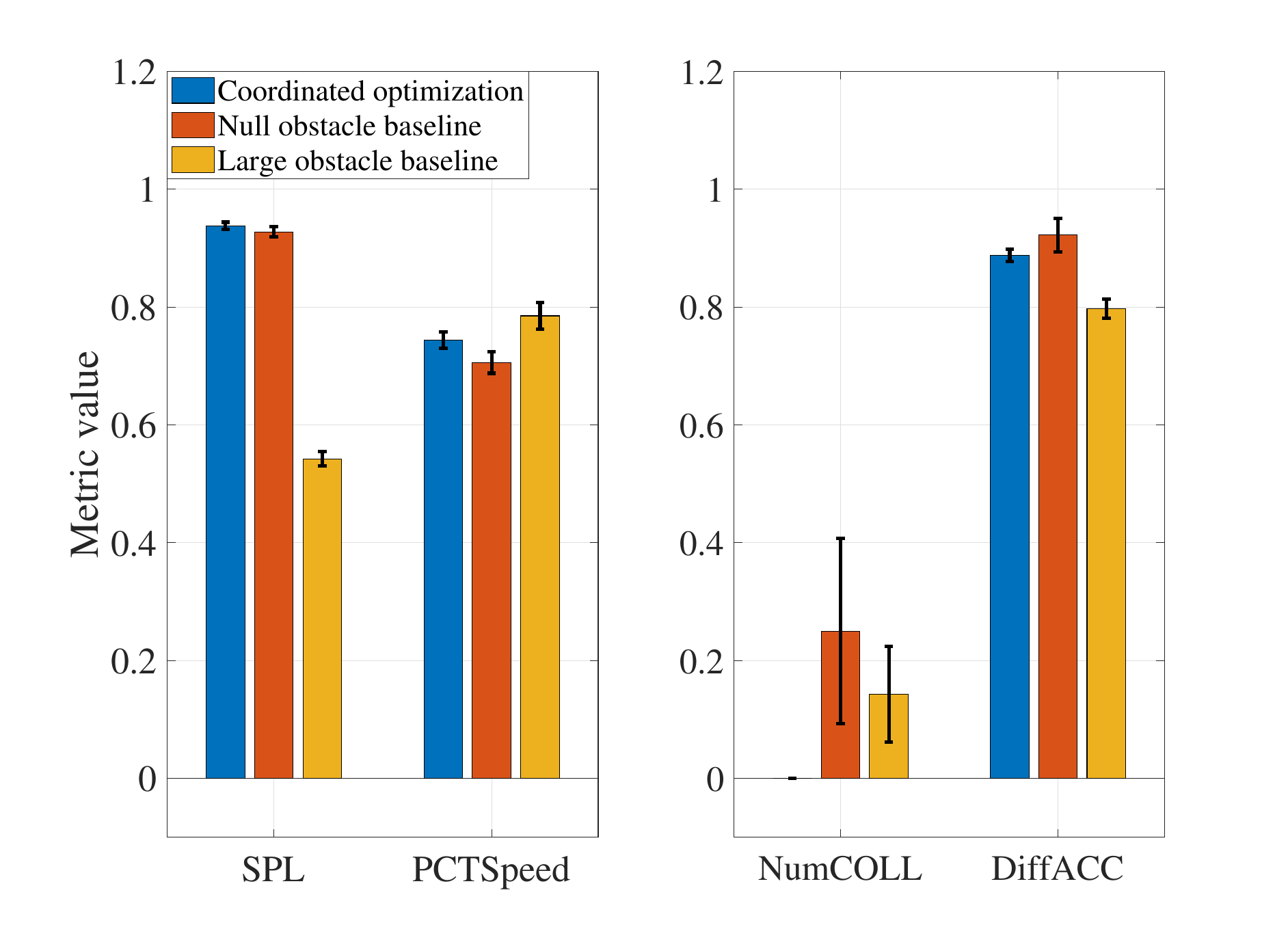}
		\caption{}
		\label{subfig:performance8}
	\end{subfigure}\hfill\hfill%
	\begin{subfigure}{0.55\columnwidth}
		\includegraphics[width=1\linewidth, height = 0.75\linewidth]{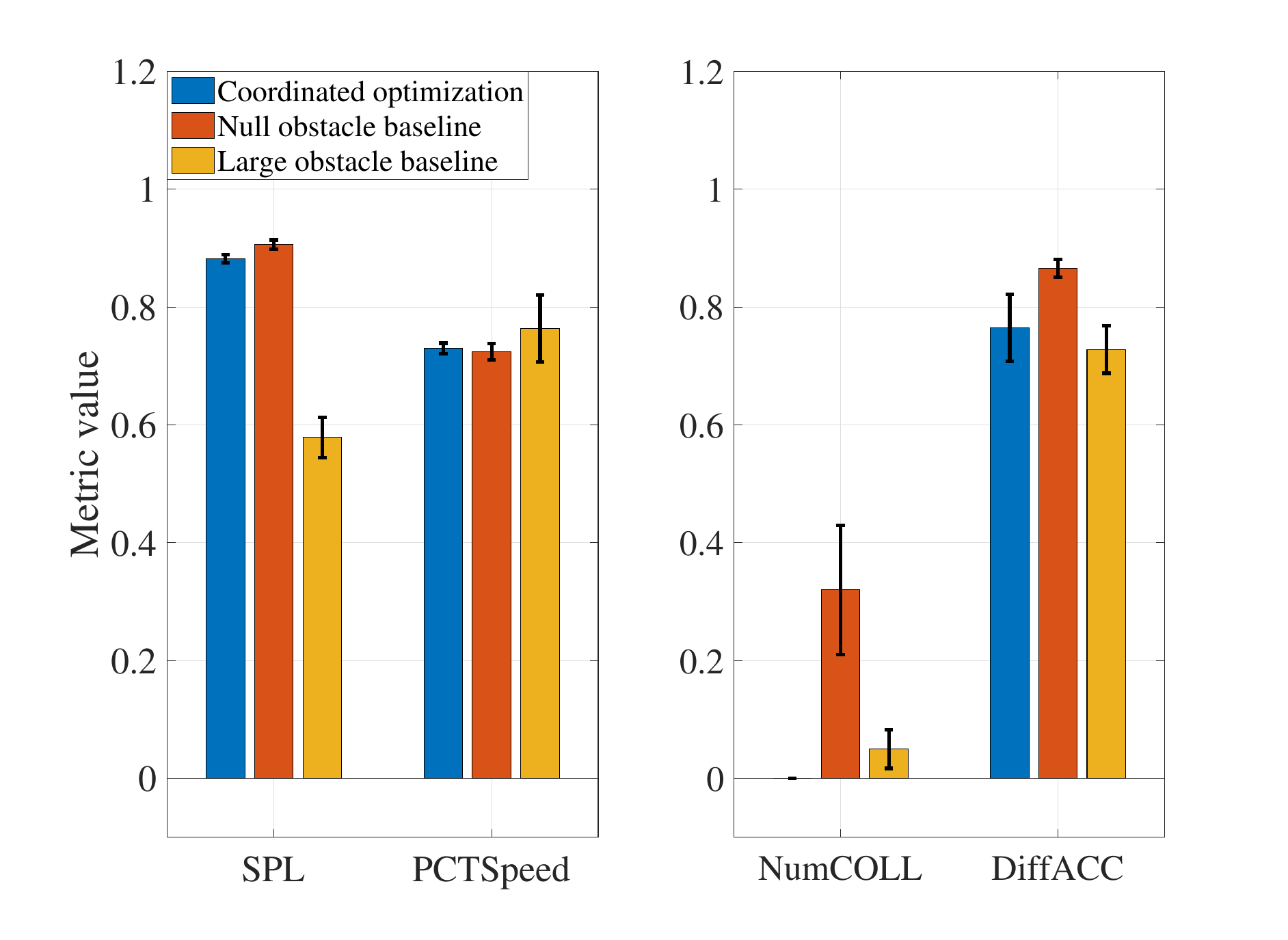}
		\caption{}
		\label{subfig:performance12}
	\end{subfigure}\hfill\hfill
	\begin{subfigure}{0.55\columnwidth}
		\includegraphics[width=1\linewidth,height = 0.75\linewidth]{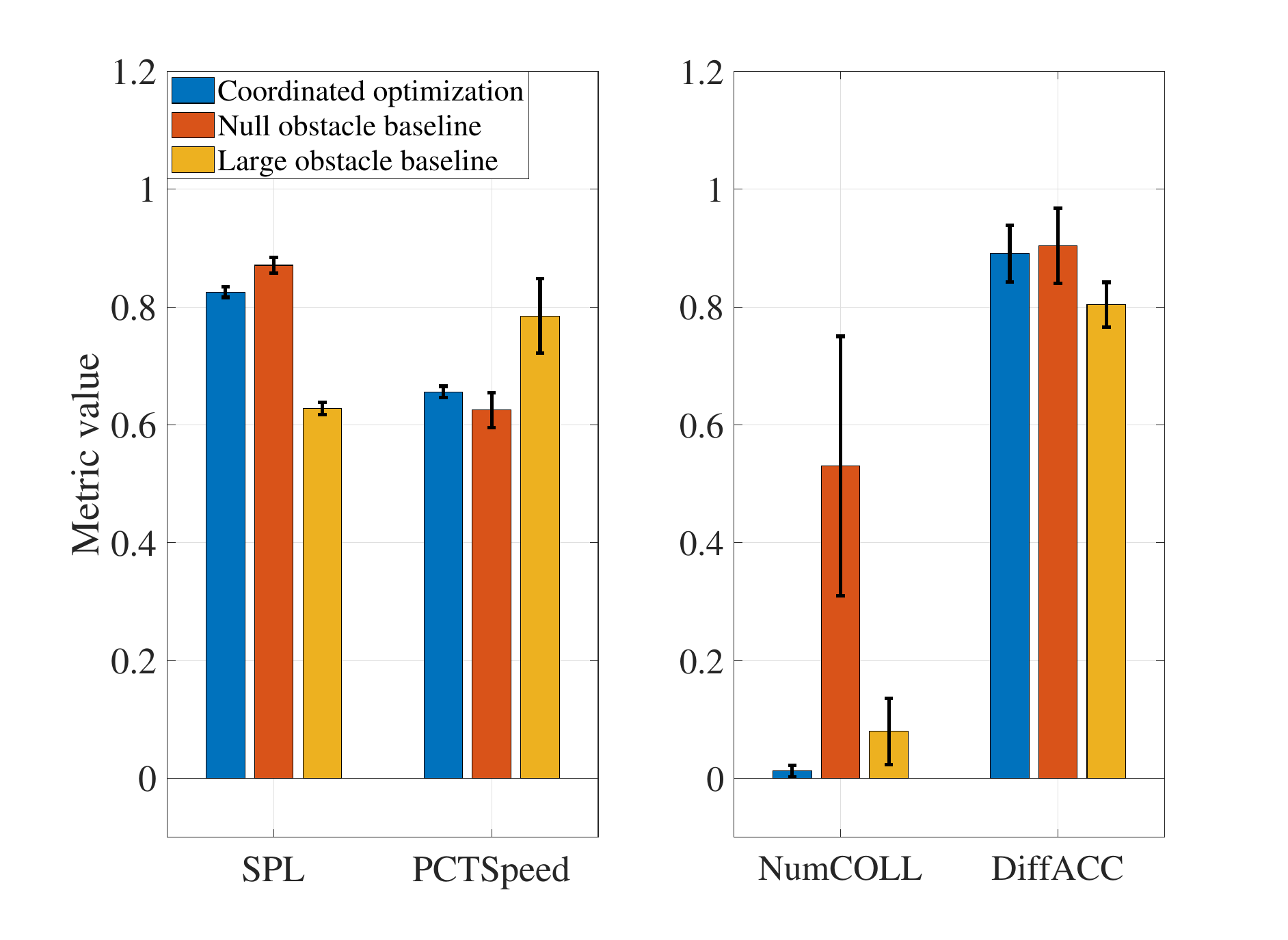}
		\caption{}
		\label{subfig:performance16}
	\end{subfigure}
	\caption{Performance of agent-environment coordinated optimization compared with two baseline scenarios in the circular setting with different numbers of agents. A higher value of SPL or PCTSpeed represents a better performance, while a lower value of NumCOLL or DiffACC represents better safety or more comfort. (a) $n=8$ agents. (b) $n=12$ agents. (c) $n=16$ agents.}\label{fig:step3-1}\vspace{-5mm}
\end{figure*}

\begin{figure*}
	\centering
	\begin{subfigure}{0.45\columnwidth}
		\includegraphics[width=0.975\linewidth,height = 0.85\linewidth]{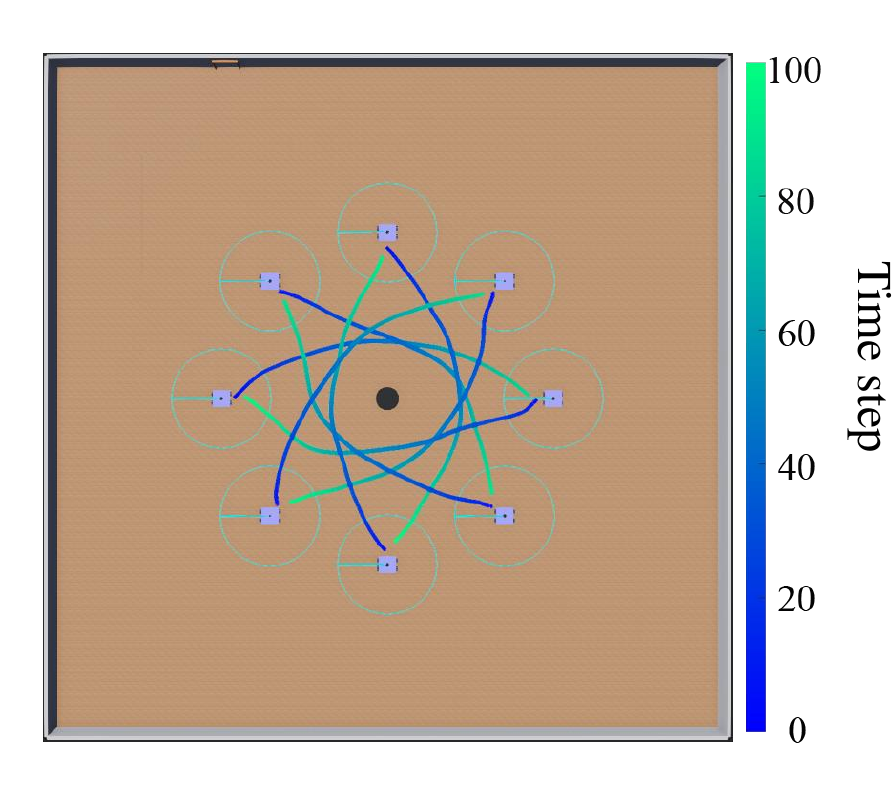}
		\caption{Environment with $8$ agents}
		\label{subfig:examplecircular8}
	\end{subfigure}\hfill\hfill
	\begin{subfigure}{0.45\columnwidth}
		\includegraphics[width=0.975\linewidth, height = 0.85\linewidth]{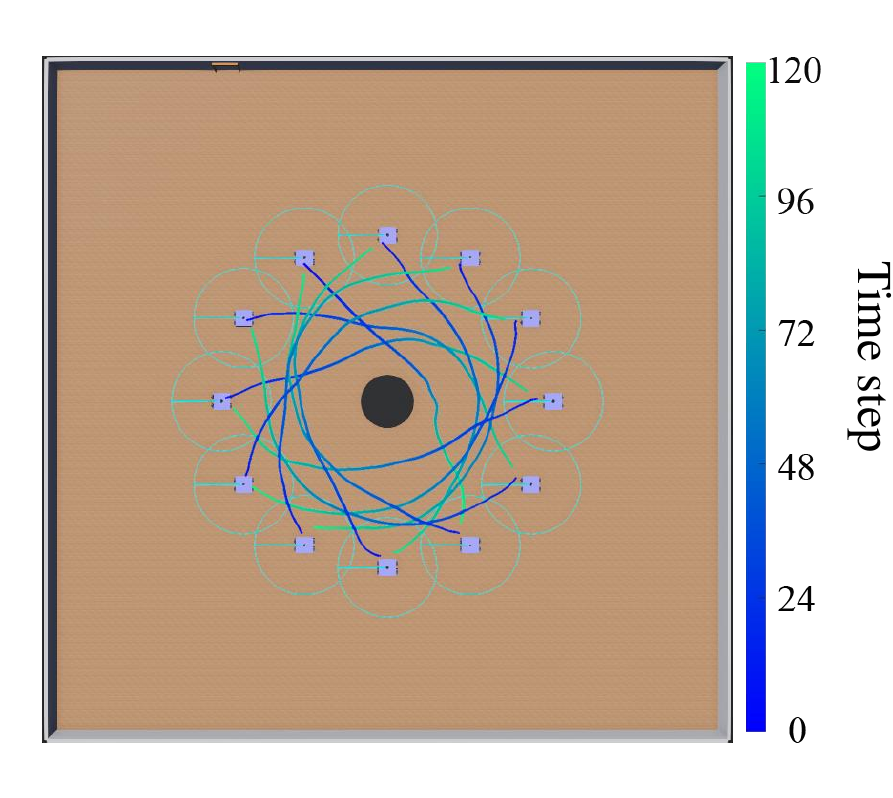}
		\caption{Environment with $12$ agents}
		\label{subfig:examplecircular12}
	\end{subfigure}\hfill\hfill
	\begin{subfigure}{0.45\columnwidth}
		\includegraphics[width=0.975\linewidth,height = 0.85\linewidth]{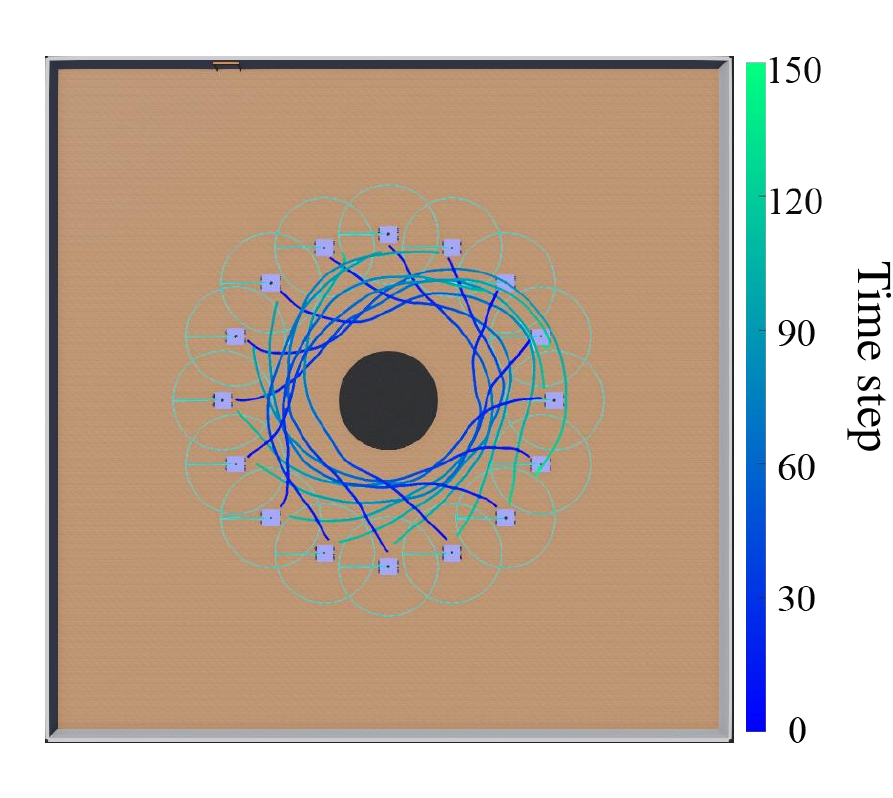}
		\caption{Environment with $16$ agents}
		\label{subfig:examplecircular16}
	\end{subfigure}
	\caption{Examples of agent-environment co-optimization in the circular setting with different numbers of agents. Blue robots $\{A_i\}_{i=1}^n$ represent agents at initial positions, green circles $\{\bbd_i\}_{i=1}^n$ represent goals, and the solid black circle represents the obstacle. Colored lines from blue to green are agent trajectories and the color bar represents the time scale. (a) $n=8$ agents. (b) $n=12$ agents. (c) $n=16$ agents.}\label{fig:step3-2}\vspace{-6mm}
\end{figure*}

We measure the navigation performance with four metrics: \textit{(i)} Success weighted by Path Length (SPL) \cite{anderson2018evaluation}, \textit{(ii)} the percentage to the maximal speed (PCTSpeed), \textit{(iii)} the average number of ``collisions'' (NumCOLL), and \textit{(iv)} the average finite difference of the acceleration (DiffACC). SPL is a stringent measure that combines success rate with path length overhead, and has been widely used as a primary quantifier in comparing navigation performance \cite{anderson2018evaluation, gervet2023navigating, wang2019reinforced}. PCTSpeed represents the ratio of the average speed to the maximal one, and provides complementary information regarding the moving speed along trajectories. These two metrics are normalized to $[0,1]$ with a higher value representing a better performance, providing a unified exposition. NumCOLL counts the number of `collisions' averaged over agents, where an agent is considered `in collision' at time $t$ if it is within the safety range of another agent or obstacle, and measures the safety of the multi-agent system. DiffACC represents the change of agents' acceleration averaged over time steps, and indicates the comfort (smoothness) of the navigation procedure \cite{bellem2016objective, eboli2016measuring}. For the latter two metrics, a lower value represents a better performance, i.e., better safety and more comfort -- see Appendix \ref{appendix:metrics}. Our results are averaged over $30$ random tasks with $10$ random trials for a total of $300$ runs. The initial and goal positions of each task are randomly sampled in the environment boundaries, while maintaining a minimum distance between the starting and goal positions of each agent for a non-trivial navigation task. 

Fig. \ref{fig:step2-1} shows the performance with expectation and standard deviation of the coordinated optimization method. We consider two baselines: a hand-designed environment as shown in Fig. \ref{fig:step21} and the randomly generated environment. The hand-designed baseline is a standard environment with a regular obstacle layout, as is common practice. Our method outperforms the baselines consistently with the highest SPL and PCTSpeed and the lowest NumCOLL and DiffACC, and maintains a good performance across different environments. We attribute this behavior to that the generative model adapts the obstacle layout to alleviate environment restrictions, based on which the navigation policy tunes the agent trajectories to improve the performance. The performance improvement is emphasized as the environment structure becomes more complex from Environment I with $4$ obstacles to Environment III with $16$ obstacles. This is because a greater number of obstacles leads to increasingly challenging navigation scenarios, in which agents derive more benefits from environments that are more compatible with their navigation policies. 

Fig. \ref{fig:step2-3} displays examples of the agent-environment co-optimization in Environments I-III. We see that: \textit{(i)} The generative model produces different obstacle layouts depending on different navigation tasks, i.e., different starting and goal positions of agents. The resulting environment exhibits an irregular structure different from the hand-designed one, which facilitates the multi-agent navigation. \textit{(ii)} The generated environment not only creates obstacle-free pathways for agents, but also prioritizes these pathways through structure irregularity and non-symmetry to implicitly de-conflict agents via prioritization. For example in Fig. \ref{subfig:step1trajectories}, while the navigation tasks of $A_1$ and $A_2$ are symmetric, the generative model produces an irregular obstacle layout that creates non-symmetric obstacle-free pathways for $A_1$ and $A_2$. This prioritizes the two agents and avoids their potential confliction by allowing $A_2$ to pass the passage before $A_1$. \textit{(iii)} The navigation policy is compatible with the generative model and selects the optimal trajectories in the generated environment, moving the agents smoothly towards goals. \textit{(iv)} The navigation policy strives to move agents as a team (or sub-team) to remain connected, because the agents rely on their neighborhood information to make collectively meaningful decisions. 

\begin{figure*}
	\centering
	\begin{subfigure}{0.475\columnwidth}
		\includegraphics[width=0.975\linewidth,height = 0.9\linewidth]{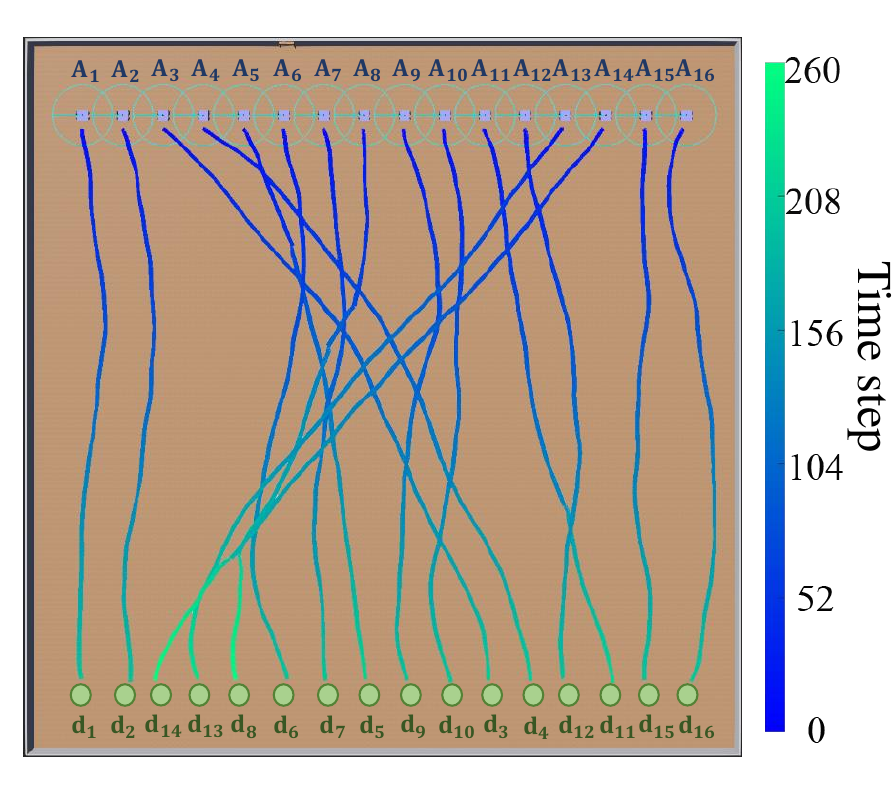}
		\caption{}
		\label{subfig:examplerandom2ObstacleNull}
	\end{subfigure}\hfill\hfill
	\begin{subfigure}{0.475\columnwidth}
		\includegraphics[width=0.975\linewidth, height = 0.9\linewidth]{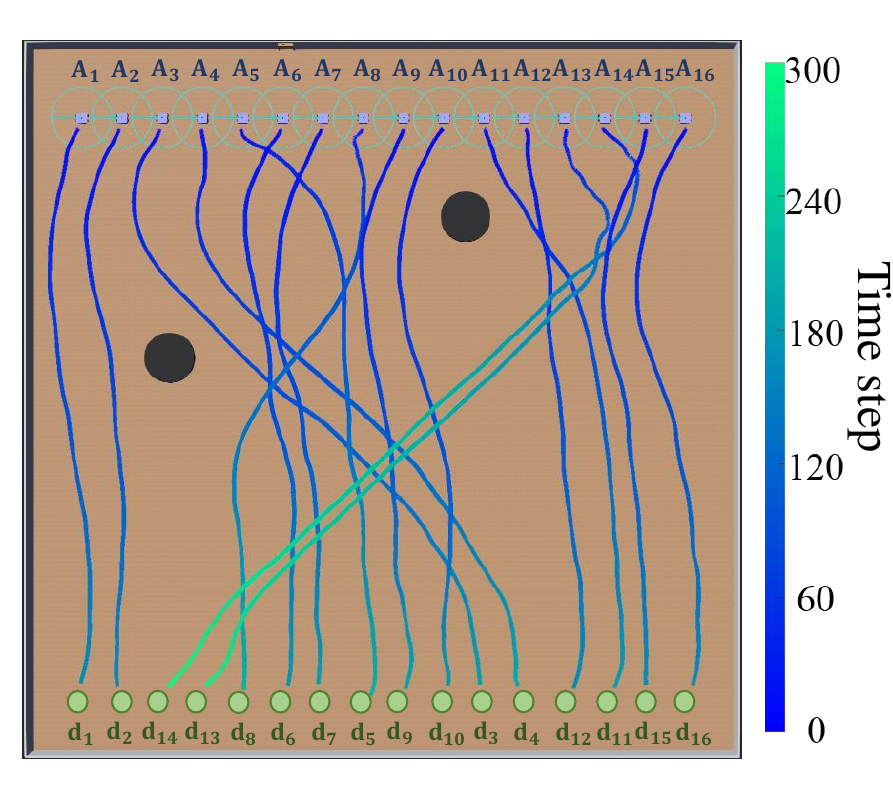}
		\caption{}
		\label{subfig:examplerandom2ObstacleSmall}
	\end{subfigure}\hfill\hfill
	\begin{subfigure}{0.475\columnwidth}
		\includegraphics[width=0.975\linewidth,height = 0.9\linewidth]{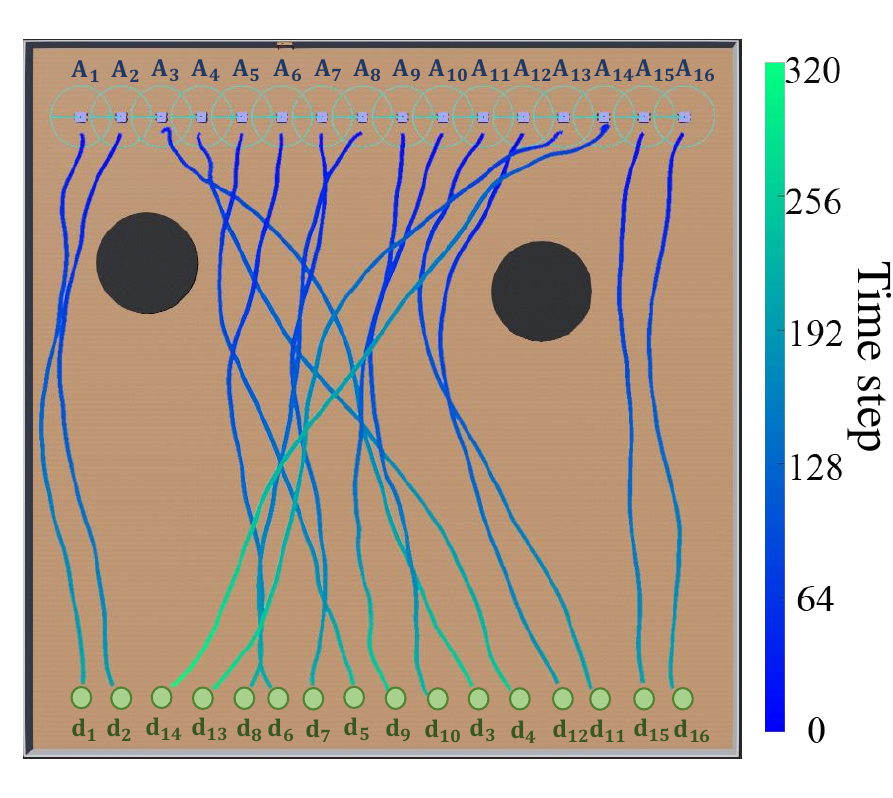}
		\caption{}
		\label{subfig:examplerandom2ObstacleLarge}
	\end{subfigure}
	\caption{Examples of agent-environment co-optimization in Case I of the randomized setting with different obstacle radii. Blue robots $\{A_i\}_{i=1}^{16}$ represent agents at initial positions, green circles $\{\bbd_i\}_{i=1}^{16}$ represent goals, and black circles represent obstacles. Colored lines from blue to green are agent trajectories and the color bar represents the time scale. (a) Null radius $r=0$. (b) Radius $r=0.5$. (c) Radius $r=1$.}\label{fig:step4-2}\vspace{-6.5mm}
\end{figure*}

\begin{figure}
	\centering
	\begin{subfigure}{0.475\columnwidth}
		\includegraphics[width=1.05\linewidth,height = 0.8\linewidth]{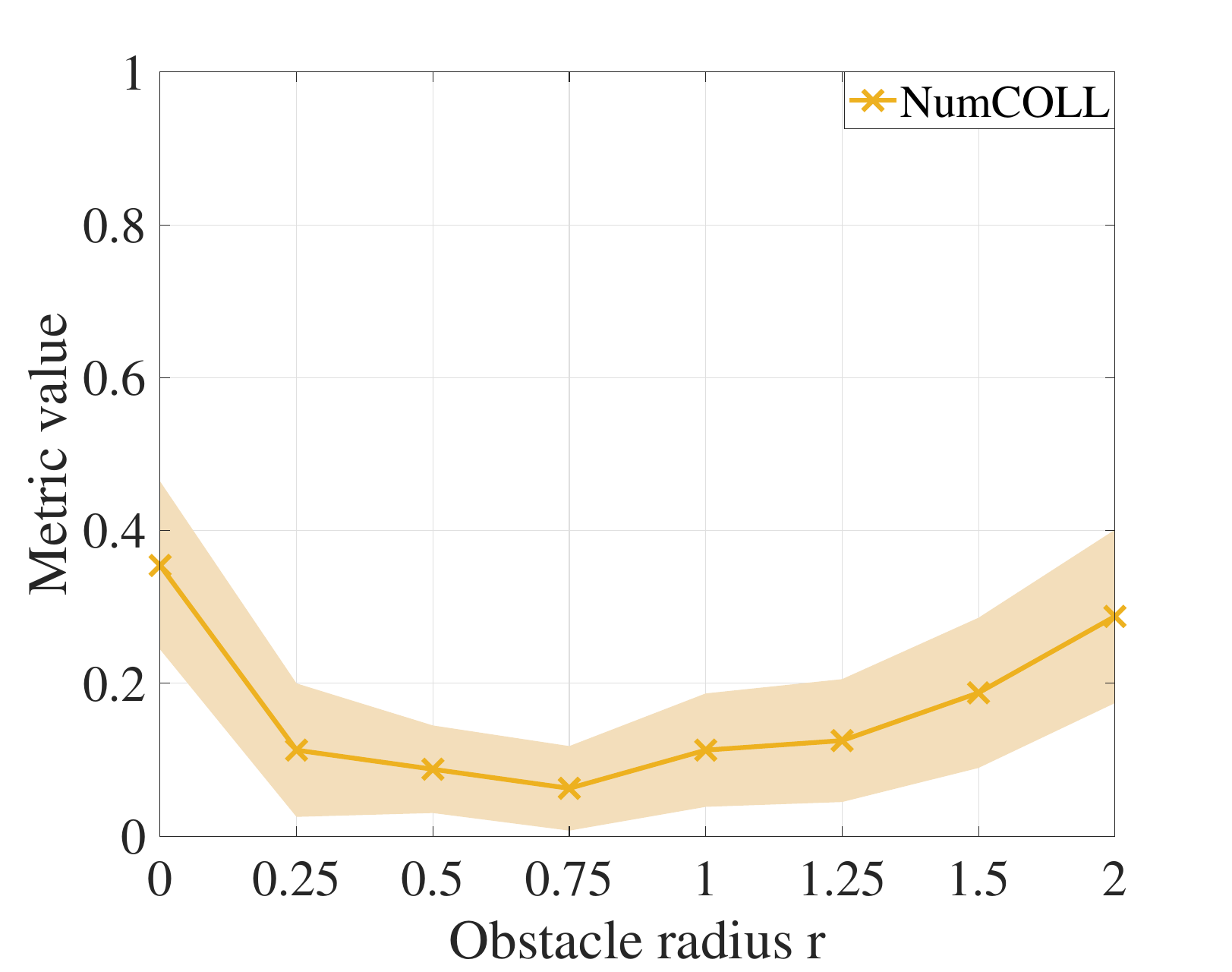}
		\caption{}
		\label{subfig:performance2Obstacles}
	\end{subfigure}\hfill\hfill
	\begin{subfigure}{0.475\columnwidth}
		\includegraphics[width=1.07\linewidth, height = 0.8\linewidth]{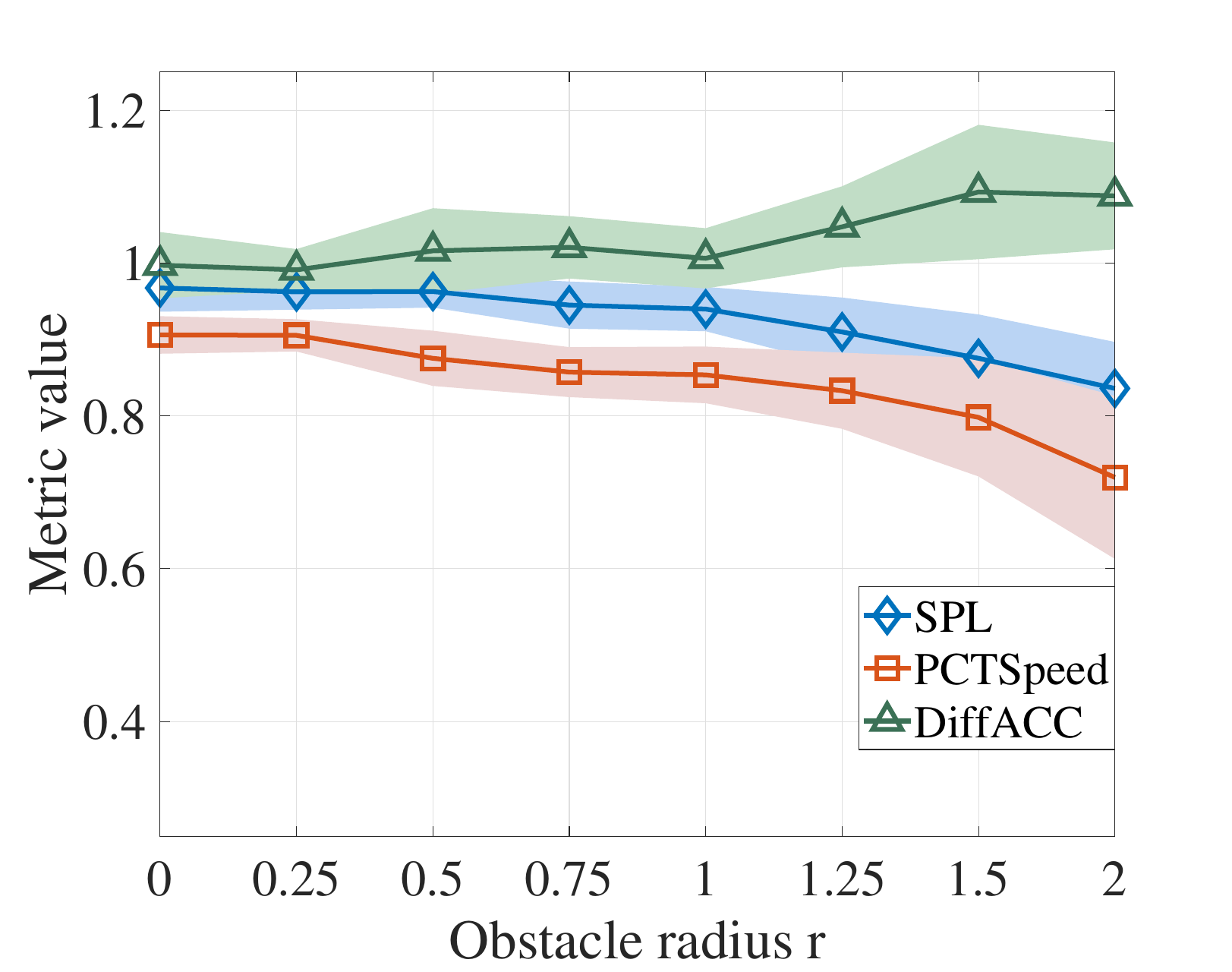}
		\caption{}
		\label{subfig:performance2Obstacles2}
	\end{subfigure}
	\caption{Performance of agent-environment co-optimization in Case I of the randomized setting with different obstacle radii. A higher SPL / PCTSpeed represents a better performance, while a lower NumCOLL / DiffACC represents better safety or more comfort.}\label{fig:step4-1}\vspace{-6mm}
\end{figure}

These results indicate that our approach establishes a symbiotic agent-environment system, where the environment and the agents rely on and adapt to each other; ultimately, yielding an improved navigation performance. 

\subsection{Hindrance vs. Guidance}\label{subsec:blockingorGuidance}

Next, we investigate the role played by the environment in multi-agent navigation by leveraging our coordinated optimization method. Obstacles are typically viewed as ``negative'' elements that create inaccessible areas or obstruct agent pathways (and hinder ideal trajectories). However, we hypothesize that an appropriate obstacle layout can have ``positive'' effects by forming designated collision-free zones for the agents to follow. Such an arrangement can provide navigation guidance and help de-conflict agents, especially in dense settings. It is somewhat counterintuitive that it outperforms the empty space. The following experiments aim to verify this hypothesis and to capture the relationship between the environment and agents. That is, we aim to uncover an implicit trade-off between hindrance and guidance roles that the environment may take on. 

We consider two environment settings motivated by traffic system design for safe navigation. 

\noindent \textbf{Circular setting.} The environment is of size $8 \times 8$, where agents are initialized along a circle -- see Fig. \ref{fig:step3-2}. The goal of the navigation policy is to cross-navigate the agents towards the opposite side while avoiding collisions, and the goal of the generative model is to determine the radius of the circular obstacle at the origin $(0,0)$ to de-conflict the agents. A large obstacle reduces the reachable space for the agents and may degrade the navigation performance, while a small obstacle provides a negligible guiding effect and may not prevent agent conflicts (e.g., all agents attempt to move along the shortest but intersecting trajectories). We explore this hindrance-guidance trade-off by performing the agent-environment co-optimization to find the optimal obstacle radius. 

Fig. \ref{fig:step3-1} shows the performance of the coordinated optimization method with different numbers of agents $n \in \{8, 12, 16\}$. We consider two baseline scenarios: a null obstacle of radius $0$ and a large obstacle of radius $2$. The presence of the obstacle with an appropriate radius improves the agents' performance with a significantly lower NumCOLL. This improvement scales with the number of agents, which can be explained by that a more cluttered system with a higher density of agents necessitates more guidance from the obstacle to facilitate agent de-confliction. The baseline scenario with a null obstacle, i.e., the empty space, achieves a comparable SPL / PCTSpeed but increases NumCOLL, despite no hindrance along agent trajectories. This is because the agents have to coordinate their navigation trajectories with only neighbors' states, and receive no guidance from the obstacle. Its NumCOLL increases from the scenario with $8$ agents to the one with $16$ agents. We attribute this behavior to that while inter-agent interactions may be sufficient for de-confliction in a sparse environment with a small number of agents, it is challenging for agents to de-conflict each other in a cluttered environment with a large number of agents and the latter requires guidance from the obstacle. The baseline scenario with a large obstacle results in a lower SPL and a slightly higher NumCOLL because it reduces the reachable space and imposes more restrictions on the feasible solution. The latter blocks the shortest pathways and forces the agents to move along inefficient trajectories. 

\begin{figure*}
	\centering
	\begin{subfigure}{0.475\columnwidth}
		\includegraphics[width=0.975\linewidth,height = 0.9\linewidth]{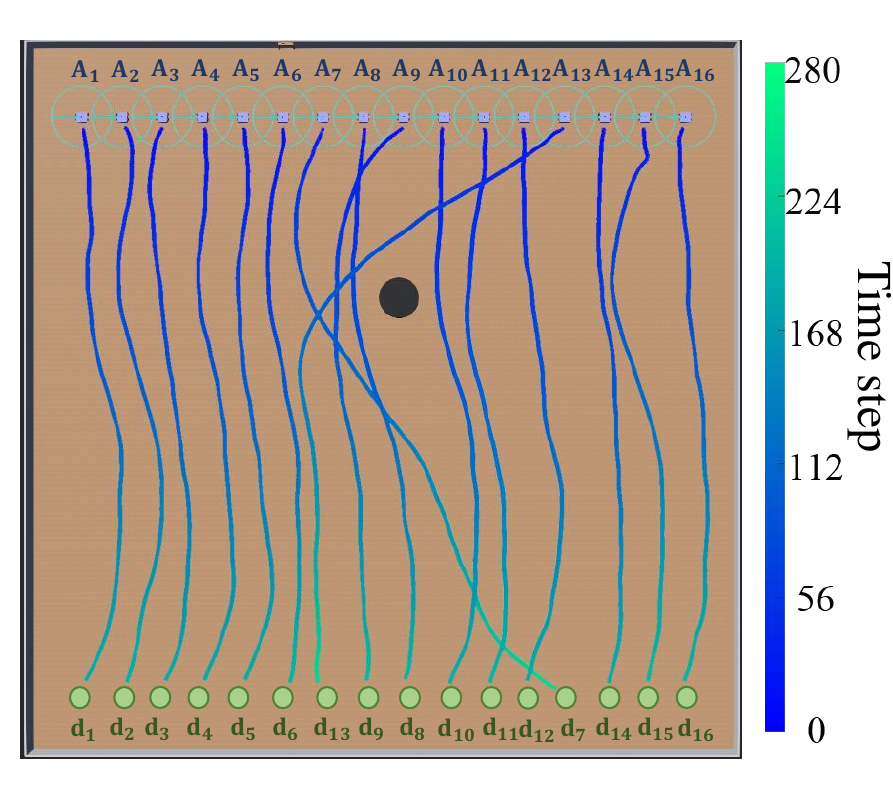}
		\caption{}
		\label{subfig:examplerandom4Randomness}
	\end{subfigure}\hfill\hfill
	\begin{subfigure}{0.475\columnwidth}
		\includegraphics[width=0.975\linewidth, height = 0.9\linewidth]{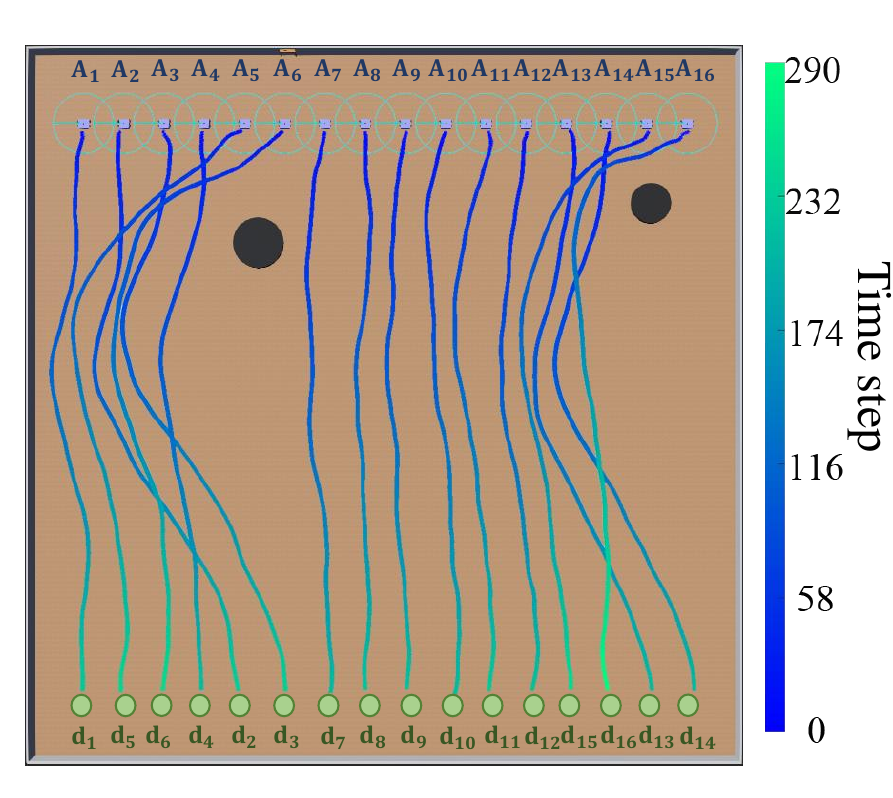}
		\caption{}
		\label{subfig:examplerandom8Randomness}
	\end{subfigure}\hfill\hfill
	\begin{subfigure}{0.475\columnwidth}
		\includegraphics[width=0.975\linewidth,height = 0.9\linewidth]{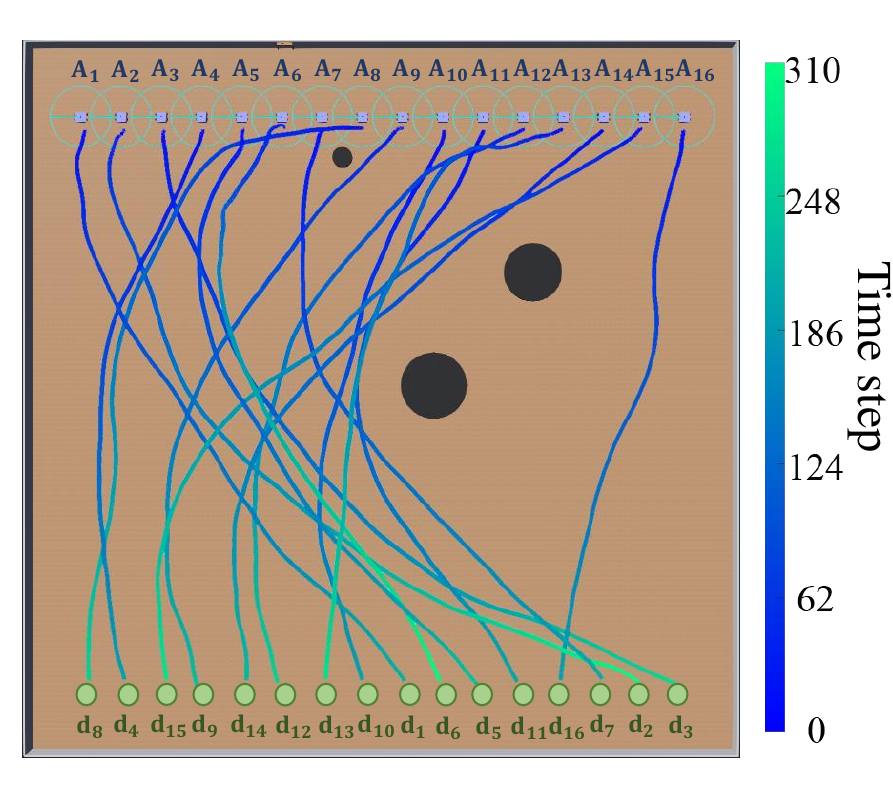}
		\caption{}
		\label{subfig:examplerandom16Randomness}
	\end{subfigure}
	\caption{Examples of agent-environment co-optimization in Case II of the randomized setting with different numbers of random agents $\eta$. Blue robots $\{A_i\}_{i=1}^{16}$ represent agents at initial positions, green circles $\{\bbd_i\}_{i=1}^{16}$ represent goals, and black circles represent obstacles. Colored lines from blue to green are agent trajectories and the color bar represents the time scale. (a) $\eta=4$. (b) $\eta=8$. (c) $\eta=16$.}\label{fig:step4-3}\vspace{-7mm}
\end{figure*}

\begin{figure}
	\centering
	\begin{subfigure}{0.45\columnwidth}
		\includegraphics[width=1.05\linewidth,height = 0.8\linewidth]{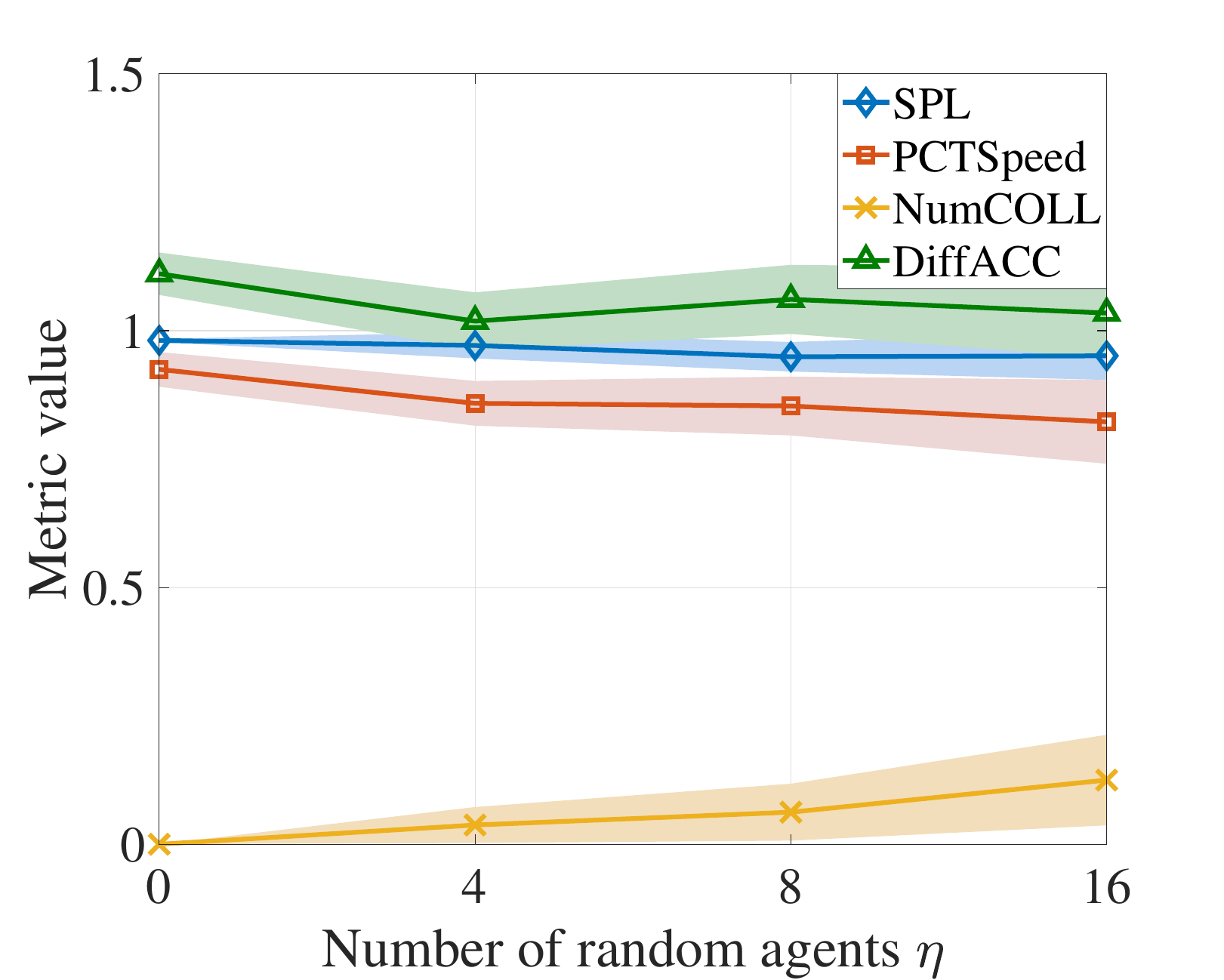}
		\caption{}
		\label{subfig:performanceRandomnessPerformance}
	\end{subfigure}
	\begin{subfigure}{0.45\columnwidth}
		\includegraphics[width=1.05\linewidth, height = 0.8\linewidth]{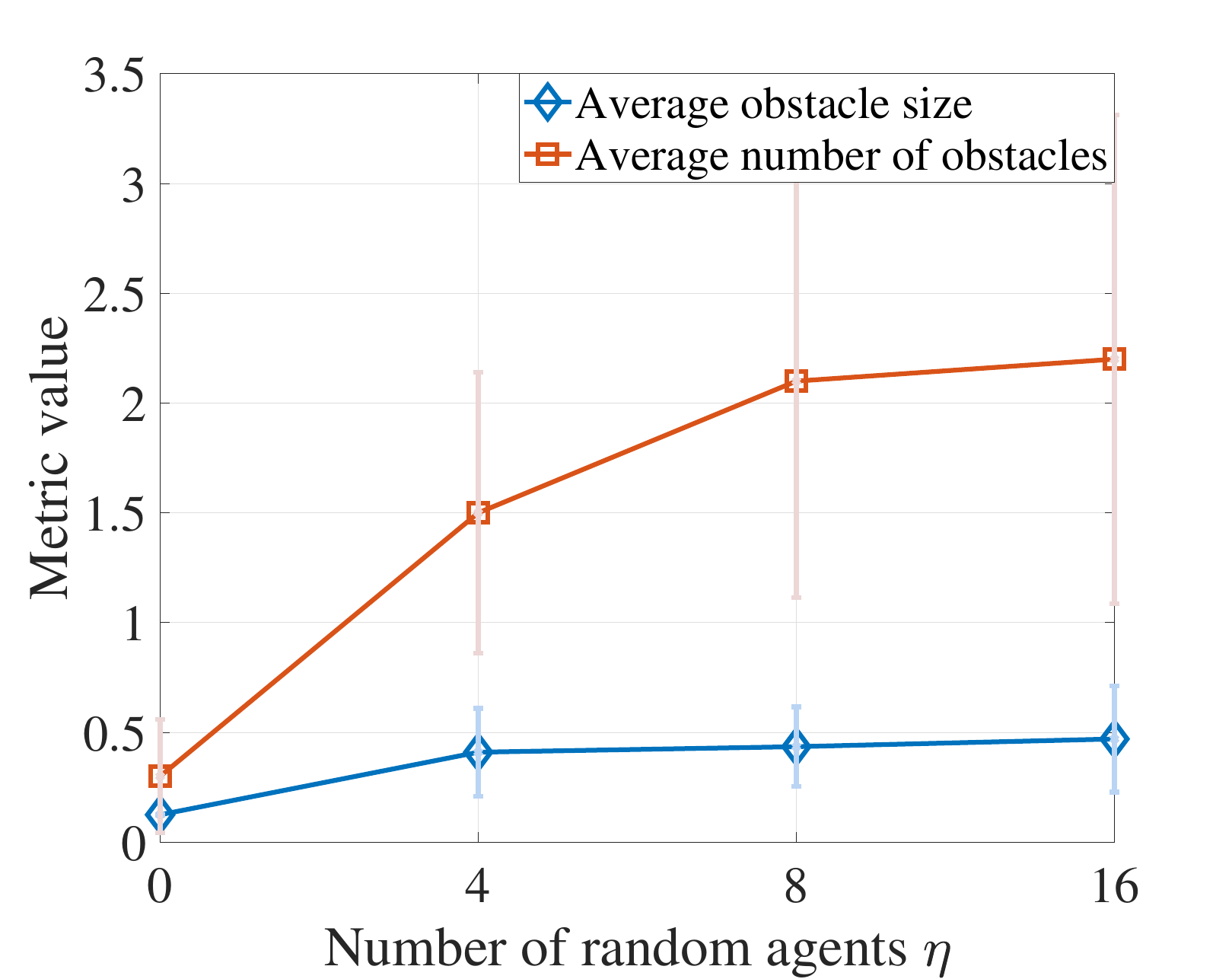}
		\caption{}
		\label{subfig:performanceRandomnessSize}
	\end{subfigure}
	\caption{(a) Performance of agent-environment coordinated optimization in Case II of the randomized setting with different numbers of random agents $\eta$. A higher value of SPL or PCTSpeed represents a better performance, while a lower value of NumCOLL or DiffACC represents better safety or more comfort. (b) Average obstacle radius and average number of obstacles determined by coordinated optimization method with different numbers of random agents $\eta$.}\label{fig:step4-4}\vspace{-7mm}
\end{figure}

Fig. \ref{fig:step3-2} plots the obstacle radius of the generative model and agent trajectories of the navigation policy. The obstacle radius generated by our method increases with the number of agents $n$. For a small $n$, the agents are sparsely distributed and need less guidance from the obstacle for de-confliction, leading to a small radius that does not hinder agent pathways. For a large $n$, the agents are densely distributed and require more help from the obstacle for coordination, leading to a large radius that guides the agents moving clockwise towards goals. 

\noindent \textbf{Randomized setting.} The environment is of size $18 \times 18$ with $16$ starting positions $\{\bbs_i\}_{i=1}^{16}$ on the top and $16$ goals $\{\bbd_i\}_{i=1}^{16}$ on the bottom -- see Fig. \ref{fig:step4-2}. There are $16$ agents $\{A_i\}_{i=1}^{16}$ initialized at $\{\bbs_i\}_{i=1}^{16}$ and tasked towards $\{\bbd_i\}_{i=1}^{16}$ in order, i.e., agent $A_i$ is initialized at $\bbs_i$ and tasked towards $\bbd_i$ for $i=1,...,16$. The goal of the navigation policy is to move the agents to their goals while avoiding collision among each other, and the goal of the generative model is to determine the obstacle layout between starting and goal positions to facilitate safe navigation. An environment with a large number of obstacles or a large obstacle size hinders agent pathways, while an empty environment might not provide enough guidance for effective agent de-confliction. By leveraging the agent-environment co-optimization, we explore this trade-off to find the optimal obstacle layout for different navigation cases. 

\noindent \textbf{Case I.} We consider an environment with $2$ circular obstacles, where the obstacle radius varies in $[0, 2]$. We randomly switch the initial and goal positions of $8$ agents, while keeping the other $8$ agents unchanged -- see Fig. \ref{subfig:examplerandom2ObstacleNull}. The generative model determines the optimal obstacle positions that de-conflict the agents while avoiding hindering their pathways, striking a balance between the guidance and the hindrance roles. 

Fig. \ref{subfig:performance2Obstacles} shows that NumCOLL first decreases with the obstacle radius $r$, and then increases for large $r$. The former is because an initial increase in $r$ allows the presence of obstacles in an empty environment, which provides the operation space for our method to de-conflict the agents with an appropriate obstacle layout. The result for large $r$ corresponds to the fact that further increasing $r$ occupies more collision-free space but provides little additional guidance, leading to an increased NumCOLL. Fig. \ref{subfig:performance2Obstacles2} shows that SPL, PCTSpeed decrease and DiffACC increases slightly with $r$, while all maintaining a good performance. This slight degradation is because \textit{(i)} increasing $r$ reduces the reachable space in the environment and \textit{(ii)} some agents make a detour under the obstacle guidance to avoid collision with the other agents. 

Fig. \ref{fig:step4-2} plots examples of how our method optimizes obstacle positions with different obstacle radii. We see that different obstacle positions and sizes have different impact on agents' trajectories, and the obstacles are typically placed around the intersections of agents' trajectories to separate the crowd and de-conflict the agents. For example, with no obstacle in Fig. \ref{subfig:examplerandom2ObstacleNull}, $A_{12}$, $A_{13}$, $A_{14}$ all attempt to move fast along short paths, which increases the risk of agent collisions. With the presence of obstacles in Fig. \ref{subfig:examplerandom2ObstacleLarge}, $A_{13}$, $A_{14}$ are guided to slow down and make a detour for $A_{12}$ to avoid potential collisions.

\noindent \textbf{Case II.} We consider an environment with at most $4$ circular obstacles. We randomly switch the initial and goal positions of $\eta$ agents, where $\eta \in [0, 16]$ represents the randomness of the navigation task, while keeping the rest $16-\eta$ agents unchanged -- see Fig. \ref{fig:step4-3}. The generative model determines: \textit{(i)} the number of obstacles presented in the environment; \textit{(ii)} the obstacle radii and \textit{(iii)} the obstacle positions, to de-conflict the agents while avoiding blocking their pathways. 

Fig. \ref{subfig:performanceRandomnessPerformance} shows the performance with different numbers of random agents $\eta$. The proposed method maintains a good performance across varying agent randomness, highlighting its capacity in handling different navigation scenarios. SPL and PCTSpeed decrease and NumCOLL increases slightly with the agent randomness because the navigation task becomes more challenging as the initial and goal positions of the agents get more random. Fig. \ref{subfig:performanceRandomnessSize} plots the average obstacle size and the average number of presented obstacles with different $\eta$. Both increase with the navigation randomness, while the increasing rates decrease as $\eta$ becomes large. The former is because when the randomness is small, the agents are capable of handling the de-confliction and need little help from the obstacles; when the randomness increases and the system becomes cluttered, the agents need more or larger obstacles to provide guidance for de-confliction. The latter is because as the obstacle number or radius reaches a saturated level, it has sufficient capacity to guide the agents and thus, reduces the increasing rate. Fig. \ref{fig:step4-3} plots examples of how our approach optimizes the obstacle layout for agents. Firstly, it guides the agent crowd and sorts out the navigation randomness; for example, it guides a subset of agents to slow down and take a detour to make way for the other agents. Secondly, it minimizes the use of collision-free space and does not obstruct agent pathways. 

The above results demonstrate that the obstacles are not always ``bad'', but instead, can play ``good'' guidance roles for the agents with an appropriate layout. These insights can be used for traffic design to facilitate safe and efficient navigation.

\begin{figure*}%
	\centering
\captionsetup[subfigure]{justification=centering}
	\begin{subfigure}{0.425\columnwidth}
		\includegraphics[width=1\linewidth,height = 0.6\linewidth]{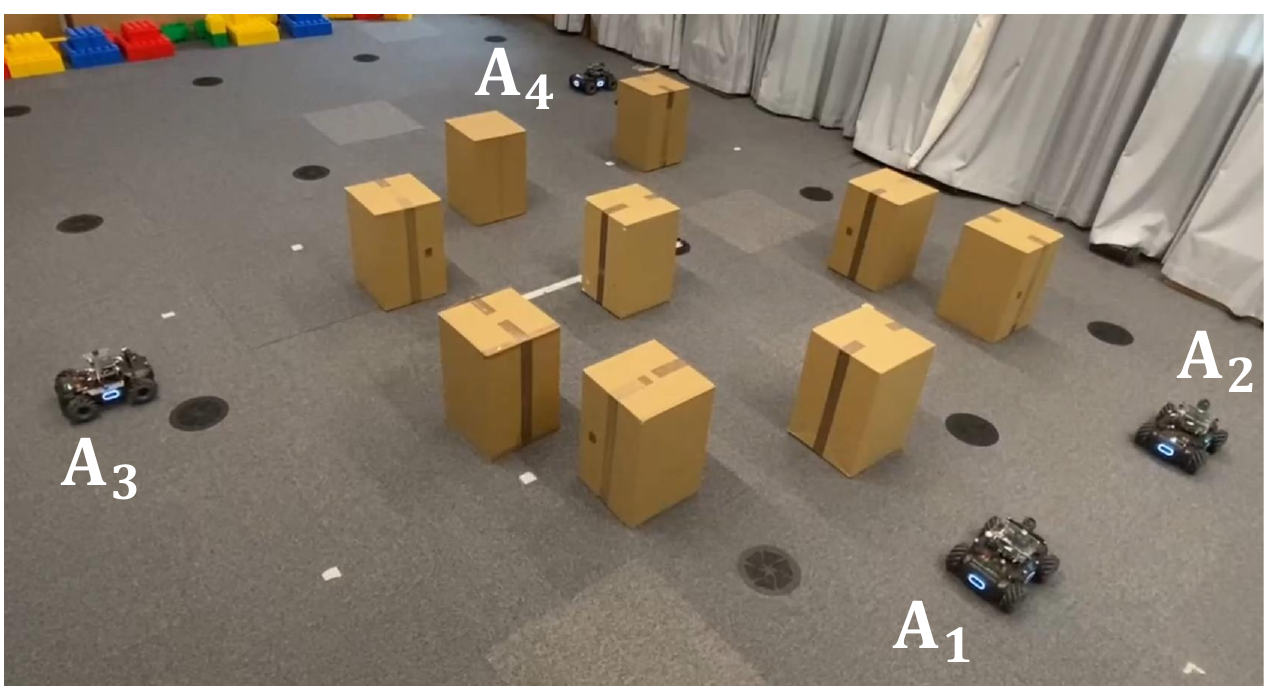}%
		\caption{$t=0$ seconds}%
		\label{subfig:0-11}%
	\end{subfigure}\hfill\hfill%
	\begin{subfigure}{0.425\columnwidth}
		\includegraphics[width=1\linewidth, height = 0.6\linewidth]{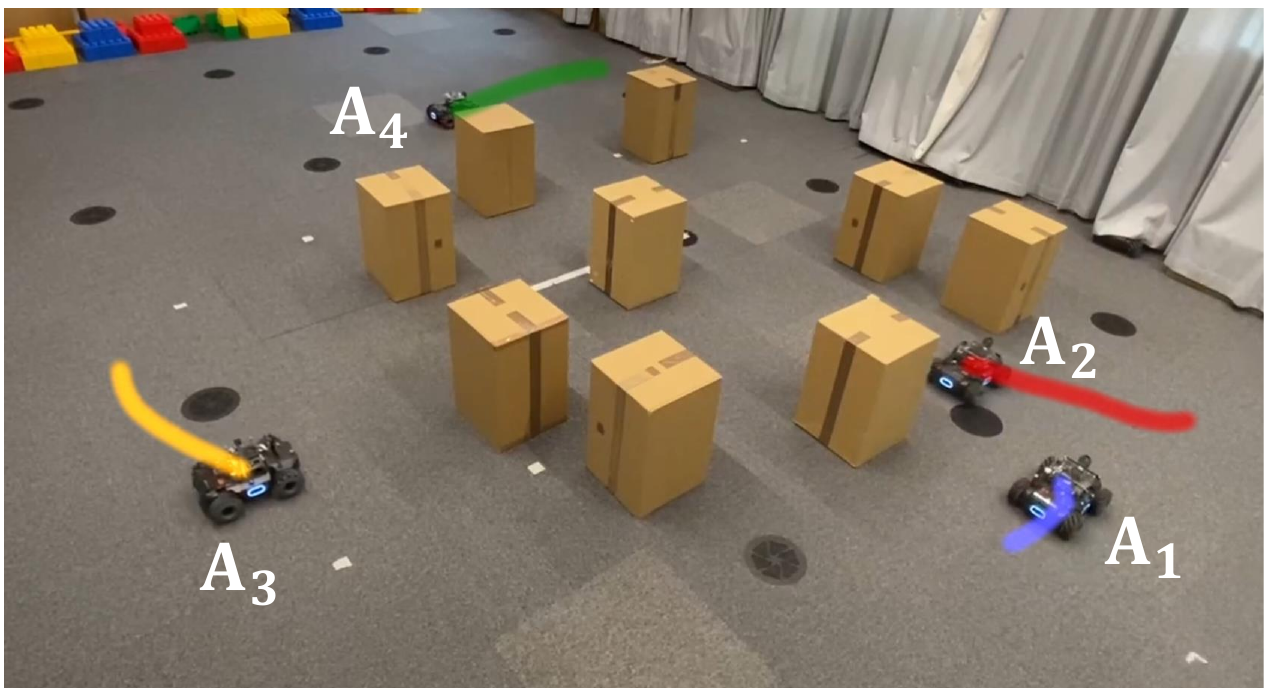}%
		\caption{$t=1$ seconds}%
		\label{subfig:0-12}
	\end{subfigure}\hfill\hfill%
	\begin{subfigure}{0.425\columnwidth}
		\includegraphics[width=1\linewidth,height = 0.6\linewidth]{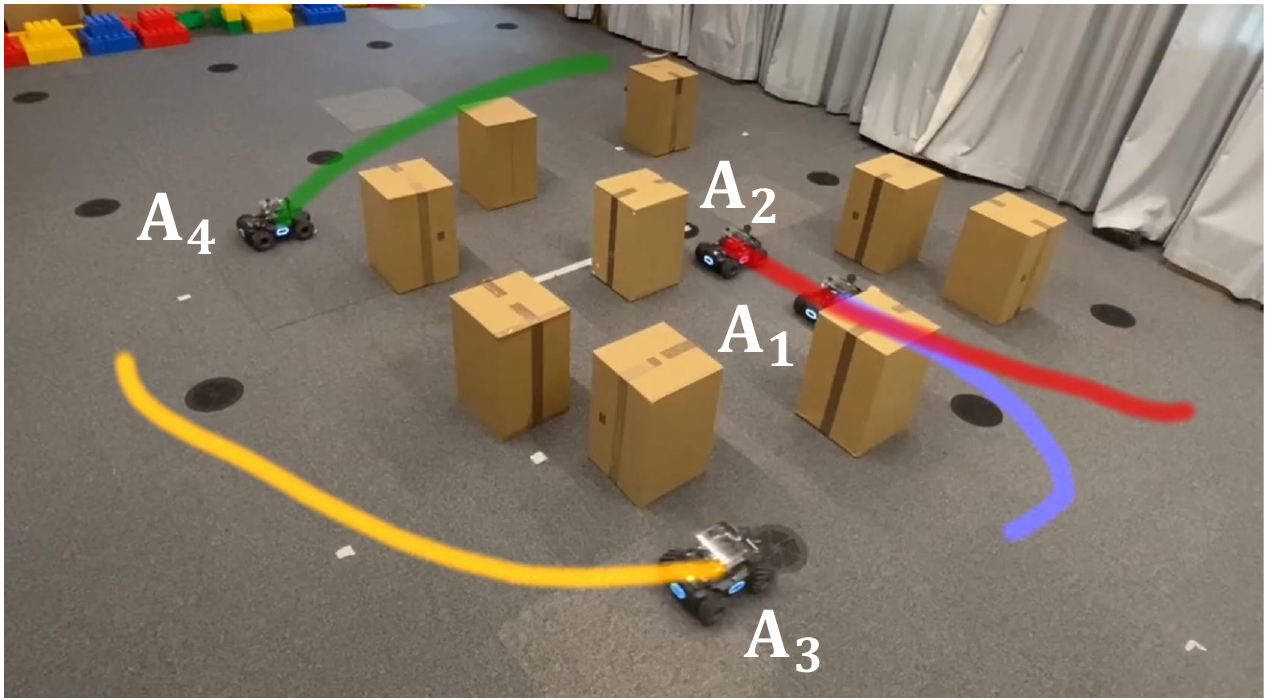}%
		\caption{$t=4$ seconds}%
		\label{subfig:0-13}%
	\end{subfigure}\hfill\hfill%
	\begin{subfigure}{0.425\columnwidth}
		\includegraphics[width=1\linewidth,height = 0.6\linewidth]{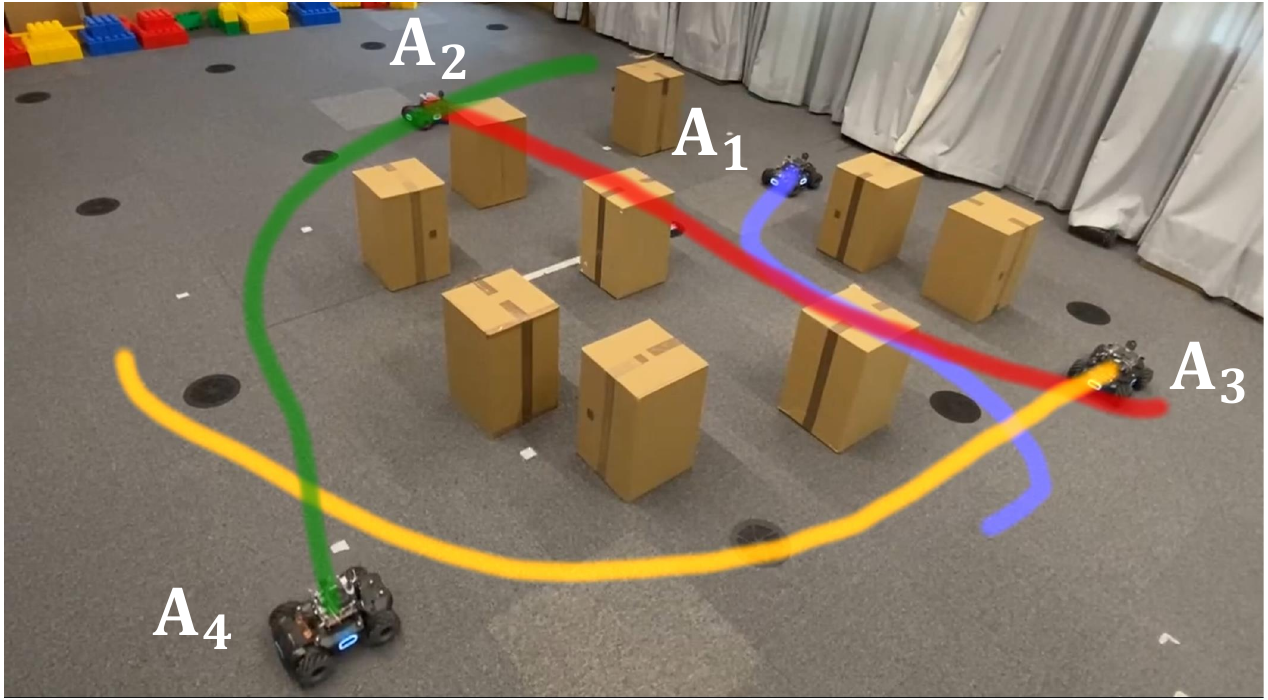}%
		\caption{$t=6$ seconds}%
		\label{subfig:0-14}%
	\end{subfigure}
	\begin{subfigure}{0.425\columnwidth}
		\includegraphics[width=1\linewidth,height = 0.6\linewidth]{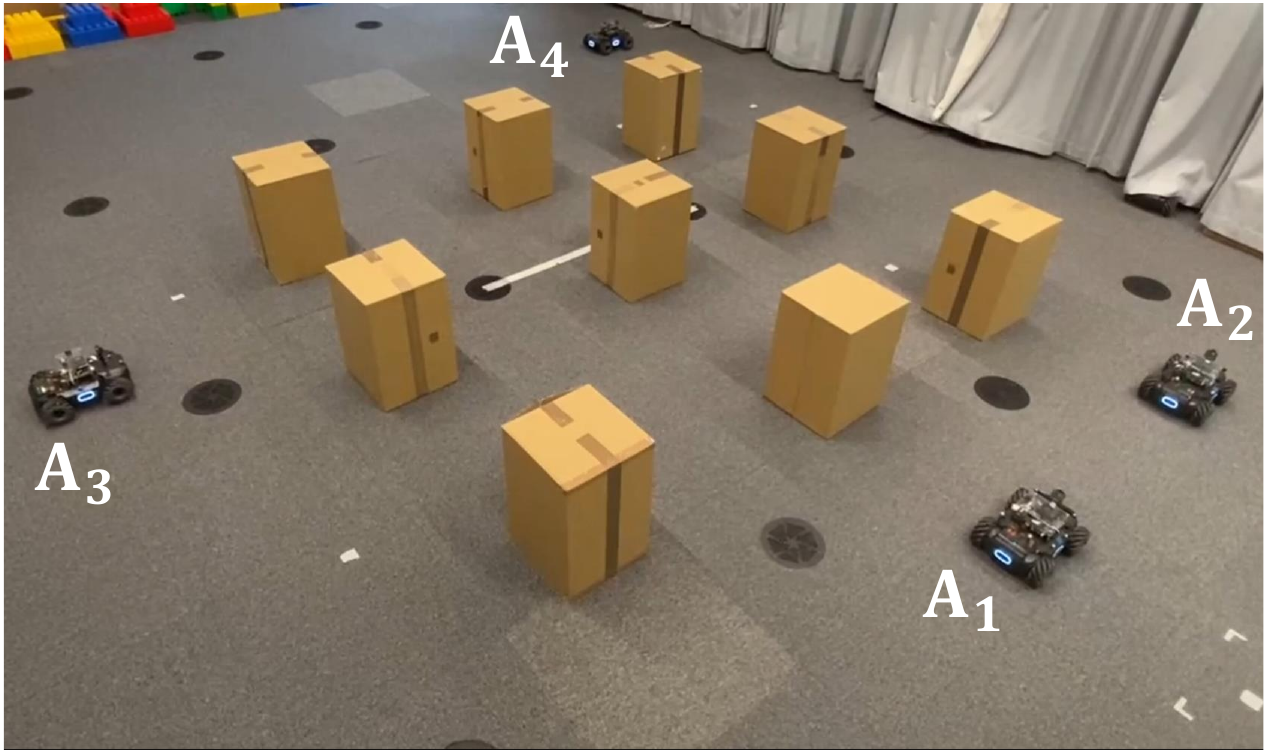}%
		\caption{$t=0$ seconds}%
		\label{subfig:0-15}%
	\end{subfigure}\hfill\hfill%
	\begin{subfigure}{0.425\columnwidth}
		\includegraphics[width=1\linewidth, height = 0.6\linewidth]{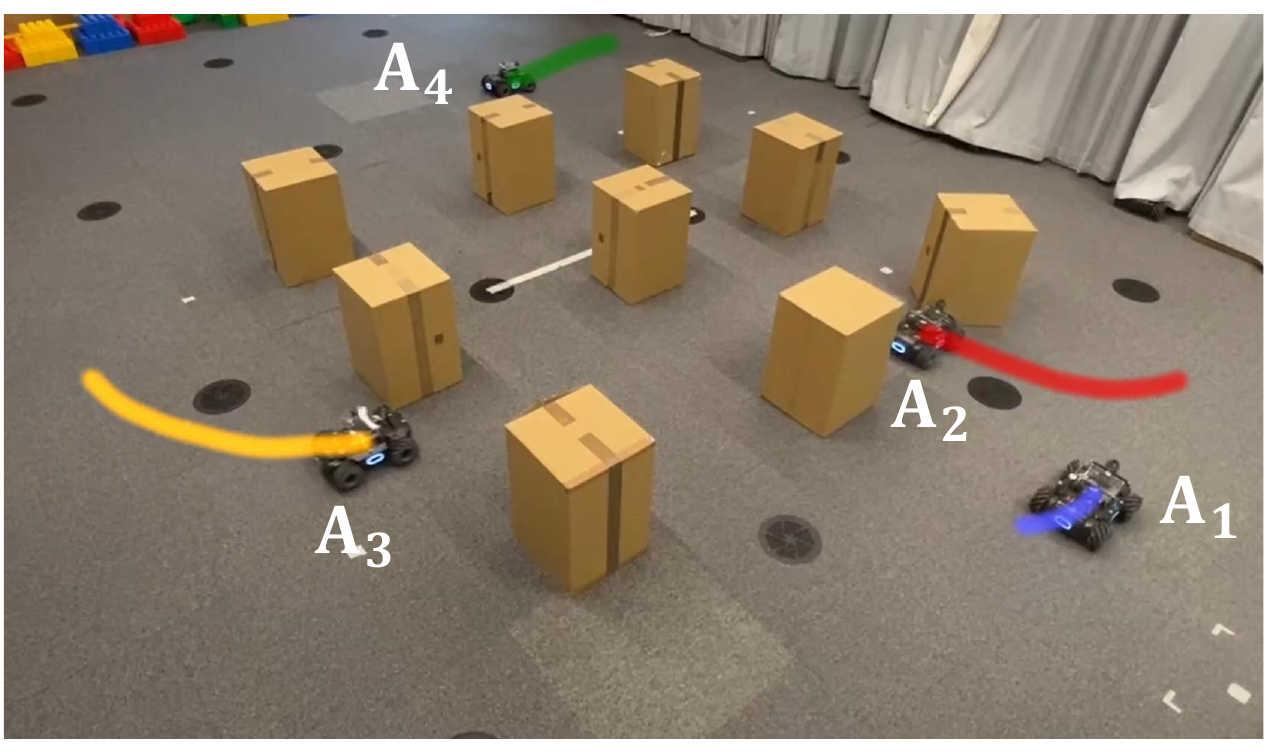}%
		\caption{$t=2$ seconds}%
		\label{subfig:0-16}
	\end{subfigure}\hfill\hfill%
	\begin{subfigure}{0.425\columnwidth}
		\includegraphics[width=1\linewidth,height = 0.6\linewidth]{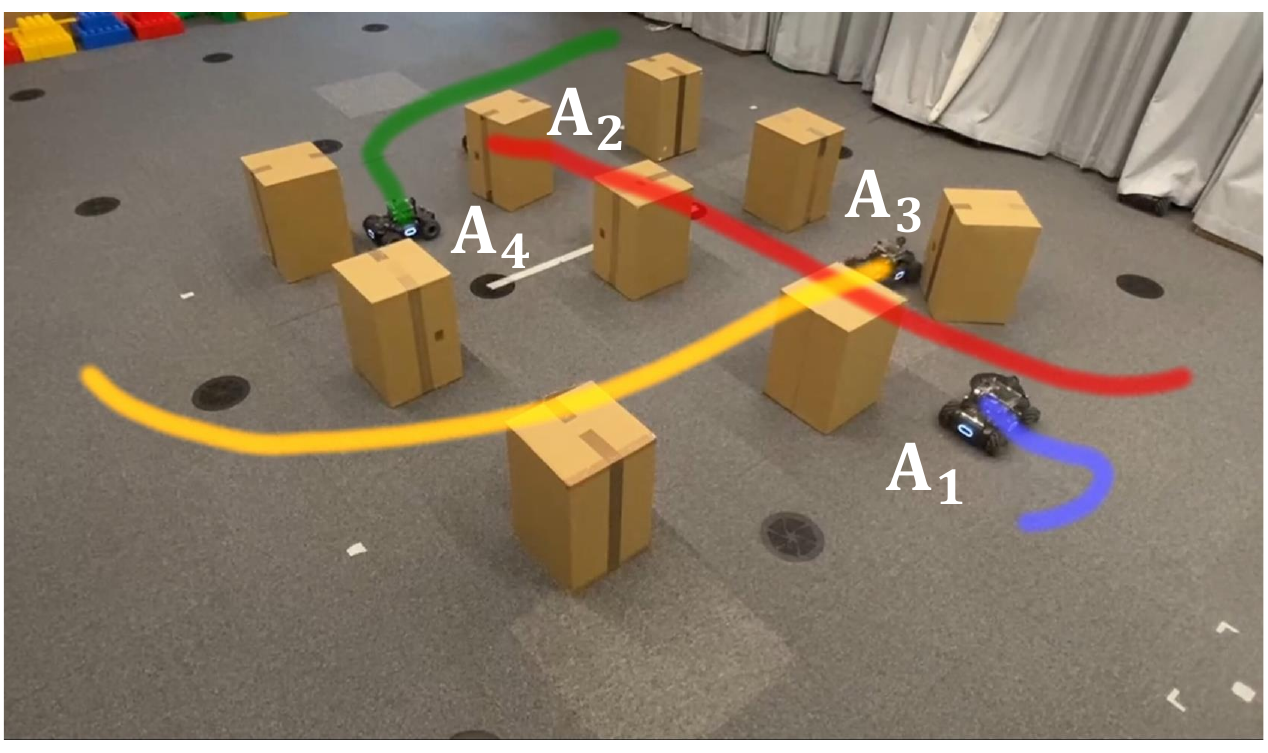}%
		\caption{$t=6$ seconds}%
		\label{subfig:0-17}%
	\end{subfigure}\hfill\hfill%
	\begin{subfigure}{0.425\columnwidth}
		\includegraphics[width=1\linewidth,height = 0.6\linewidth]{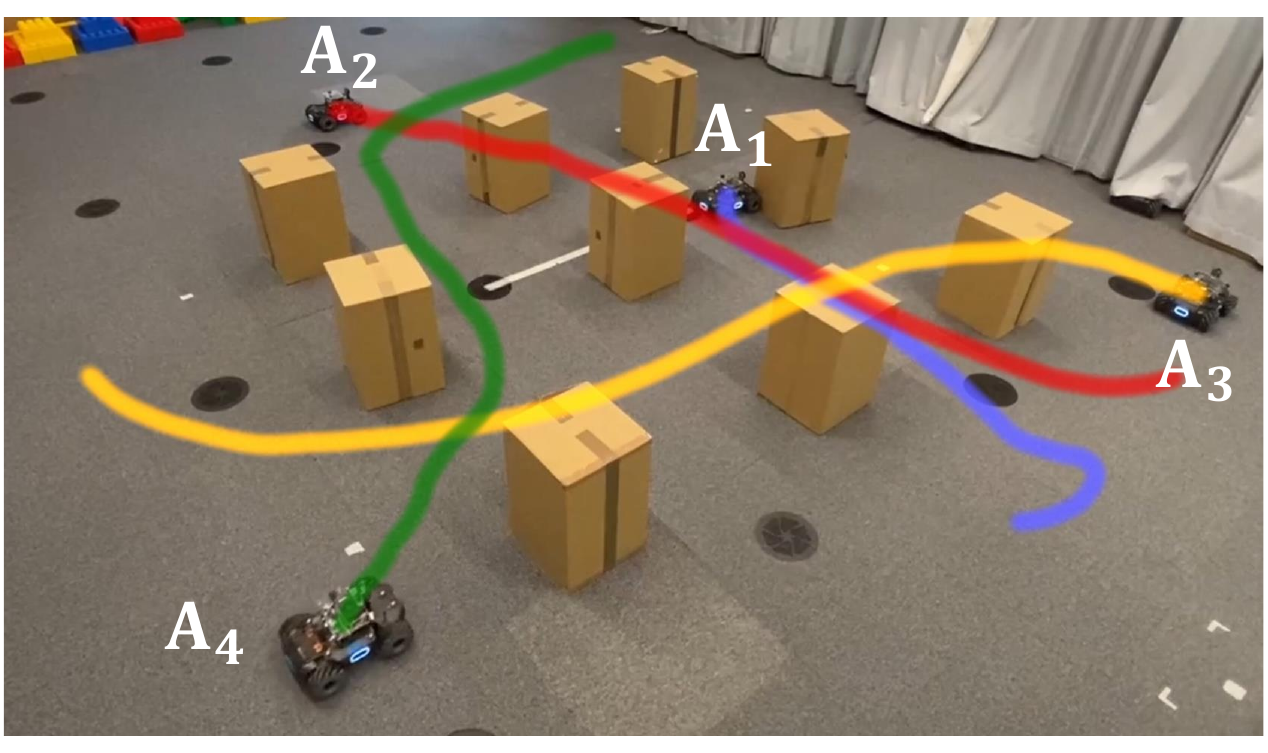}%
		\caption{$t=12$ seconds}%
		\label{subfig:0-18}%
	\end{subfigure}
	\caption{Real-world experiments in the warehouse setting. Robots are required to pass through obstacles towards goals, where colored lines are agent trajectories. (a-d) Performance of agent-environment coordinated optimization. (e-h) Performance of hand-designed baseline. Qualitatively, the optimized scenario in (a-d) showcases visually smoother trajectories than the hand-designed one in (e-h).}\label{fig:realExperimentsWarehouse}\vspace{-6mm}
\end{figure*}

\subsection{Real-World Experiments}

We conduct real-world experiments to corroborate simulation results. We consider two scenarios, i.e., the warehouse setting in Sec. \ref{subsec:performance} and the circular setting in Sec. \ref{subsec:blockingorGuidance}, where the former leverages agent-environment co-optimization to improve navigation performance and the latter explores the guidance role of the environment in multi-agent de-confliction. We use the Cambridge RoboMaster robots \cite{blumenkamp2024cambridge} equipped with Raspberry Pi in the workspace of size $4.5$m$\times 4.5$m, where each robot has a partially observable space with a communication range of $1$m. We use ROS2 as the communication middle-ware and employs external telemetry (OptiTrack) for localization. At each time, the robot captures its own state, receives neighbors' states, and generates the control output with the navigation policy in a decentralized manner. We use Model Predictive Control (MPC) as the low-level controller, which converts the control output to the wheel speed using inverse kinematics, moving the robot to its goal -- see Appendix \ref{appendix:dynamics}.

\begin{figure*}%
	\centering
	\captionsetup[subfigure]{justification=centering}
	\begin{subfigure}{0.41\columnwidth}
		\includegraphics[width=1\linewidth,height = 0.7\linewidth]{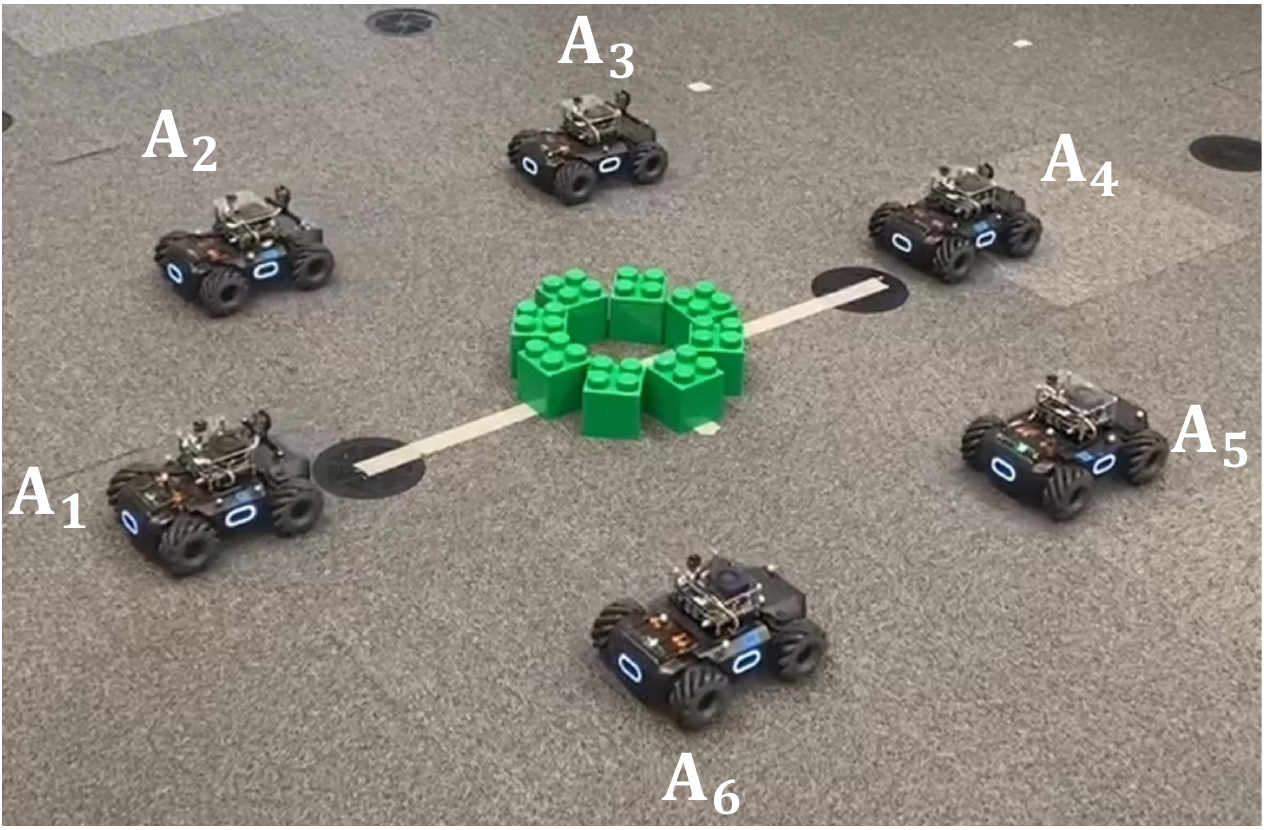}%
		\caption{$t=0$ seconds}%
		\label{subfig:0-21}%
	\end{subfigure}\hfill\hfill%
	\begin{subfigure}{0.41\columnwidth}
		\includegraphics[width=1\linewidth, height = 0.7\linewidth]{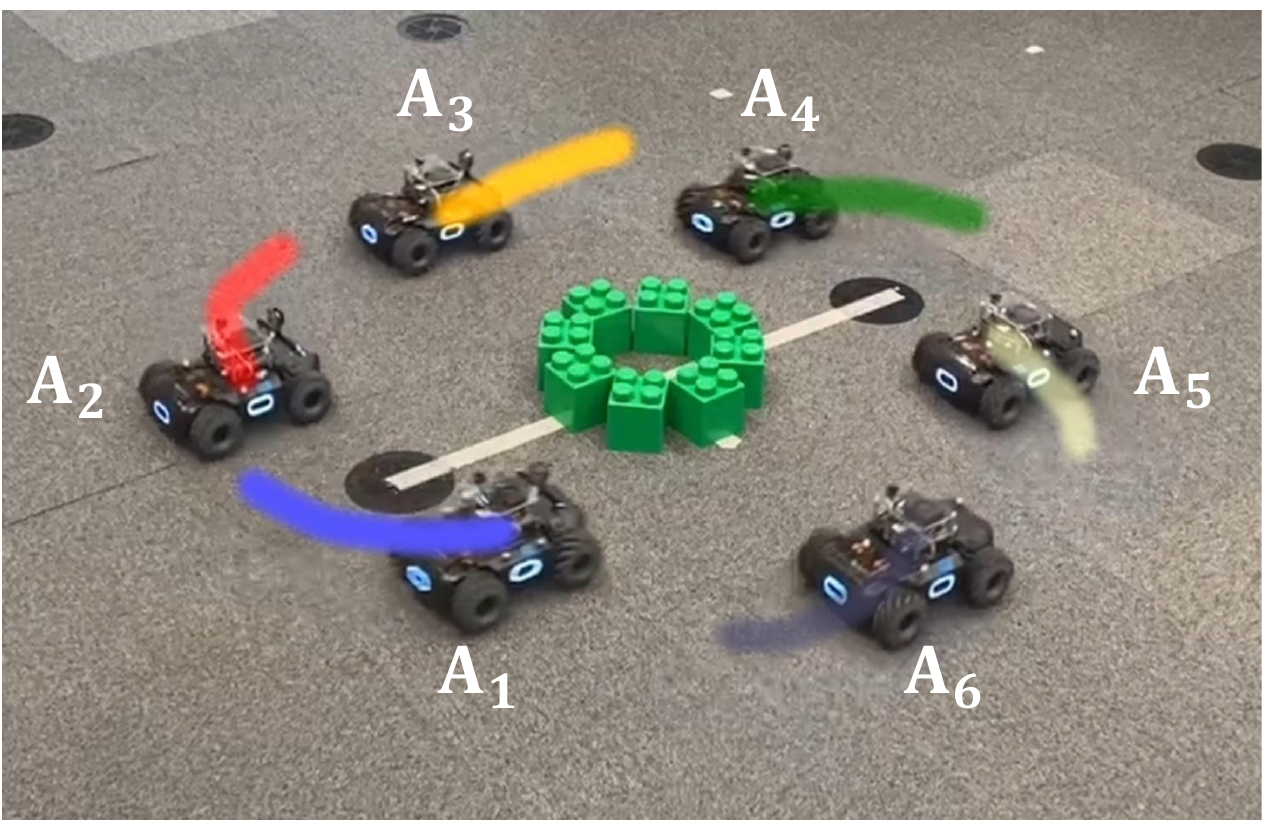}%
		\caption{$t=1$ seconds}%
		\label{subfig:0-22}
	\end{subfigure}\hfill\hfill%
	\begin{subfigure}{0.41\columnwidth}
		\includegraphics[width=1\linewidth,height = 0.7\linewidth]{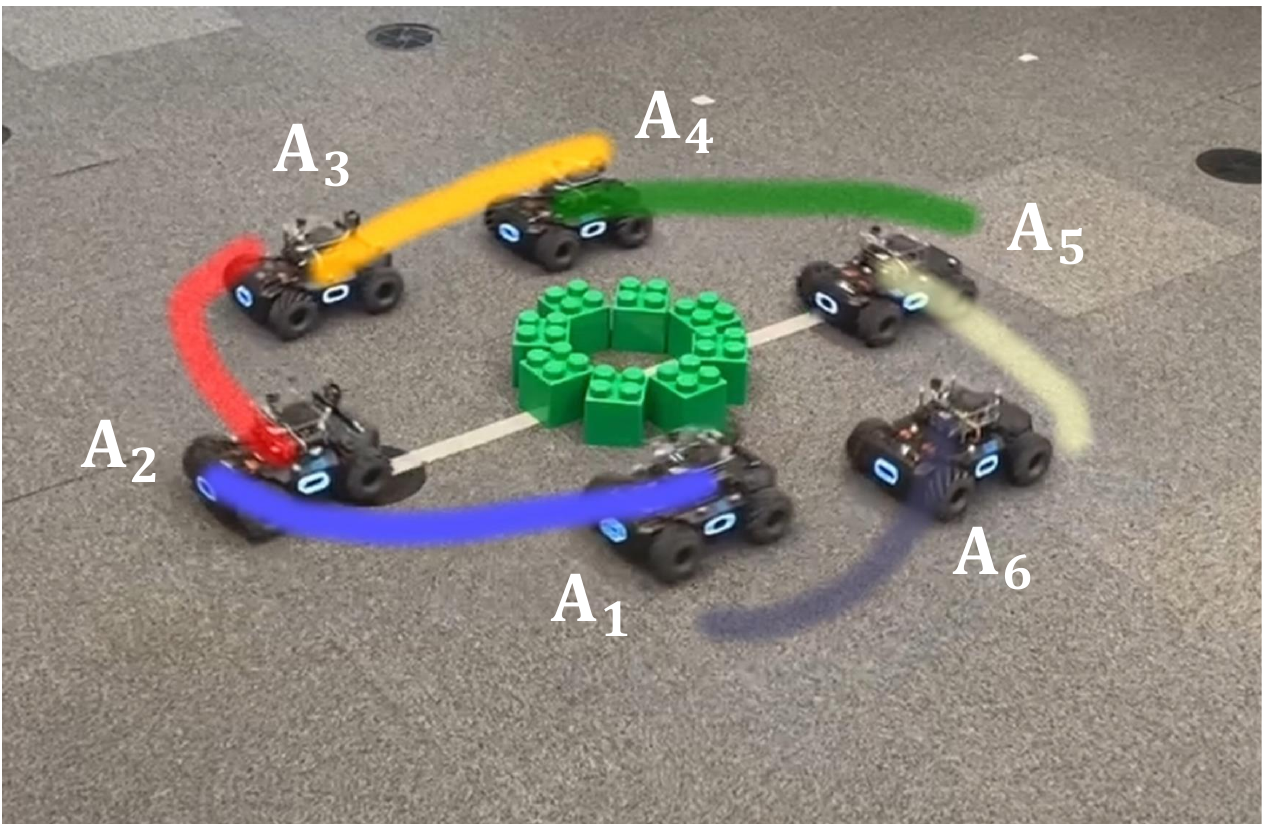}%
		\caption{$t=2$ seconds}%
		\label{subfig:0-23}%
	\end{subfigure}\hfill\hfill%
	\begin{subfigure}{0.41\columnwidth}
		\includegraphics[width=1\linewidth,height = 0.7\linewidth]{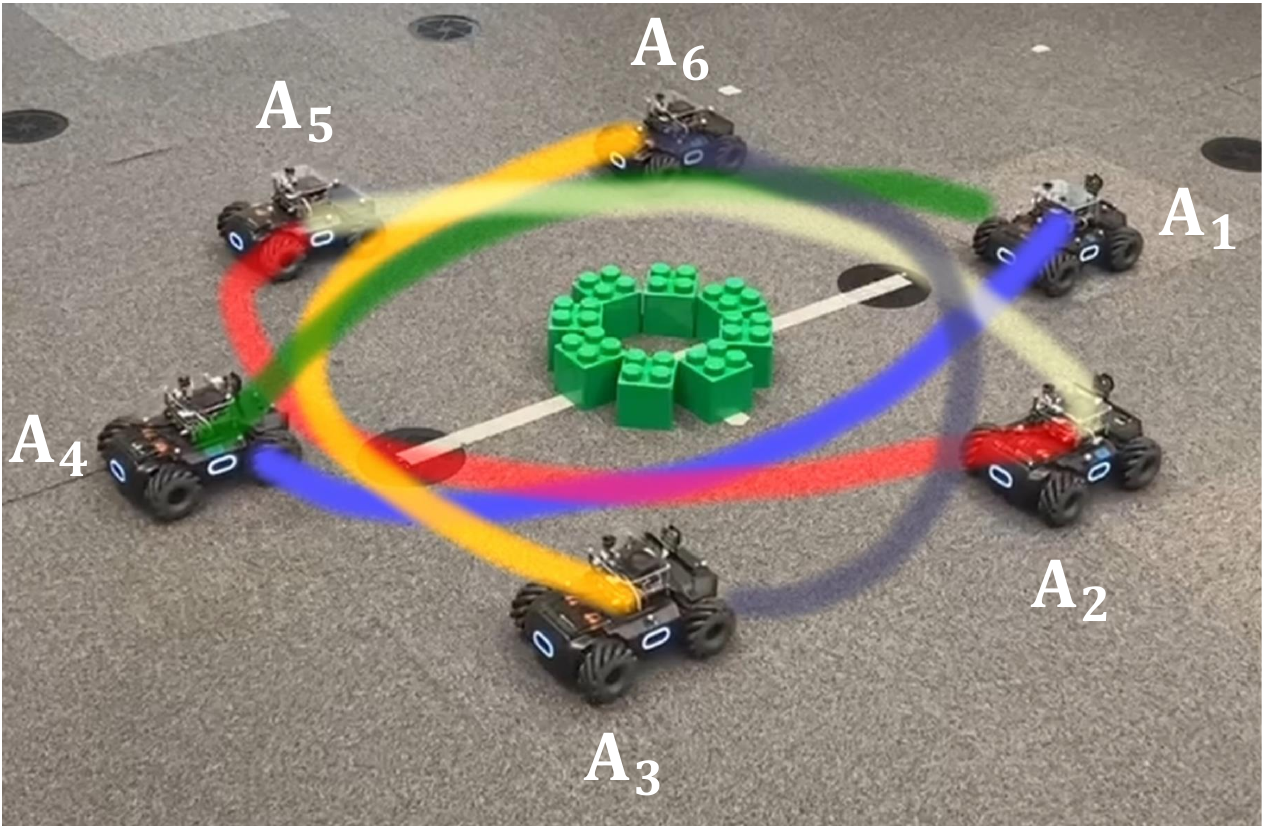}%
		\caption{$t=3$ seconds}%
		\label{subfig:0-24}%
	\end{subfigure}
	\begin{subfigure}{0.41\columnwidth}
		\includegraphics[width=1\linewidth,height = 0.7\linewidth]{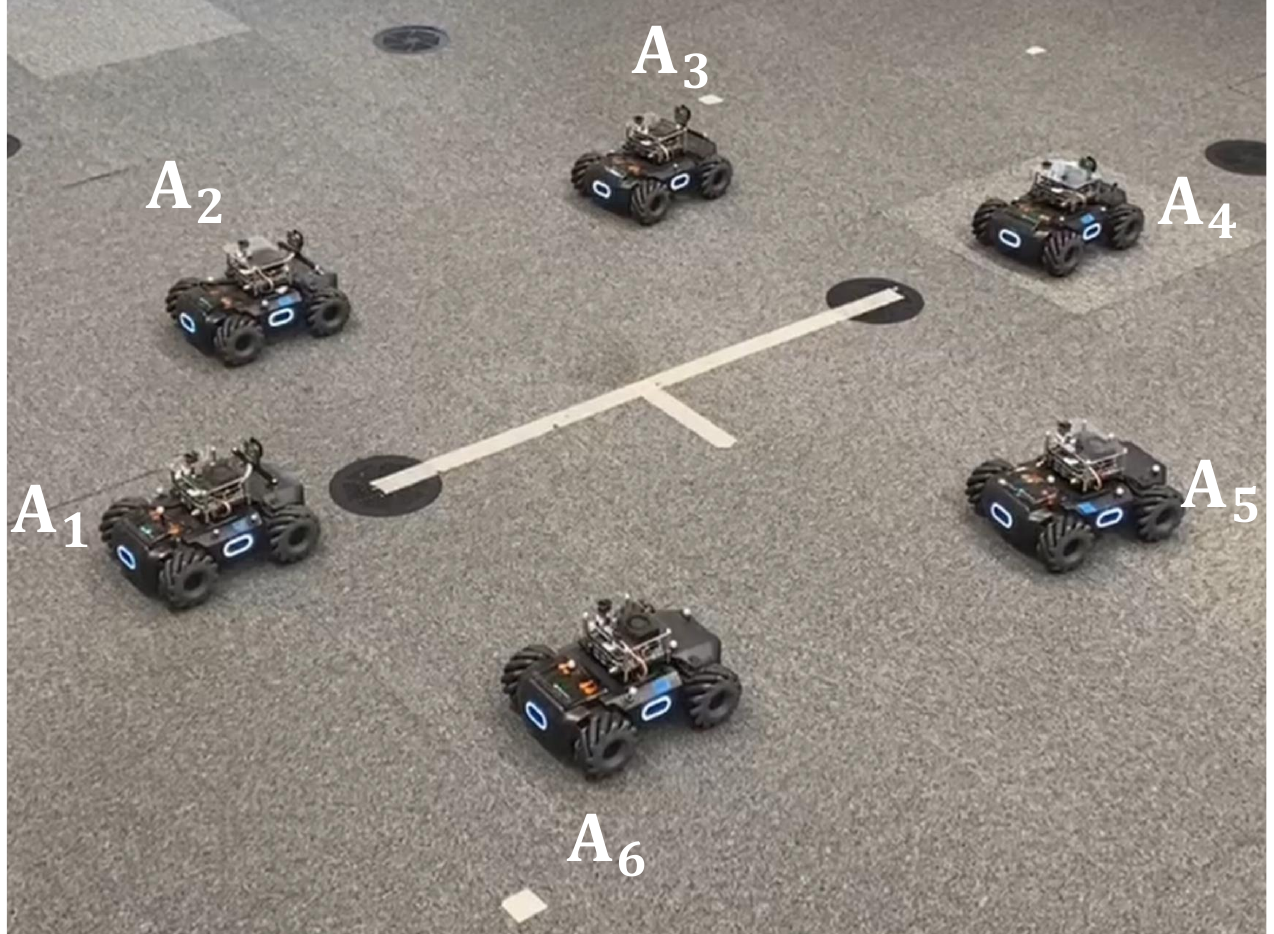}%
		\caption{$t=0$ seconds}%
		\label{subfig:0-25}%
	\end{subfigure}\hfill\hfill%
	\begin{subfigure}{0.41\columnwidth}
		\includegraphics[width=1\linewidth, height = 0.7\linewidth]{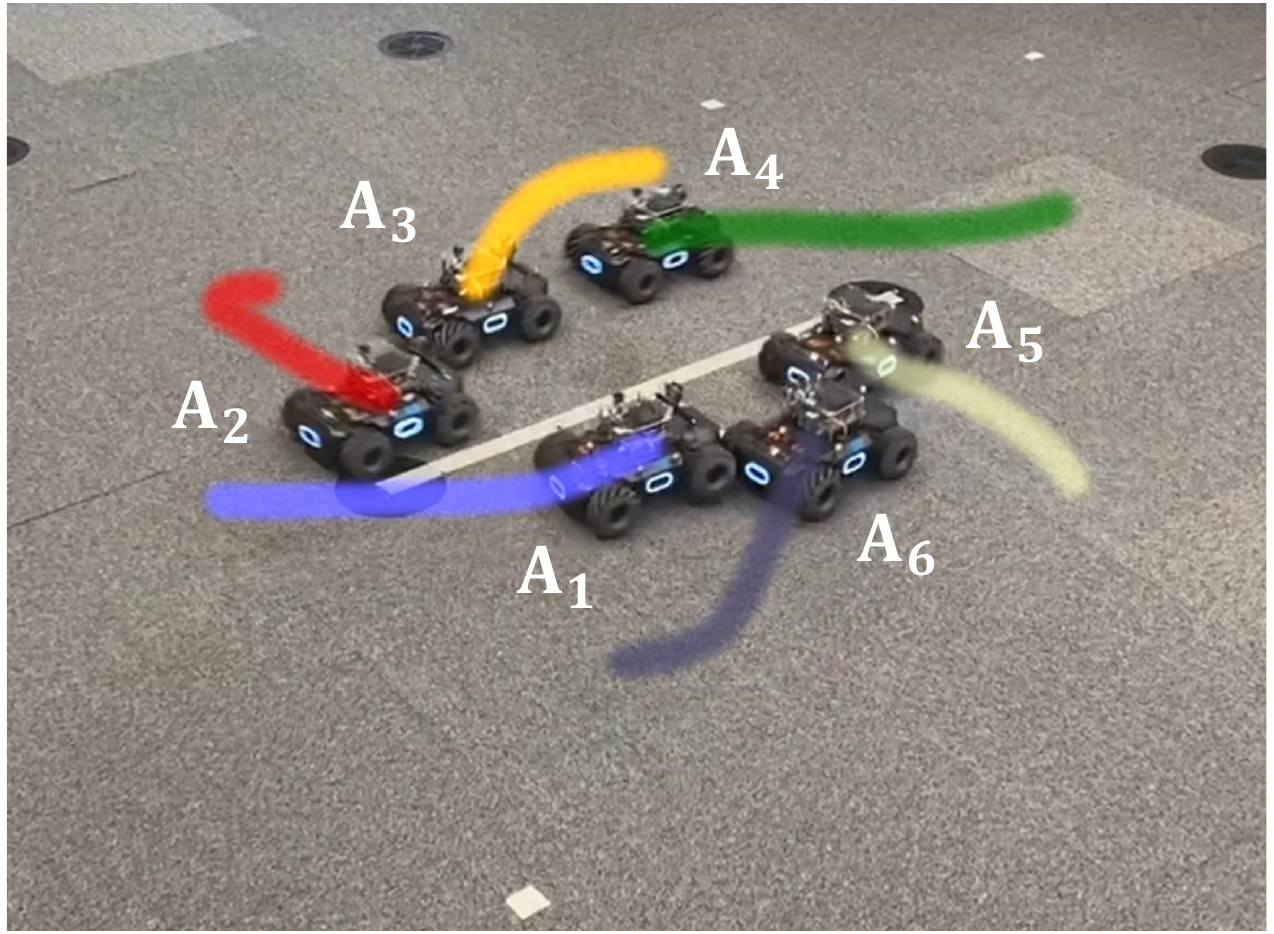}%
		\caption{$t=2$ seconds}%
		\label{subfig:0-26}
	\end{subfigure}\hfill\hfill%
	\begin{subfigure}{0.41\columnwidth}
		\includegraphics[width=1\linewidth,height = 0.7\linewidth]{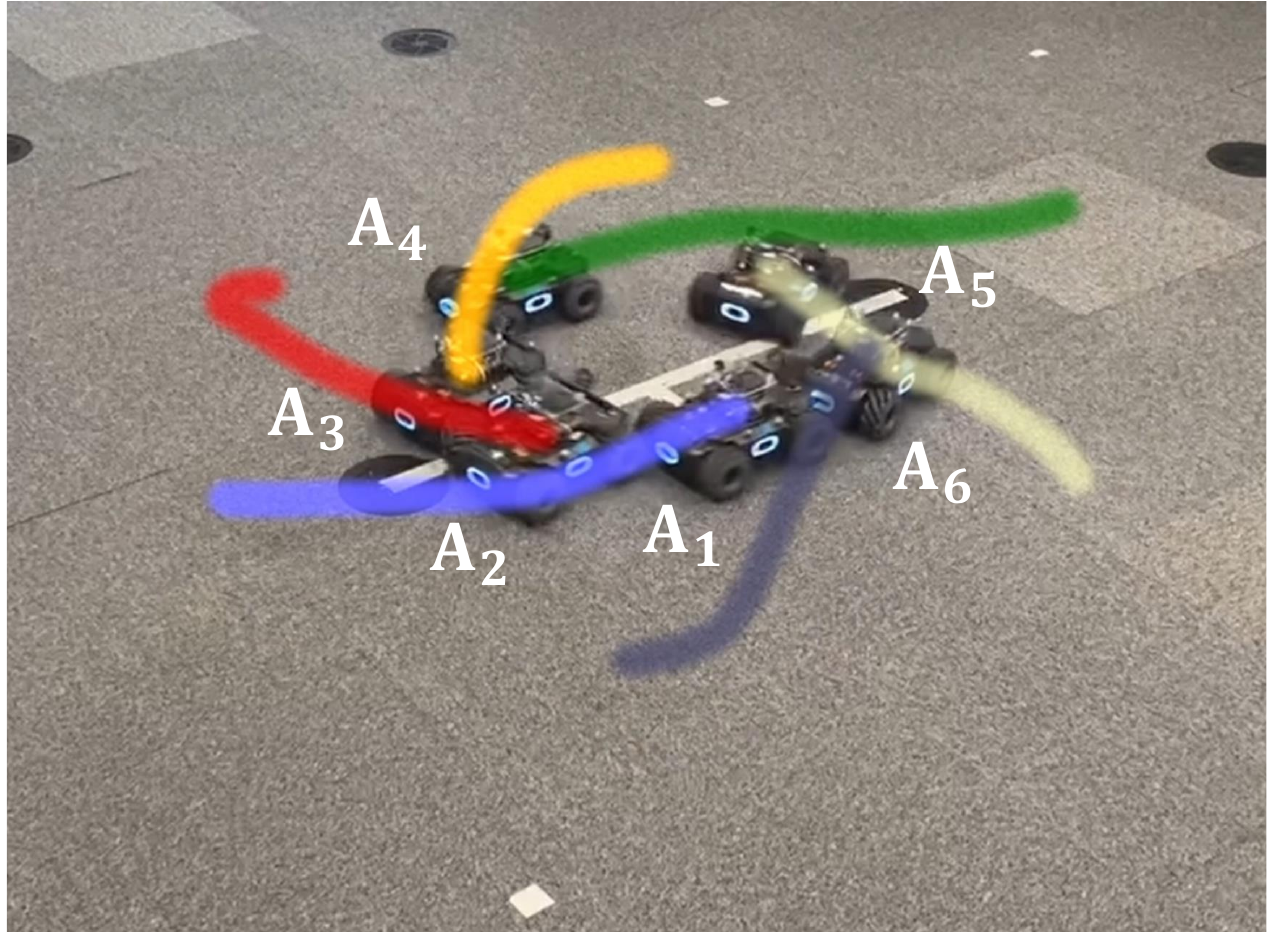}%
		\caption{$t=3$ seconds}%
		\label{subfig:0-27}%
	\end{subfigure}\hfill\hfill%
	\begin{subfigure}{0.41\columnwidth}
		\includegraphics[width=1\linewidth,height = 0.7\linewidth]{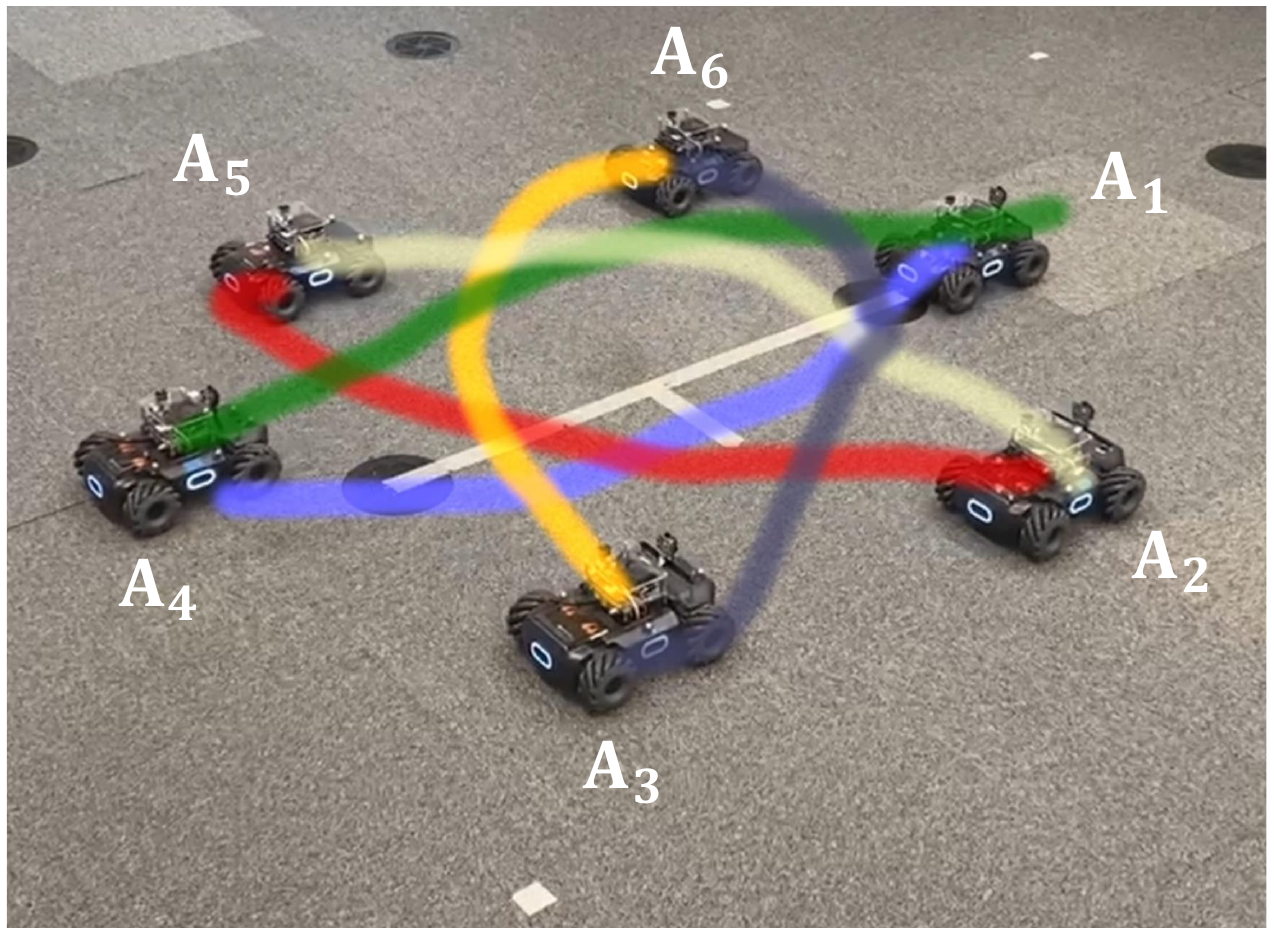}%
		\caption{$t=5$ seconds}%
		\label{subfig:0-28}%
	\end{subfigure}
	\caption{Real-world experiments in the circular setting. Robots are required to cross-navigate towards the opposite side while avoiding collisions. (a-d) Performance of agent-environment coordinated optimization. (e-h) Performance of hand-designed default (i.e., empty) baseline. Qualitatively, the optimized scenario in (a-d) showcases visually smoother trajectories than the hand-designed one in (e-h).}\label{fig:realExperiments}\vspace{-6.5mm}
\end{figure*}

Figs. \ref{subfig:0-11}-\ref{subfig:0-14} and Figs. \ref{subfig:0-21}-\ref{subfig:0-24} show the performance of our method in two scenarios, where the environment generative model produces compatible obstacle layouts and multi-agent policies generate smooth trajectories without collision. It outperforms the hand-designed baseline in Figs. \ref{subfig:0-15}-\ref{subfig:0-18} and Figs. \ref{subfig:0-25}-\ref{subfig:0-28}, which corroborates simulation results. Moreover, our method exhibits similarly good performance in real-world experiments as in numerical simulations. This is due to that the optimized environment creates more collision-free space and provides more de-confliction guidance, which ease the navigation task. It is thus more robust to the sim-to-real impact during real-world deployment. In contrast, the hand-designed baseline performs worse (e.g., lower path efficiencies, slower speed and more collisions) in real-world experiments than in numerical simulations. This is because the hand-designed environment is not optimally compatible with the navigation task, requiring more accurate execution of the navigation policy during real-world deployment, and is hence affected more seriously by the sim-to-real gap. 

\subsection{Discussions}

In summary, the coordinated optimization method builds a symbiotic agent-environment co-existing system, where environment configurations and multi-agent policies adapt to and rely on each other to jointly improve performance. For one thing, our method optimizes the obstacle layout to create efficient obstacle-free pathways for agents and generates compatible agent trajectories for smooth navigation -- see Sec. \ref{subsec:performance}. For another, our method exploits the obstacle layout (e.g., structure irregularity and non-symmetry) to prioritize created obstacle-free pathways and de-conflict agents via prioritization, which provide guidance for agents to facilitate safe navigation -- see Sec. \ref{subsec:blockingorGuidance}. Compared to hand-designed environments, our approach achieves higher success rates and path efficiencies (SPL), moving speed (PCTSpeed), in the meantime, providing better navigation safety (NumCOLL) and acceleration comfort (DiffACC). It offers a comprehensively improved performance for multi-agent navigation and is well-applicable to various settings. Additional experiments are provided in Appendix \ref{appendix:additional} to further validate the proposed method.

Moreover, our framework allows high degrees of freedom for environment optimization. Specifically, in our experiments, the environment generative model allows three design choices for each obstacle: (i) The presence or absence of an obstacle; (ii) the obstacle position; (iii) the obstacle size. These re-configurable parameters together yield a rich family of possible environment configurations, which can be used as potential candidates for different multi-agent navigation tasks.

\section{Conclusion} 
\label{sec:conclusion}

This paper proposed an agent-environment coordinated optimization framework for multi-agent systems. It comprises two components: \textit{(i)} multi-agent navigation and \textit{(ii)} environment optimization. The former seeks a navigation strategy that moves agents from origins to goals, and the latter seeks a design scheme that generates an appropriately accommodating environment (w.r.t the task). These components must be optimized jointly such that an overarching objective is achieved (in our case: improved navigation performance). We solved this problem by alternating the optimization of the multi-agent navigation policy and an environment generative model. We followed a model-free learning-based approach, and introduced a novel combination of reinforcement learning (through an actor-critic mechanism) and unsupervised learning (through policy gradient ascent) for coordinated optimization. Furthermore, we analyzed the convergence of our method by exploring its relation with an associated time-varying non-convex optimization problem and ordinary differential equations. Numerical results and real-world experiments corroborated theoretical findings and showed the benefit of our method, i.e., the agent-environment co-optimization finds agent-friendly environments and efficient navigation trajectories. Finally, we investigated the role of the environment in multi-agent navigation to show that it does not just impose physical restrictions, but can also provide guidance for agent de-confliction if designed appropriately. The latter explores an inherent relationship between the environment and agents in a way previously unaddressed. Our framework has limitations when the obstacle layout of the environment is not reconfigurable or with limited flexibility for reconfiguration. It reduces to nominal multi-agent navigation when the environment is not reconfigurable and has a reduced performance improvement when the environment has limited reconfigurable flexibility. 

Future work will extend the agent-environment co-optimization to life-long multi-agent navigation with multiple target positions through a problem re-formulation. We plan to model multiple targets as a sequence of waypoints, assign them as intermediate sub-goals, and move the agents to pass through all sub-goals towards the final destination. The key step is to develop scheduling strategies that assign targets to successive waypoints for each agent and re-design reward functions to learn multi-agent policies that pass through all waypoints.

\appendices 

\section{Proof of Theorem \ref{thm:Convergence}}\label{proof:Theorem1}

First, from Theorem 3.3 in \cite{khalil2002nonlinear}, the ODE \eqref{eq:ODE} has a unique solution $\bbtheta_o(\alpha)$ starting from the point $\bbtheta_o^{(0)} = \bbtheta_o(0)$. Then, we show the discrete parameters $\big\{\bbtheta_o^{(k)}\big\}_k$ of the proposed method converge to this solution $\bbtheta_o(\alpha)$. 
From Taylor’s theorem, we can expand the ODE solution $\bbtheta_o(\alpha)$ at $\alpha^{(k+1)}$ as
\begin{align}\label{proof:thm1eq1}
	&\bbtheta_o(\alpha^{(k+1)}) = \bbtheta_o(\alpha^{(k)}) + \Delta \alpha \bbtheta'_o(\alpha^{(k)}) + \ccalO(\Delta \alpha^2), 
\end{align}
where $\ccalO(\Delta \alpha^2)$ can be bounded by $C_h \Delta \alpha^2$ with $C_h$ the bounding constant of the higher-order term over the finite horizon $\alpha^{(k)} \in [0, T]$. By substituting \eqref{eq:ODE} into \eqref{proof:thm1eq1}, we have
\begin{align}\label{proof:thm1eq2}
	&\bbtheta_o\!(\!\alpha^{(k\!+\!1)}\!) \!=\! \bbtheta_o\!(\!\alpha^{(k)}\!) \!-\! \frac{\Delta \alpha}{\eta}\! \nabla_{\bbtheta_o} g(\bbS,\!\bbD,\!\bbtheta_o\!(\!\alpha^{(k)}\!),\! \alpha^{(k)}\!) \!+\! \ccalO(\Delta \alpha^2) . 
\end{align}
Recall the proposed method updates the generative parameters $\bbtheta^{(k)}_o$ by the policy gradient ascent w.r.t. the objective function $f(\bbS,\bbD,\bbtheta_o^{(k)}, \alpha^{(k)})$ of step-size $\Delta \beta = \Delta \alpha / \eta$, i.e., 
\begin{align}\label{proof:thm1eq4}
	\bbtheta_o^{(k+1)} &= \bbtheta_o^{(k)} + \frac{\Delta \alpha}{\eta} \widetilde{\nabla}_{\bbtheta_o} f(\bbS,\bbD,\bbtheta_o^{(k)}, \alpha^{(k)}) \\
	& = \bbtheta_o^{(k)} - \frac{\Delta \alpha}{\eta} \widetilde{\nabla}_{\bbtheta_o} g(\bbS,\bbD,\bbtheta_o^{(k)}, \alpha^{(k)}), \nonumber
\end{align}
where $ \widetilde{\nabla}_{\bbtheta_o} f$ and $\widetilde{\nabla}_{\bbtheta_o} g$ represent the policy gradients that approximates the true gradients $ \nabla_{\bbtheta_o} f$ and $\nabla_{\bbtheta_o} g$. By subtracting \eqref{proof:thm1eq4} from \eqref{proof:thm1eq2}, we get
\begin{align}\label{proof:thm1eq5}
	&\bbe_o^{(k+1)} = \bbe_o^{(k)} + \ccalO(\Delta \alpha^2) \\
	& +\! \frac{\Delta \alpha}{\eta} \Big(\widetilde{\nabla}_{\bbtheta_o} g(\bbS,\bbD,\bbtheta_o^{(k)}\!, \alpha^{(k)}) \!-\! \nabla_{\bbtheta_o} g(\bbS,\bbD,\bbtheta_o(\!\alpha^{(k)}), \alpha^{(k)})\!\!\Big), \nonumber
\end{align}
where $\bbe_o^{(k)} = \bbtheta_o(\alpha^{(k)}) - \bbtheta_o^{(k)}$ is the deviation error. By using Assumption \ref{as2} and the triangle inqueality, we rewrite \eqref{proof:thm1eq5} as
\begin{align}\label{proof:thm1eq6}
	&\|\bbe_o^{(k+1)}\| \le \|\bbe_o^{(k)}\| + C_h \eta^2 \Big(\frac{\Delta \alpha}{\eta}\Big)^2 + \varepsilon\frac{\Delta \alpha}{\eta} \\
	& +\! \frac{\Delta \alpha}{\eta} \|\nabla_{\bbtheta_o} g(\bbS,\bbD,\bbtheta_o^{(k)}\!, \alpha^{(k)}) \!-\! \nabla_{\bbtheta_o} g(\bbS,\bbD,\bbtheta_o(\!\alpha^{(k)}\!), \alpha^{(k)})\|. \nonumber
\end{align}
By using Assumption \ref{as1} and $\bbtheta_o(\alpha^{(k)}) = \bbtheta_o^{(k)} + \bbe_o^{(k)}$, we have
\begin{align}\label{proof:thm1eq7}
	\|\!\nabla_{\bbtheta_o} g(\bbS,\!\bbD,\!\bbtheta_o^{(k)}\!\!,\! \alpha^{(k)}\!) \!-\! \nabla_{\bbtheta_o} g(\bbS,\!\bbD,\!\bbtheta_o(\!\alpha^{(k)}\!),\! \alpha^{(k)}\!)\|\!\!\le\!\! C_L \| \bbe_o^{(k)}\! \|, 
\end{align}
and substituting \eqref{proof:thm1eq7} into \eqref{proof:thm1eq6} yields
\begin{align}\label{proof:thm1eq8}
	\|\bbe_o^{(k+1)}\| &\!\le\! (1 \!+\! \frac{\Delta \alpha}{\eta} C_L)\|\bbe_o^{(k)}\| \!+\! C_h \eta^2 \Big(\frac{\Delta \alpha}{\eta}\Big)^2 \!+\! \varepsilon \frac{\Delta \alpha}{\eta}.
\end{align}
Then, we show by induction that 
\begin{align}\label{proof:thm1eq9}
	\|\bbe_o^{(k)}\| \le \frac{C_h \eta \Delta \alpha + \varepsilon}{C_L}\Big(\big(1+ \frac{\Delta \alpha}{\eta} C_L\big)^k - 1\Big). 
\end{align}
For $k=0$, we need to show $\|\bbe_o^{(0)}\| \le 0$, i.e., $\bbe_o^{(k)} = \bb0$. This is true because the initial points $\bbtheta_o^{(0)}$ and $\bbtheta_o(0)$ are the same. For iteration $k$, we assume \eqref{proof:thm1eq9} holds and consider iteration $k+1$. By combining \eqref{proof:thm1eq9} with \eqref{proof:thm1eq8}, we bound $\|\bbe_o^{(k+1)}\|$ by 
\begin{align}\label{proof:thm1eq10}
	&\big(\!1 \!\!+\! \frac{\Delta \alpha}{\eta} C_L\!\big)\!\frac{C_h \eta \Delta \alpha \!\!+\!\! \varepsilon}{C_L}\!\Big(\!\!\big(\!1\!\!+\!\frac{\Delta \alpha}{\eta} C_L\!\big)^k \!\!\!-\!\! 1\!\!\Big) \!\!+\!C_h \eta^2 \!\Big(\!\!\frac{\Delta \alpha}{\eta}\!\Big)^2\!\!\!\!+\!\varepsilon \frac{\Delta \alpha}{\eta} \nonumber \\
	& = \frac{C_h \eta \Delta \alpha + \varepsilon}{C_L}\Big((1+\frac{\Delta \alpha}{\eta} C_L)^{k+1} - 1\Big).
\end{align}
This completes the inductive argument and proves \eqref{proof:thm1eq9}. 

Since the term $\Delta \alpha C_L / \eta > 0$ is positive, we have $1 + \Delta \alpha C_L / \eta \le e^{\Delta \alpha C_L / \eta}$ and thus $(1+ \Delta \alpha C_L/\eta )^k \le e^{ k \Delta \alpha C_L/\eta} \le e^{\lfloor T/\Delta \alpha \rfloor \Delta \alpha C_L/\eta }$, where $k \le \lfloor T/\Delta \alpha \rfloor$ is used. Further using the fact that $\Delta \alpha \lfloor T/\Delta \alpha \rfloor \le T$ yields
\begin{align}\label{proof:thm1eq12}
	\big(1+ \frac{\Delta \alpha}{\eta} C_L\big)^k \le e^{\frac{C_L}{\eta}T}.
\end{align}
By substituting \eqref{proof:thm1eq12} into \eqref{proof:thm1eq9}, we get
\begin{align}\label{proof:thm1eq13}
	\|\bbe_o^{(k)}\| &\le \frac{C_h \eta}{C_L}\Big(e^{\frac{C_L}{\eta}T} - 1\Big) \Delta \alpha \!+\! \frac{e^{\frac{C_L}{\eta}T} - 1}{C_L}\varepsilon.
\end{align}
The first term in \eqref{proof:thm1eq13} decreases with the step-size $\Delta \alpha$. This indicates that for any $\eps > 0$ and $T$, there exists a $\Delta \alpha$ s.t. 
\begin{align}\label{proof:thm1eq14}
	\frac{C_h \eta}{C_L}\Big(e^{\frac{C_L}{\eta}T} - 1\Big) \Delta \alpha \le \eps.
\end{align}
The second term in \eqref{proof:thm1eq13} is proportional to $\varepsilon$ w.r.t. a constant $C$. By using these results in \eqref{proof:thm1eq13}, we complete the proof
\begin{align}\label{proof:thm1eq15}
	\|\bbe_o^{(K)}\| \le \eps + C \varepsilon. 
\end{align}

\section{Information Processing Architecture}\label{sec:NNs}

Information processing architectures parameterize the navigation policy and the generative model, allowing to represent infinite dimensional policy functions with finite parameters -- see challenge (iii) of Sec. \ref{subsec:AgentEnvironment}. The selection of information processing architectures depends on specific problem requirements, where we consider a graph neural network (GNN) for the navigation policy due to its decentralized implementation and a deep neural network (DNN) for the environment generative model due to its strong expressive power. 

\subsection{GNN-Based Multi-Agent Policy}

The multi-agent system can be modeled as a graph $\ccalG$ with nodes $\{1,..., n\}$ and edges $\{(i,j)\}_{ij}$. Each node represents an agent, while each edge represents a link between two neighboring agents within communication radius. The graph structure is captured by a support 
matrix $\bbE$ with $[\bbE]_{ij} \ne 0$ if there is an edge between $i, j$ or $i=j$ and $[\bbE]_{ij} = 0$ otherwise. The node states are captured by a graph signal $\bbX$, which is a matrix with the $i$th row $[\bbX]_i$ representing the state of node $i$. For example, the support matrix is the adjacency matrix $\bbA$ and graph signals are positions and velocities of the agents.

\begin{figure}[!t]
\centering
    \resizebox{0.35 \textwidth}{!}{\input{NNs}}

\caption{General framework of parameterization. The information processing architecture parameterizes the navigation policy or the generative model. These parameters determine the policy distributions of the agents or the generative distribution of the environment, and the distribution samples the agent actions or the obstacle layout.}
\label{fig:framework}\vspace{-6mm}
\end{figure}

GNNs are layered architectures that exploit a message passing mechanism to extract features from graph signals \cite{scarselli2008graph, velivckovic2018graph, gao2021stochastic}. At each layer $\ell$, a GNN consists of the message aggregation function $\ccalF_{\ell,m}$ and the feature update function $\ccalF_{\ell,u}$. With the input signal $\bbX_{\ell-1}$ generated at the previous layer, $\ccalF_{\ell,m}$ collects the signal values of the neighboring nodes and generates the intermediate features as
\begin{align}\label{eq:intermediateFeature}
	[\bbU_{\ell}]_i \!=\! \sum_{j \in \ccalN_i}\! \ccalF_{\ell, m}\big( [\bbX_{\ell\!-\!1}]_i, [\bbX_{\ell\!-\!1}]_j, [\bbE]_{ij} \big),~i=1,...,n,
\end{align}
where $\ccalN_i$ is the neighbor set. These intermediate features are processed by $\ccalF_{\ell,u}$ to generate the output signal as
\begin{align}\label{eq:GNNLayer}
	[\bbX_{\ell}]_i \!=\! \ccalF_{\ell, u}\Big( [\bbX_{\ell-1}]_i, [\bbU_{\ell}]_i \Big),~i=1,...,n.
\end{align}
The GNN inputs are the node states $\bbX$ and the GNN output is the last layer output $\bbX_L$. We define the GNN as a nonlinear mapping $\bbPhi(\bbX, \bbE, \bbtheta_a)$ where the architecture parameters $\bbtheta_a$ collect all function parameters of $\{\ccalF_{\ell,m}, \ccalF_{\ell,u}\}_{\ell=1}^L$.  

We parameterize the navigation policy with GNNs because they allow for a \textit{decentralized implementation}. Specifically, the message aggregation in \eqref{eq:intermediateFeature} requires only signal values of the neighboring nodes and the latter can be obtained by communications, while the feature update in \eqref{eq:GNNLayer} is a local operation and does not affect the decentralized nature (e.g., see \cite{li2020graph, tolstaya2020learning, gao2022wide}); hence, the output signal can be computed at each node locally with neighborhood information only. 

\subsection{DNN-Based Generative Model}

The generative model generates the obstacle layout based on the navigation task, i.e., the initial and goal positions, before agents start to move. In this context, we can deploy \emph{centralized} information processing architectures to parameterize the generative model. DNNs are well-suited candidates that have achieved success in a wide array of applications \cite{liu2017survey, sanchez2018real, gao2019optimal}. Specifically, DNNs comprise multiple layers, where each layer consists of a linear operation and a nonlinearity. At layer $\ell$, the input signal $\bbX_{\ell-1} \in \mathbb{R}^{F_{\ell-1} \times F}$ is processed by a linear operator $\bbGamma_\ell \!\in\! \mathbb{R}^{F_\ell \!\times\! F_{\ell-1}}$ and passed through a nonlinearity $\sigma(\cdot)$ to generate $\bbX_{\ell} \in \mathbb{R}^{F_{\ell} \times F}$, where $F_\ell$ is the number of units at layer $\ell$ and $F$ is the number of features at each unit, i.e.,
\begin{align}
	&\bbX_0 \!=\! \bbX,~\bbX_\ell \!=\! \sigma\big(\bbGamma_\ell \bbX_{\ell\!-\!1}\big)~\for~\ell=1,...,L,
\end{align}  
where the initial and goal positions $\bbX = [\bbS, \bbD]$ are the DNN inputs, the last layer output $\bbX_L= \bbPhi_o(\bbX, \bbtheta_o)$ is the DNN output, and the architecture parameters $\bbtheta_o$ are the weights of all $\{\bbGamma_\ell\}_{\ell=1}^L$. We remark that DNNs offer a 
near-universal parameterization, i.e., they can approximate any function to any desired degree of accuracy given sufficiently large layers and features, providing a strong representational capacity \cite{hornik1989multilayer}. 

\subsection{Policy Distribution}\label{subsec:policyDistribution}

We consider the GNN (or DNN) output $\bbX_L$ as the parameters of the policy (or generative) distribution and sample the agent actions $\bbU_a$ (or obstacle layout $\ccalO$) from the distribution instead of directly computing them. This not only allows to train the parameters $(\bbtheta_a, \bbtheta_o)$ with policy gradient in a model-free manner [cf. \eqref{eq:policyGradient}-\eqref{eq:optimizationParametersUpdate}], but also makes the generated actions satisfy the allowable space, i.e., $\bbU_a \in \ccalU_a$ and $\ccalO \in \ccalP_o$ in \eqref{eq:agentEnvironmentCoProblem}. Specifically, these constraints restrict the space of feasible solutions, which are usually difficult to satisfy and require specific post-processing techniques. We overcome this issue by selecting different policy / generative distributions to satisfy different constraints -- see details in the following examples.

\noindent \textbf{Truncated Gaussian distribution.} If the action is within a bounded interval $[0, C]$, we select the distribution as a truncated Gaussian distribution. Specifically, it is a Gaussian distribution with mean $\mu$ and variance $\sigma^2$, and lies in the interval $[0, C]$ with the probability density function 
\begin{align}\label{eq:truncatedGaussian}
	f(x, \mu, \sigma | 0, C) = \frac{1}{\sigma}\frac{\Psi\big(\frac{x - \mu}{\sigma}\big)}{\bbPsi\big(\frac{C-\mu}{\sigma}\big) - \bbPsi\big(\frac{-\mu}{\sigma}\big)},
\end{align}
where $\Psi(x)$ is the probability density function and $\bbPsi(x)$ is the cumulative distribution function of the standard Gaussian distribution. The distribution parameters $\mu$ and $\sigma$ are determined by the GNN (or DNN) output $\bbX_L$.

\noindent \textbf{Categorical distribution.} If the action is to select one value from a set of potential values, e.g., the agent takes one of $K$ discrete actions and the obstacle either exists or not, we select the distribution as a categorical distribution. Specifically, it takes one of $K$ categories and the probability of each one is 
\begin{align}\label{eq:category}
	f(x = k) = p_k~\text{s.t.}~\sum_{k=1}^K p_k = 1.
\end{align}
The distribution parameters $\{p_k\}_{k=1}^K$ are determined by the GNN (or DNN) output $\bbX_L$. For comprehensive restrictions including both \eqref{eq:truncatedGaussian} and \eqref{eq:category}, we select a combination of two distributions. Fig. \ref{fig:framework} illustrates the aforementioned framework. 

\begin{figure*}%
	\centering
	\captionsetup[subfigure]{justification=centering}
	\begin{subfigure}{0.425\columnwidth}
		\includegraphics[width=1\linewidth,height = 0.75\linewidth]{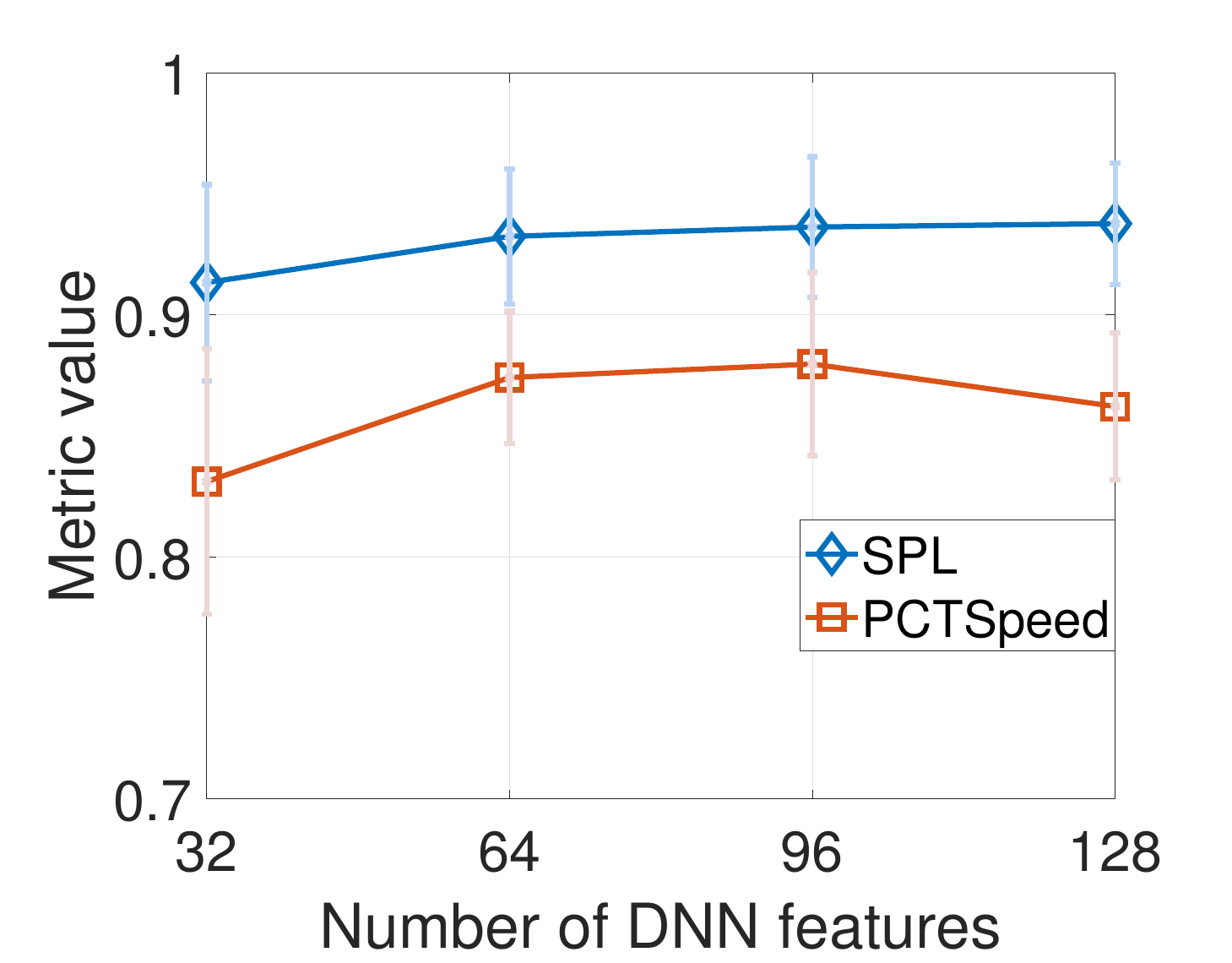}%
		\caption{}%
		\label{subfig:1-1}%
	\end{subfigure}\hfill\hfill%
	\begin{subfigure}{0.425\columnwidth}
		\includegraphics[width=1\linewidth, height = 0.75\linewidth]{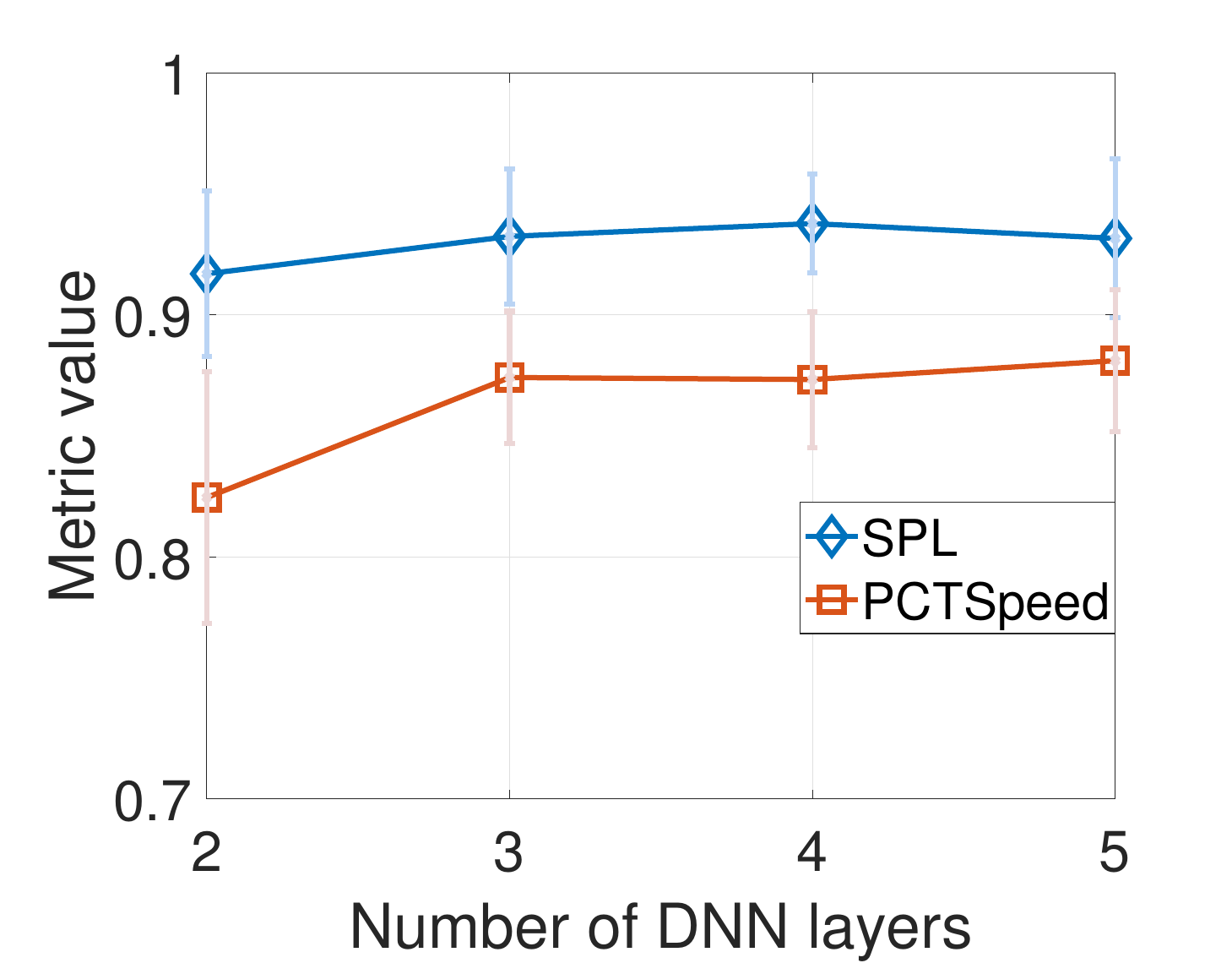}%
		\caption{}%
		\label{subfig:1-2}
	\end{subfigure}\hfill\hfill%
	\begin{subfigure}{0.425\columnwidth}
		\includegraphics[width=1\linewidth,height = 0.75\linewidth]{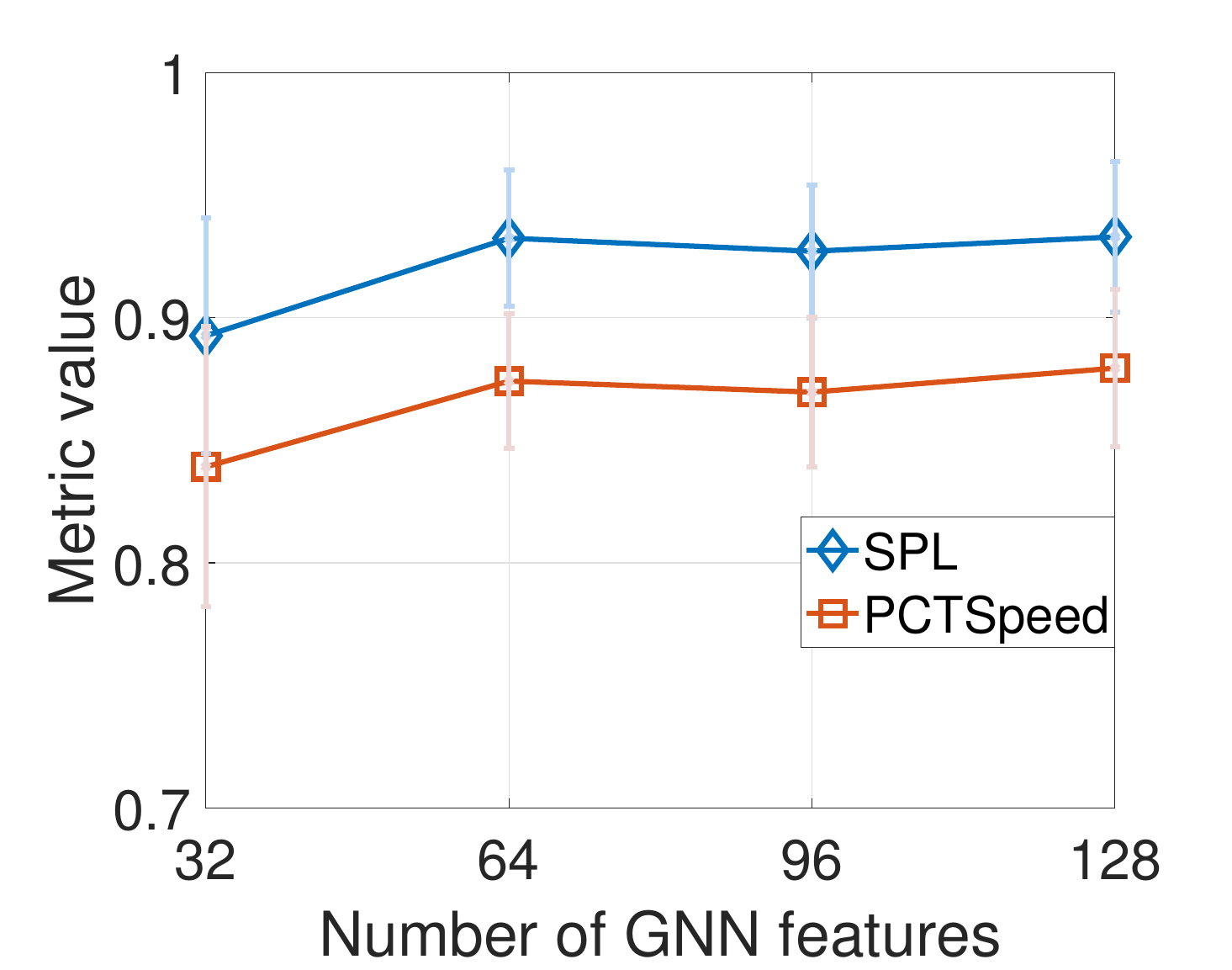}%
		\caption{}%
		\label{subfig:1-3}%
	\end{subfigure}\hfill\hfill%
	\begin{subfigure}{0.425\columnwidth}
		\includegraphics[width=1\linewidth,height = 0.75\linewidth]{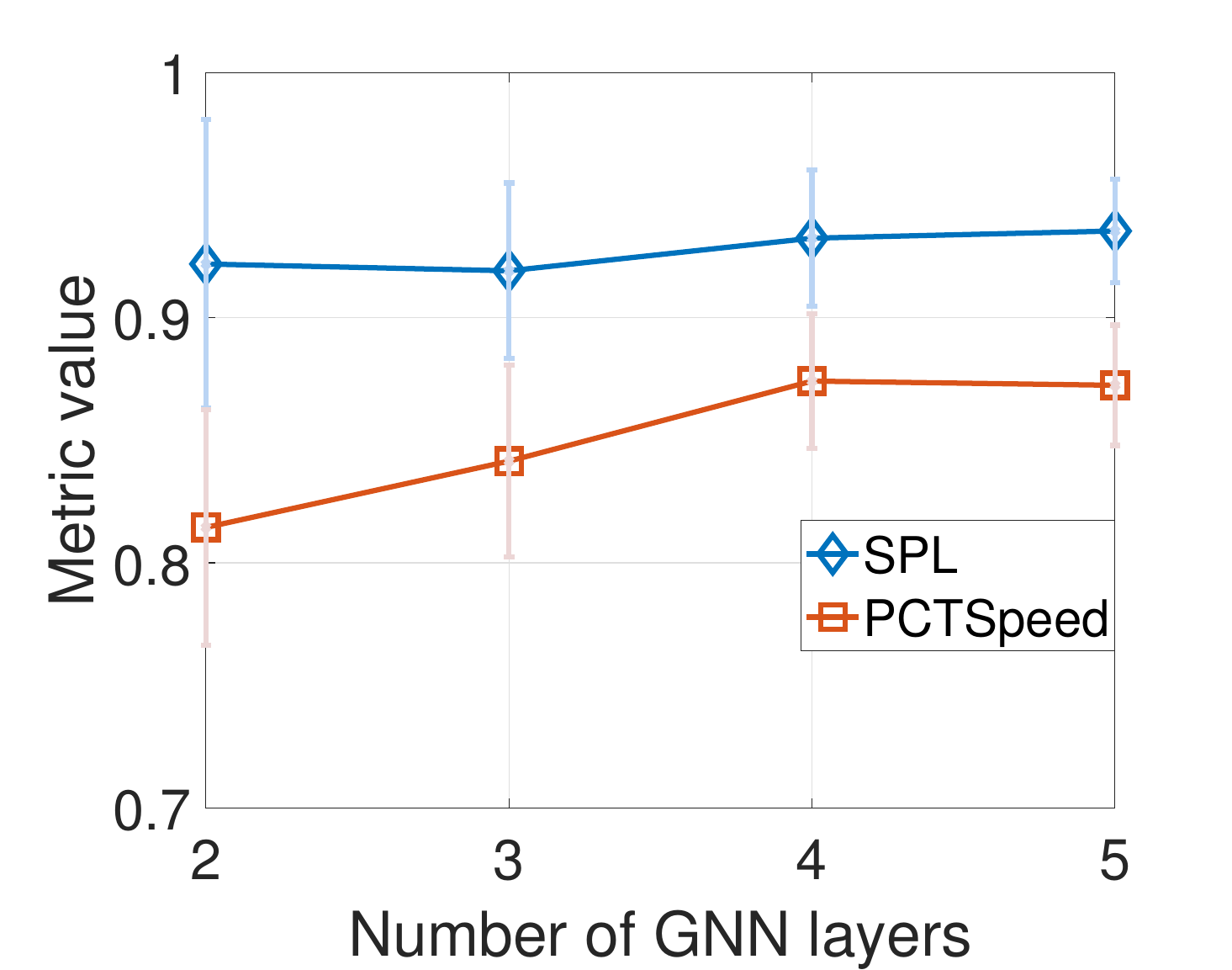}%
		\caption{}%
		\label{subfig:1-4}%
	\end{subfigure}
	\begin{subfigure}{0.425\columnwidth}
		\includegraphics[width=1\linewidth,height = 0.75\linewidth]{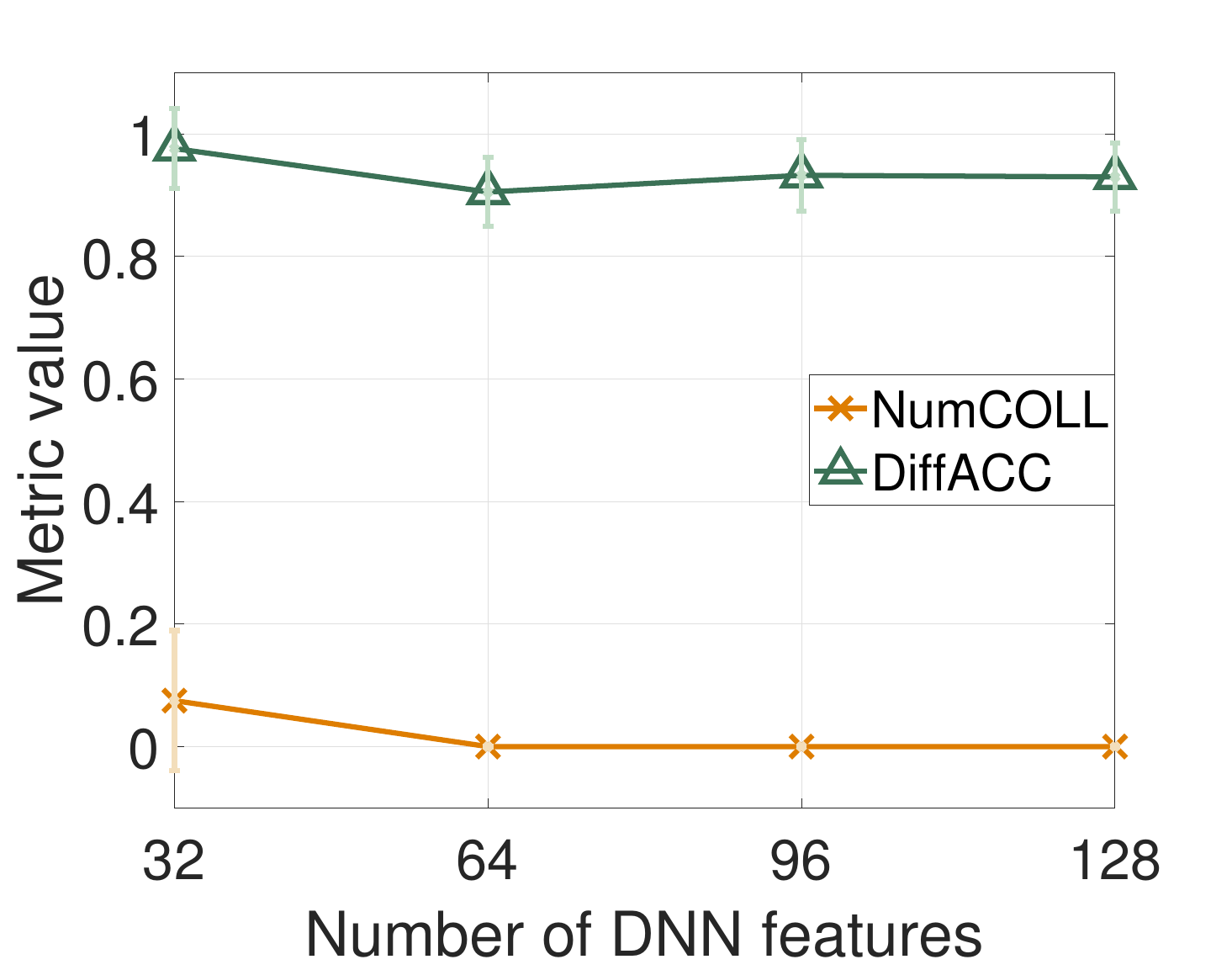}%
		\caption{}%
		\label{subfig:1-5}%
	\end{subfigure}\hfill\hfill%
	\begin{subfigure}{0.425\columnwidth}
		\includegraphics[width=1\linewidth, height = 0.75\linewidth]{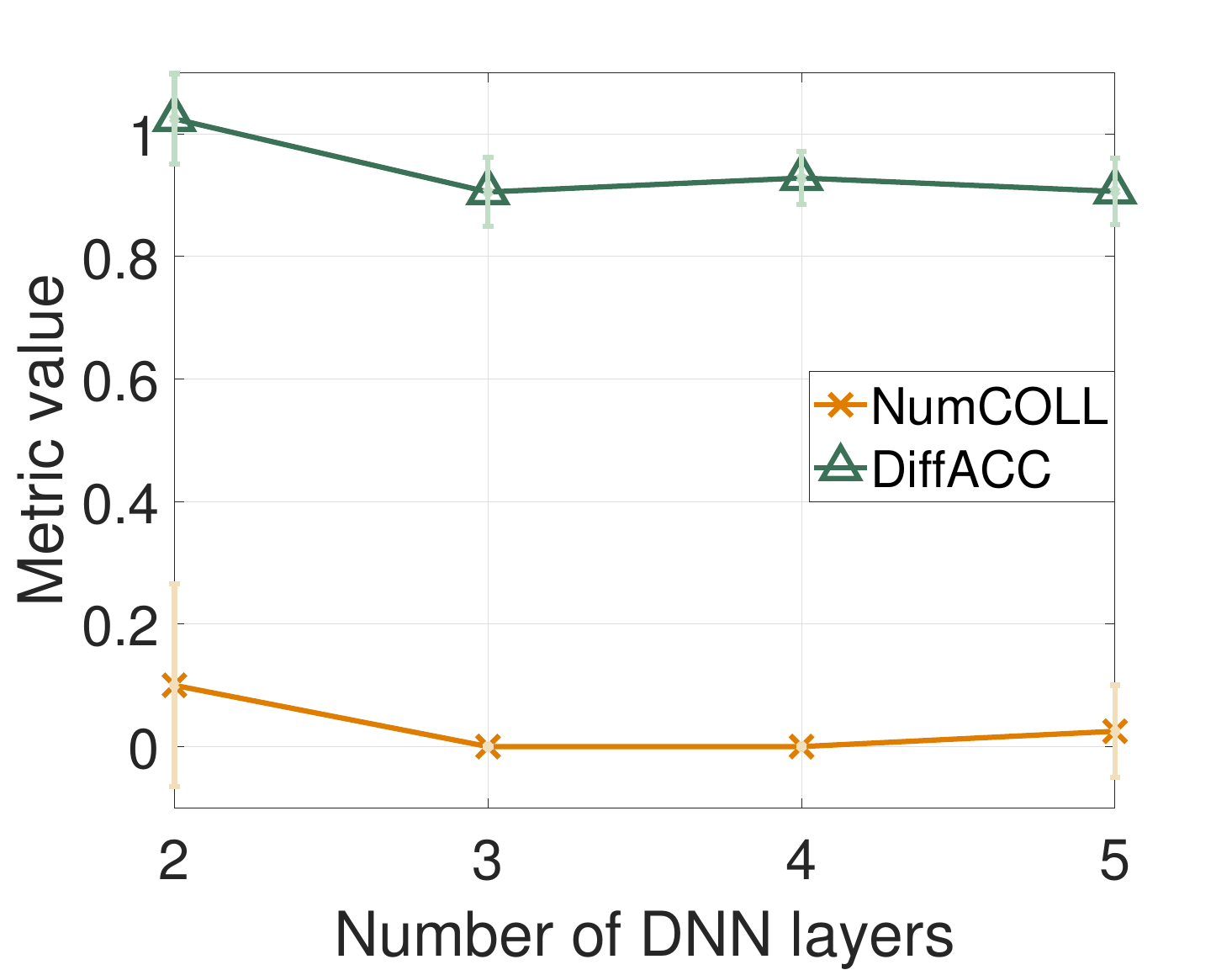}%
		\caption{}%
		\label{subfig:1-6}
	\end{subfigure}\hfill\hfill%
	\begin{subfigure}{0.425\columnwidth}
		\includegraphics[width=1\linewidth,height = 0.75\linewidth]{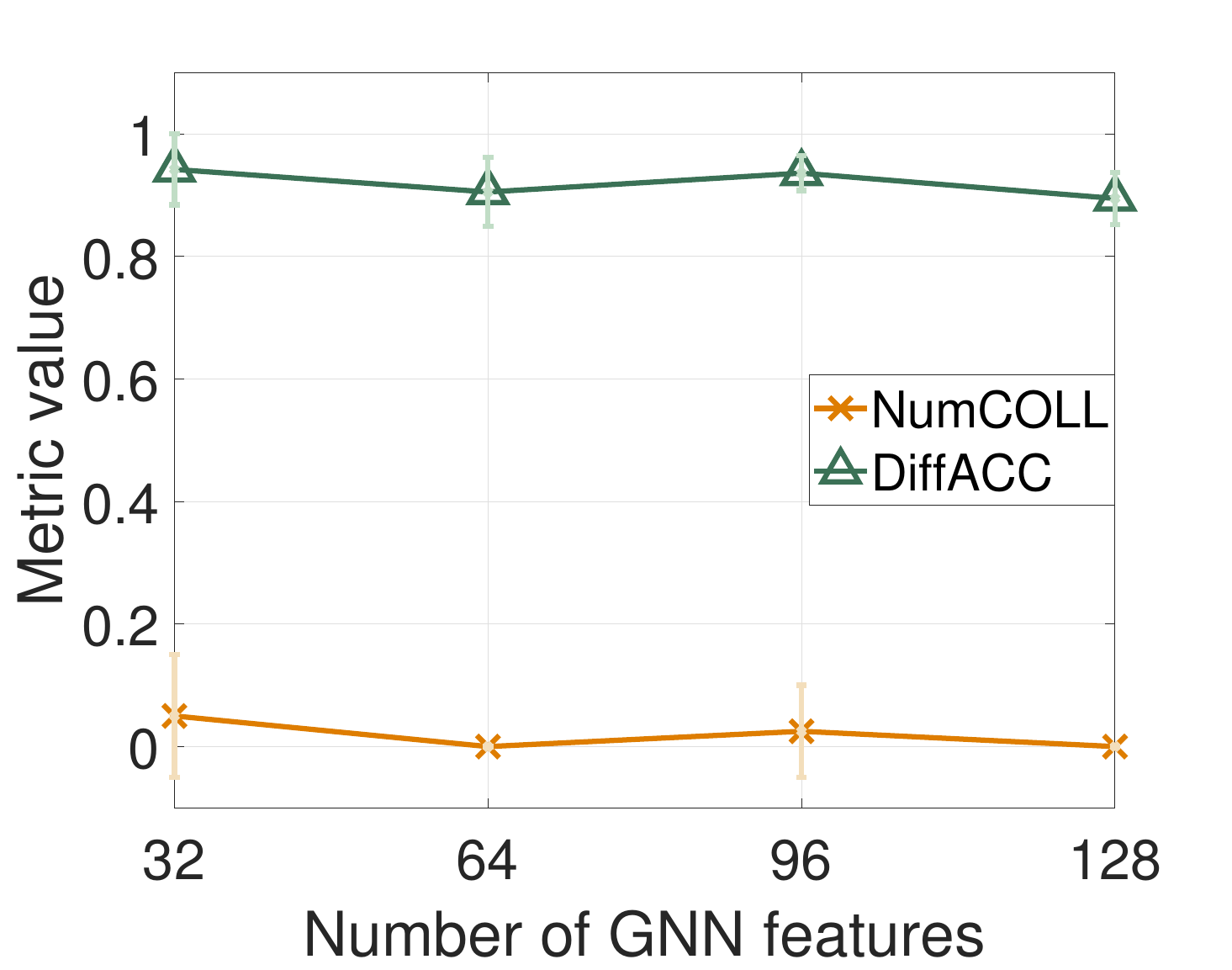}%
		\caption{}%
		\label{subfig:1-7}%
	\end{subfigure}\hfill\hfill%
	\begin{subfigure}{0.425\columnwidth}
		\includegraphics[width=1\linewidth,height = 0.75\linewidth]{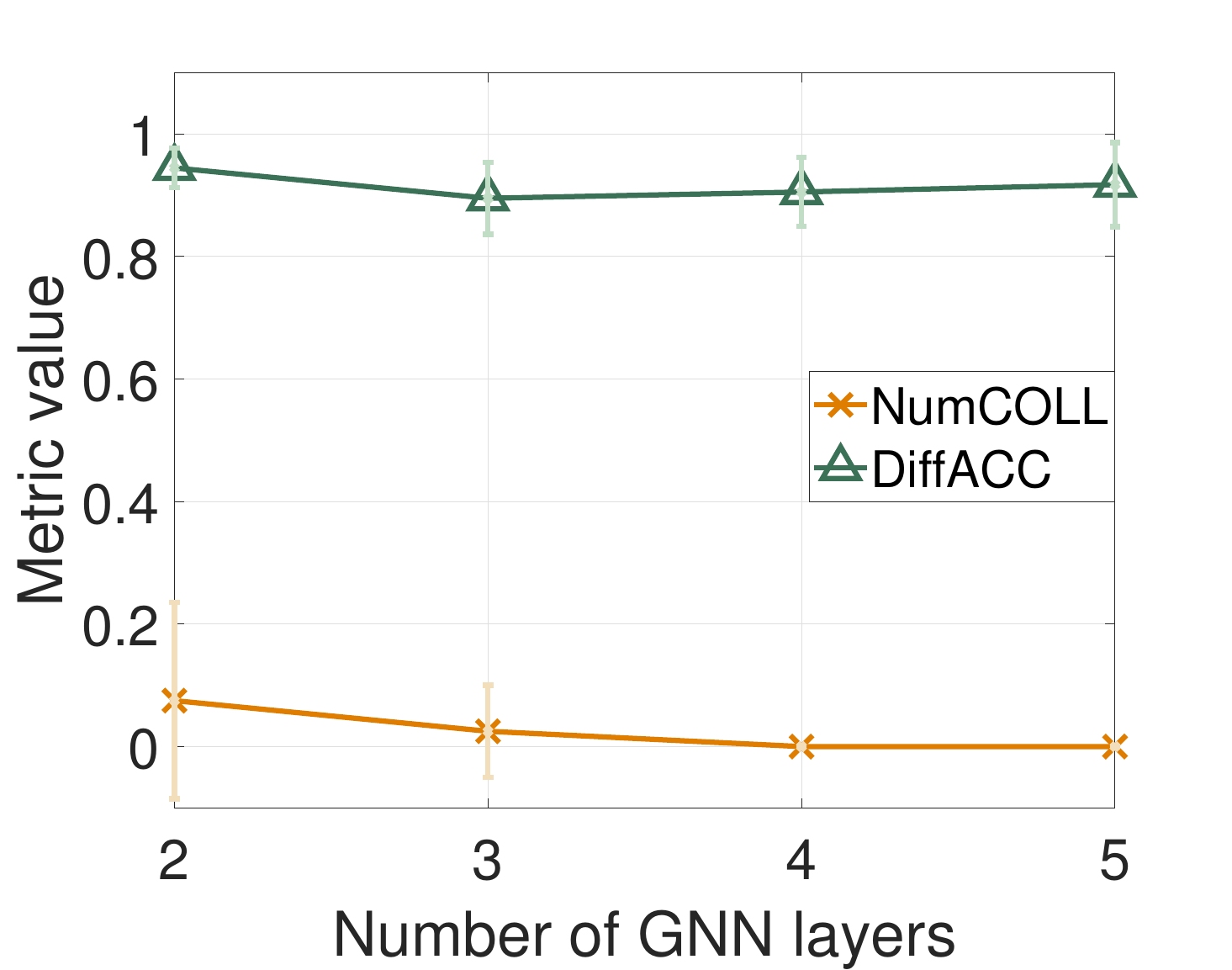}%
		\caption{}%
		\label{subfig:1-8}%
	\end{subfigure}
	\caption{Performance of agent-environment coordinated optimization with different layers and features of DNNs for multi-agent navigation policies and GNNs for environment generative models. (a-d) SPL and PCTSpeed with different features and layers of DNNs and GNNs, where a higher value represents a better performance, i.e., higher success rate, path efficiency and speed. (e-h) NumCOLL and DiffACC with different features and layers of DNNs and GNNs, where a lower value represents a better performance, i.e., more safety and comfort.}\label{fig:hyperparameters}\vspace{-6mm}
\end{figure*} 

The agent-environment co-optimization can be applied to more agents and obstacles because GNNs allow for a decentralized implementation that is scalable to larger multi-agent systems and DNNs have a strong expressive power to handle larger environment settings. As the first stride in this research direction, our work aims to formalize the problem setting and develop a general solution framework, to explore the relationship between agents and environment and to lay a foundation to promote extensions. For future exploration, we can consider simulators that allow to leverage GPUs (e.g., NVIDIA Isaac Sim) to simulate a growing large number of robots and obstacles with physical properties (e.g., inertia, friction, wheel torque, etc.), and evaluate the agent-environment co-optimization for larger-scale problems.

\section{Performance Metrics}\label{appendix:metrics}

This paper considers four metrics, i.e., SPL, PCTSpeed, NumCOLL and DiffACC. These metrics include the information about agents' success, path efficiencies, speed, safety and comfort, which provide a comprehensive performance evaluation to show the effectiveness of the proposed method. 

\textbf{SPL} is the gold standard to measure the performance of robot navigation in the literature \cite{anderson2018evaluation, gervet2023navigating, wang2019reinforced} defined as 
\begin{align}
	\text{SPL} = \frac{1}{n} \sum_{i=1}^n s_i \frac{\ell_i}{\max\{p_i, \ell_i\}},
\end{align}
where $s_i$ is a binary indicator for the success of agent $A_i$, $p_i$ is the traveled distance and $\ell_i$ is the shortest distance. SPL leverages $\{s_i\}_{i=1}^n$ to represent whether navigation tasks of $\{A_i\}_{i=1}^n$ are successful, and the distance ratios $\big\{\ell_i / \max\{p_i, \ell_i\}\big\}_{i=1}^n$ to represent the path efficiencies of successful agents. It is ready to see that: (1) SPL is proportional to the success rate, i.e., the higher the success rate, the larger the SPL; (2) Given the same success rate, SPL is proportional to the path efficiency, i.e., the higher the path efficiency, the larger the SPL. Therefore, it is a stringent measure that can be used as a primary quantifier in comparing the navigation performance.
 
\textbf{PCTSpeed}, \textbf{NumCOLL} and \textbf{DiffACC} are defined as
\begin{align}
	\text{PCTSpeed} &= \frac{1}{nT}\sum_{i=1}^n \sum_{t=1}^T \frac{\|\bbv_i^{(t)}\|}{\|\bbv_{\max}\|},\\
	\text{NumCOLL} \!=\!\!\! \sum_{i=1}^n\! \frac{N_i}{n}&,~\!\text{DiffACC} \!=\!\!\! \sum_{i=1}^n\! \sum_{t=1}^T\! \frac{\|\bba_i^{(t)} \!-\! \bba_i^{(t-1)}\|}{nT}\!,
\end{align}
where $\bbv_i^{(t)}$ is the velocity of agent $A_i$ at time step $t$, $\bbv_{\max}$ is the maximal velocity, $N_i$ is the number of collisions, $\bba_i^{(t)}$ is the acceleration of $A_i$, and $\bba_i^{(0)} \!=\! \bb0$ by default. PCTSpeed leverages the speed ratios to represent how fast agents move along trajectories. NumCOLL characterizes how safe agents are w.r.t. each other and obstacles along trajectories, while DiffACC computes the expected change of agents' accelerations, which measures how comfortable agents are along trajectories. 

SPL and PCTSpeed normalize the metric value to $[0, 1]$ with a unified exposition for a joint comparison, where a higher value represents a better performance. NumCOLL and DiffACC are auxiliary metrics, where a lower value represents a better performance (i.e., more safety and comfort).

\section{Additional Experiments}\label{appendix:additional}

\begin{figure*}%
	\centering
	\captionsetup[subfigure]{justification=centering}
	\begin{subfigure}{0.425\columnwidth}
		\includegraphics[width=1\linewidth,height = 1\linewidth]{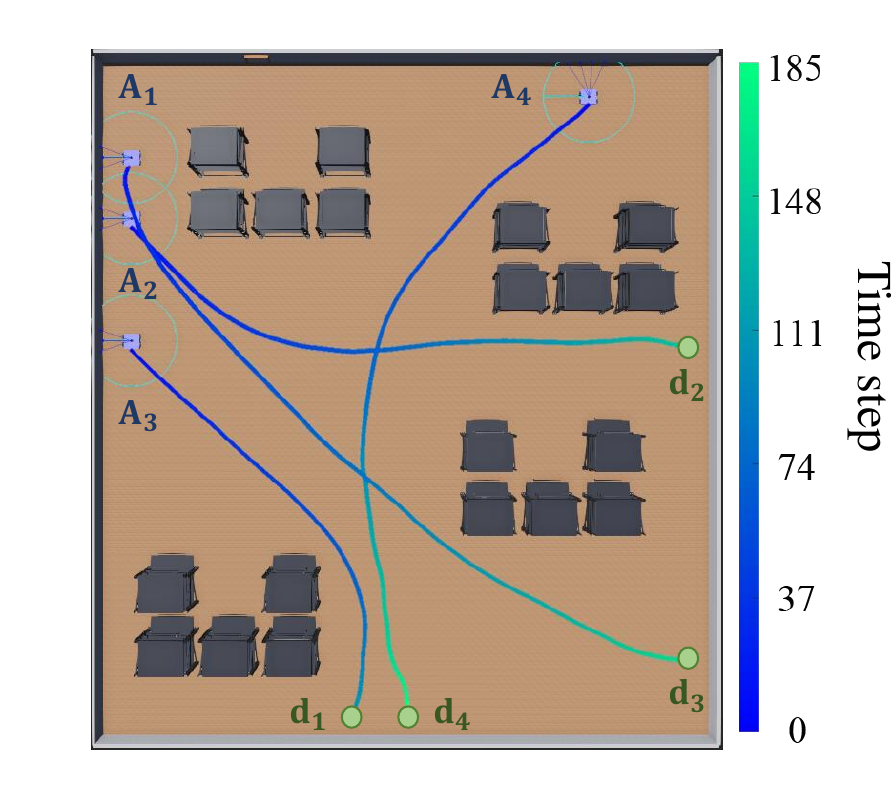}%
		\caption{}%
		\label{subfig:U}%
	\end{subfigure}\hfill\hfill%
	\begin{subfigure}{0.425\columnwidth}
		\includegraphics[width=1\linewidth,height = 1\linewidth]{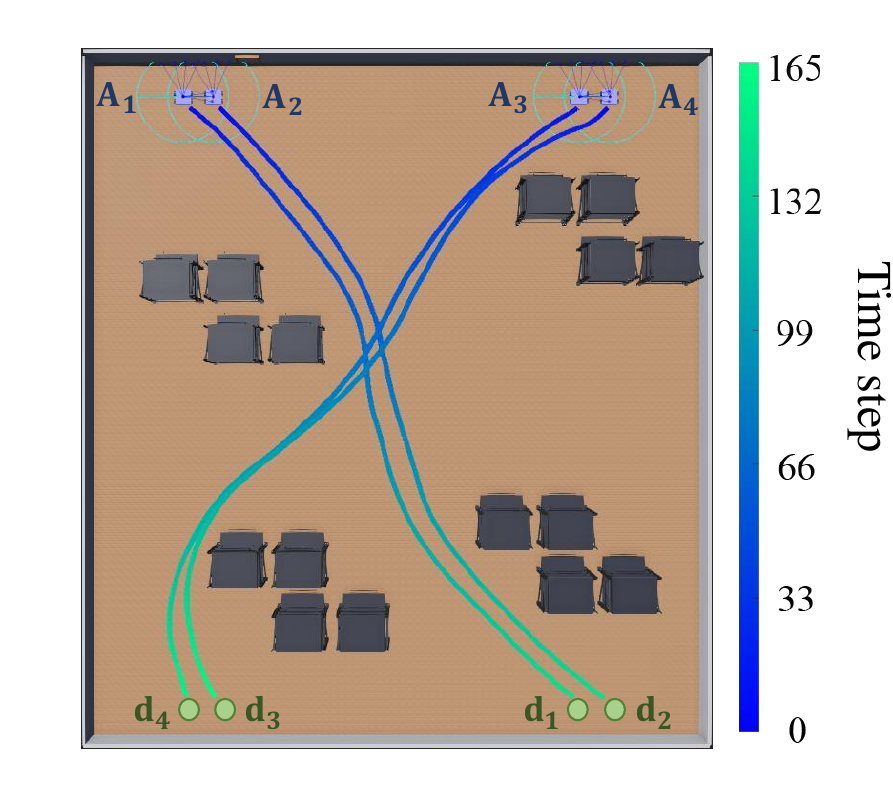}%
		\caption{}%
		\label{subfig:Z}%
	\end{subfigure}\hfill\hfill%
	\begin{subfigure}{0.425\columnwidth}
		\includegraphics[width=1\linewidth, height = 1\linewidth]{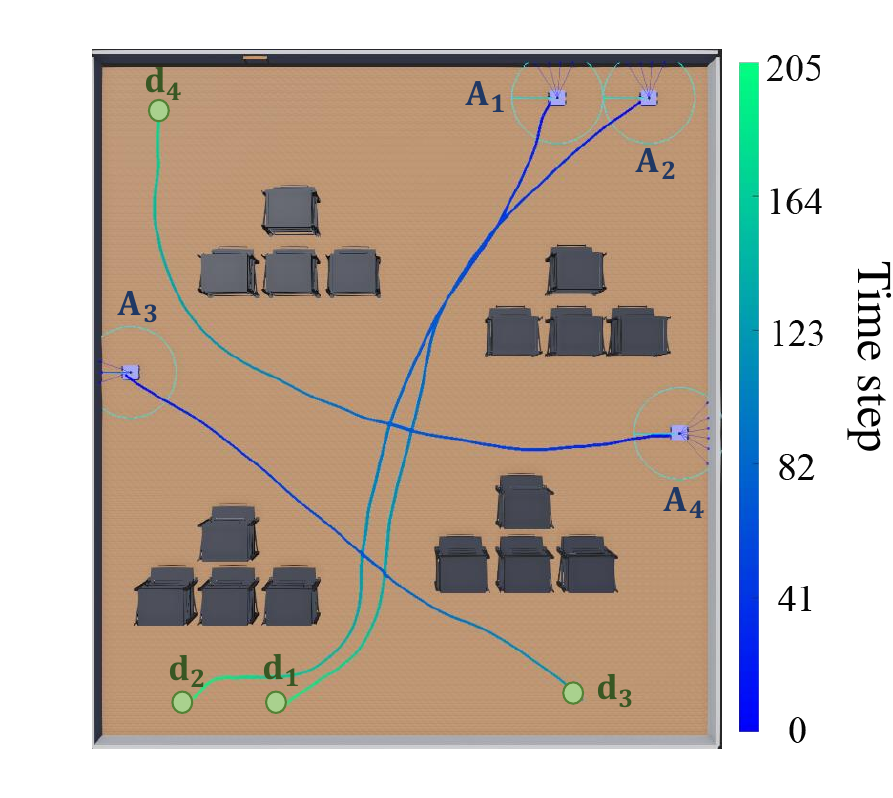}%
		\caption{}%
		\label{subfig:T}
	\end{subfigure}\hfill\hfill%
	\begin{subfigure}{0.425\columnwidth}
		\includegraphics[width=1\linewidth,height = 1\linewidth]{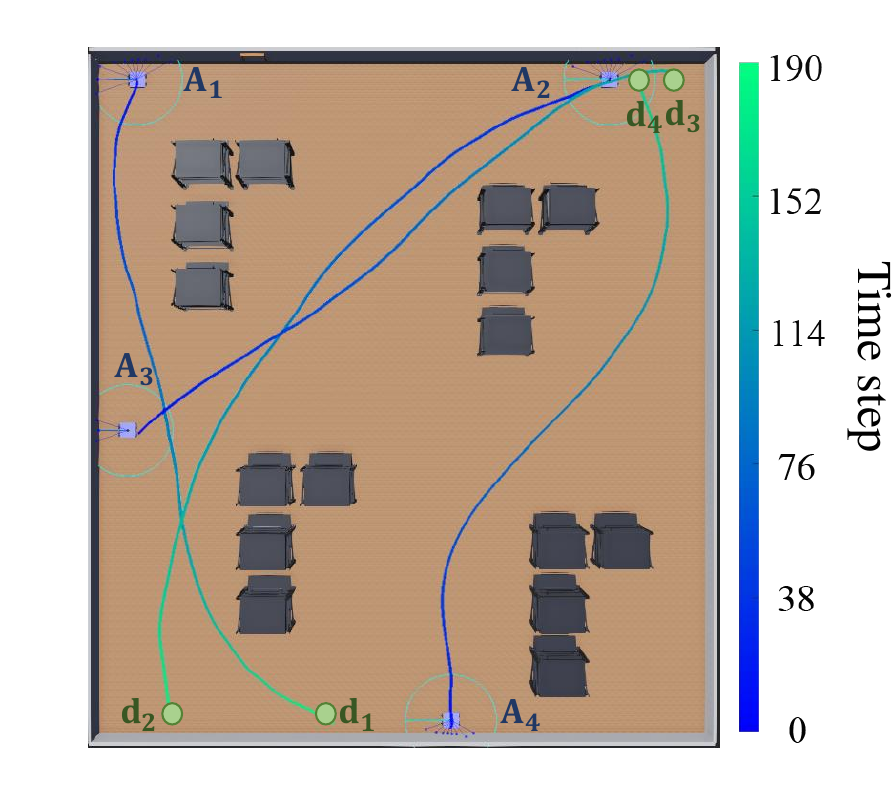}%
		\caption{}%
		\label{subfig:L}%
	\end{subfigure}
	\caption{Examples of agent-environment co-optimization with non-convex irregular obstacles of different shapes. Blue robots $\{A_i\}_{i=1}^4$ represent agents at initial positions and green circles $\{\bbd_i\}_{i=1}^4$ represent goal positions. Colored lines from blue to green are agent trajectories and the color bar represents the time scale. (a) U-shaped obstacles. (b) Z-shaped obstacles. (c) $\perp$-shaped obstacles. (d) $\Gamma$-shaped obstacles.}\label{fig:irregular}\vspace{-6mm}
\end{figure*}

\begin{figure}%
\centering
\includegraphics[width=0.425\linewidth,height = 0.375\linewidth]{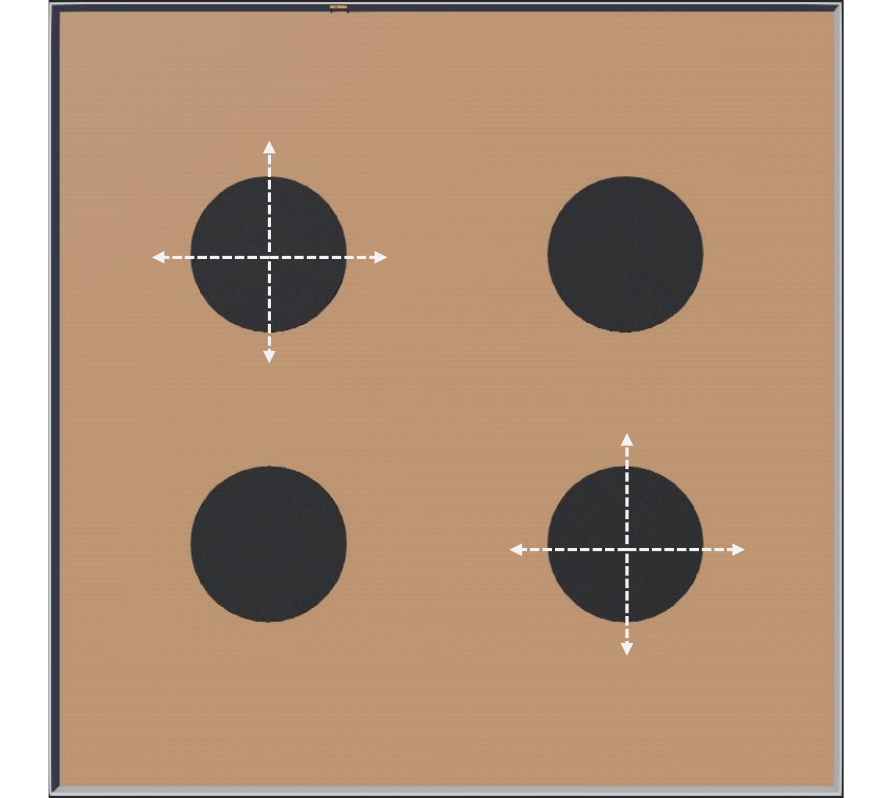}%
\caption{Environment for evaluating the effect of obstacle / agent size on the agent-environment co-optimization. The obstacle positions are re-configurable along x- and y-axes in a neighborhood of $5 \times 5$.}\label{fig:sizeSetting}\vspace{-6mm}
\end{figure}

We conduct additional experiments to further validate our method. First, we evaluate it with different hyperparameters, i.e., features and layers of DNNs and GNNs for the environment generative model and multi-agent navigation policies. Second, we consider irregular obstacles with U, Z, $\perp$ and $\Gamma$ shapes, to show the general applicability of our method. Third, we characterize the effect of obstacle / agent size on the performance of our method. Lastly, we investigate the role of inter-agent collision penalty in guiding agent de-confliction, in addition to the obstacle guidance stated in Sec. \ref{subsec:blockingorGuidance}.

\subsection{Hyperparameter Sensitivity} 

In our experiments of Sec. \ref{sec:experiments}, we consider a 3-layered DNN and a 1-layered GNN, where message aggregation and feature update functions are $4$-layered MLPs with $64$ units and ReLU nonlinearity per layer. This is a common selection, which has already exhibited superior performance. To further study hyperparameter effects, we evaluate our approach with different layers and features of DNNs and GNNs in Fig. \ref{fig:hyperparameters}\footnote{We maintain a 1-layered GNN for an efficient decentralized implementation with only $1$-hop communication, and evaluate different layers and features of message aggregation and feature update functions.}. 

We see that SPL / PCTSpeed increases and NumCOLL / DiffACC decreases with the number of layers or features for both DNNs and GNNs. This is because more layers or features yield a stronger representational capacity for parameterization and thus, an improved performance for coordinated optimization. When the number of features or layers is large, the representational capacity reaches a saturated level and the performance fluctuates around the optimal value. We remark that the proposed method achieves a good performance even with a small number of layers (e.g., $2$ layers) or features (e.g., $32$ features). This indicates that the proposed method is not sensitive to the selection of neural network hyperparameters.

\subsection{Obstacle Irregularity}

We consider rectangular and circular obstacles in Sec. \ref{sec:experiments}, following the standard setting of multi-agent navigation in the literature \cite{ismail2018survey, hildreth2019coordinating, yu2023learning}. Our method can be extended to irregular obstacles of other shapes. Given any design choices of irregular obstacles, we can consider these re-configurable parameters as outputs of the environment generative model and carry out the coordinated optimization in the same manner. 

To validate this extension, we evaluate the proposed method for non-convex obstacles with different irregular shapes, which include U-shaped, Z-shaped, $\perp$-shaped and $\Gamma$-shaped obstacles. Fig. \ref{fig:irregular} shows examples of the agent-environment co-optimization in these scenarios. We see that the proposed method achieves good performance for non-convex irregular obstacles as well as regular ones. The environment generative model produces compatible obstacle layouts and the multi-agent navigation policy generates smooth agent trajectories, yielding superior performance in a joint manner.

\subsection{Agent and Obstacle Size}

From Sec. \ref{sec:experiments}, we see that more obstacles result in more complex navigation scenarios, require more compatible environments for multi-agent navigation, and emphasize the performance improvement introduced by agent-environment co-optimization (from Fig. \ref{subfig:Performance4Obstacles} to Fig. \ref{subfig:Performance16Obstacles}). Moreover, more agents lead to more cluttered navigation scenarios, require more guidance from obstacles for de-confliction, and increase the performance improvement obtained by agent-environment co-optimization (from Fig. \ref{subfig:performance8} to Fig. \ref{subfig:performance16}). This subsection studies the influence of obstacle / agent size on the co-optimization performance. Specifically, we consider the environment of size $12 \times 12$, where obstacle positions can be re-configured along x- and y-axes in a neighborhood of size $5 \times 5$ -- see Fig. \ref{fig:sizeSetting}.

\begin{figure*}%
	\centering
\captionsetup[subfigure]{justification=centering}
	\begin{subfigure}{0.425\columnwidth}
		\includegraphics[width=1\linewidth,height = 0.85\linewidth]{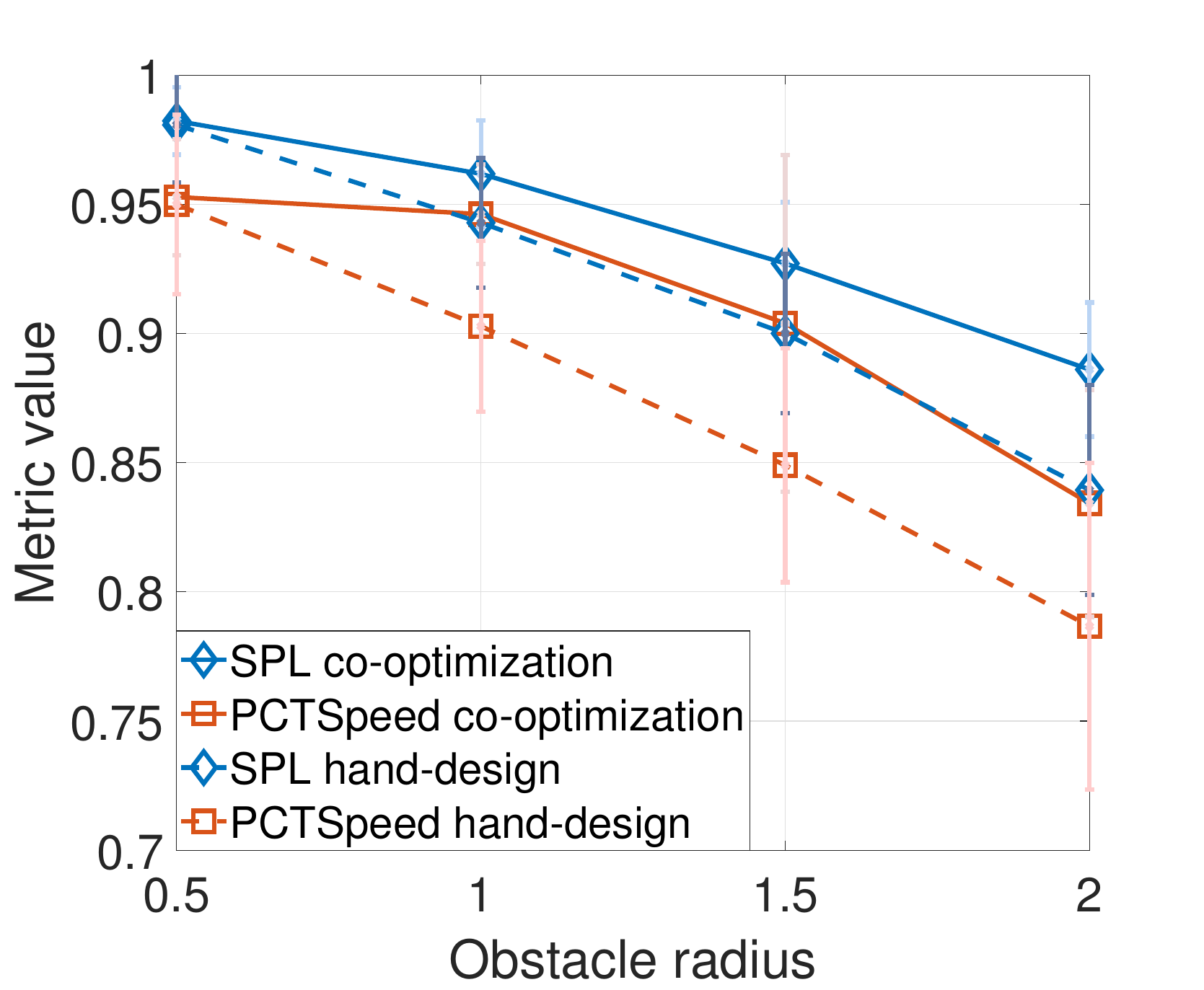}%
		\caption{}%
		\label{subfig:2-1}%
	\end{subfigure}\hfill\hfill%
	\begin{subfigure}{0.425\columnwidth}
		\includegraphics[width=1\linewidth, height = 0.85\linewidth]{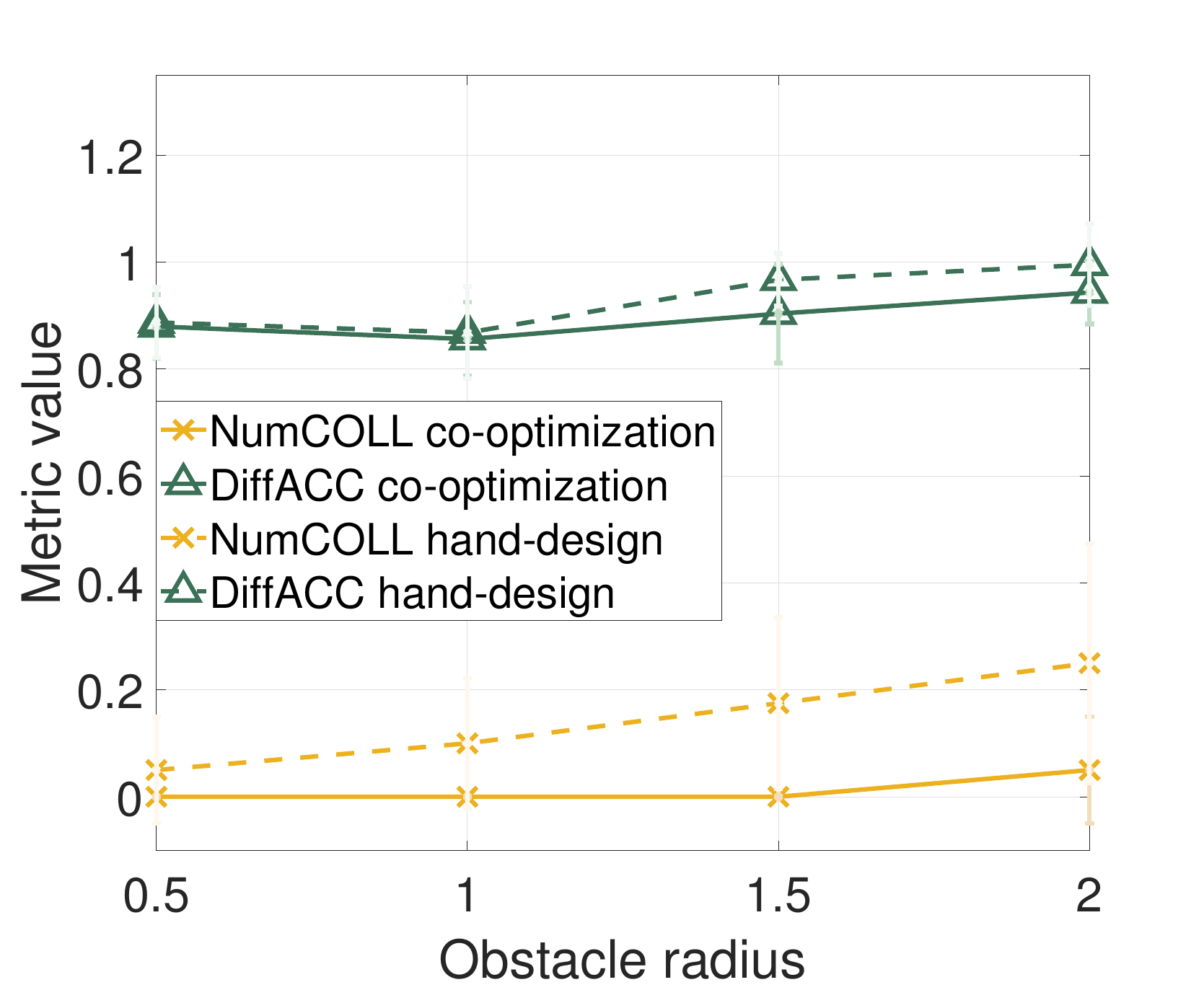}%
		\caption{}%
		\label{subfig:2-2}
	\end{subfigure}\hfill\hfill%
	\begin{subfigure}{0.425\columnwidth}
		\includegraphics[width=1\linewidth,height = 0.85\linewidth]{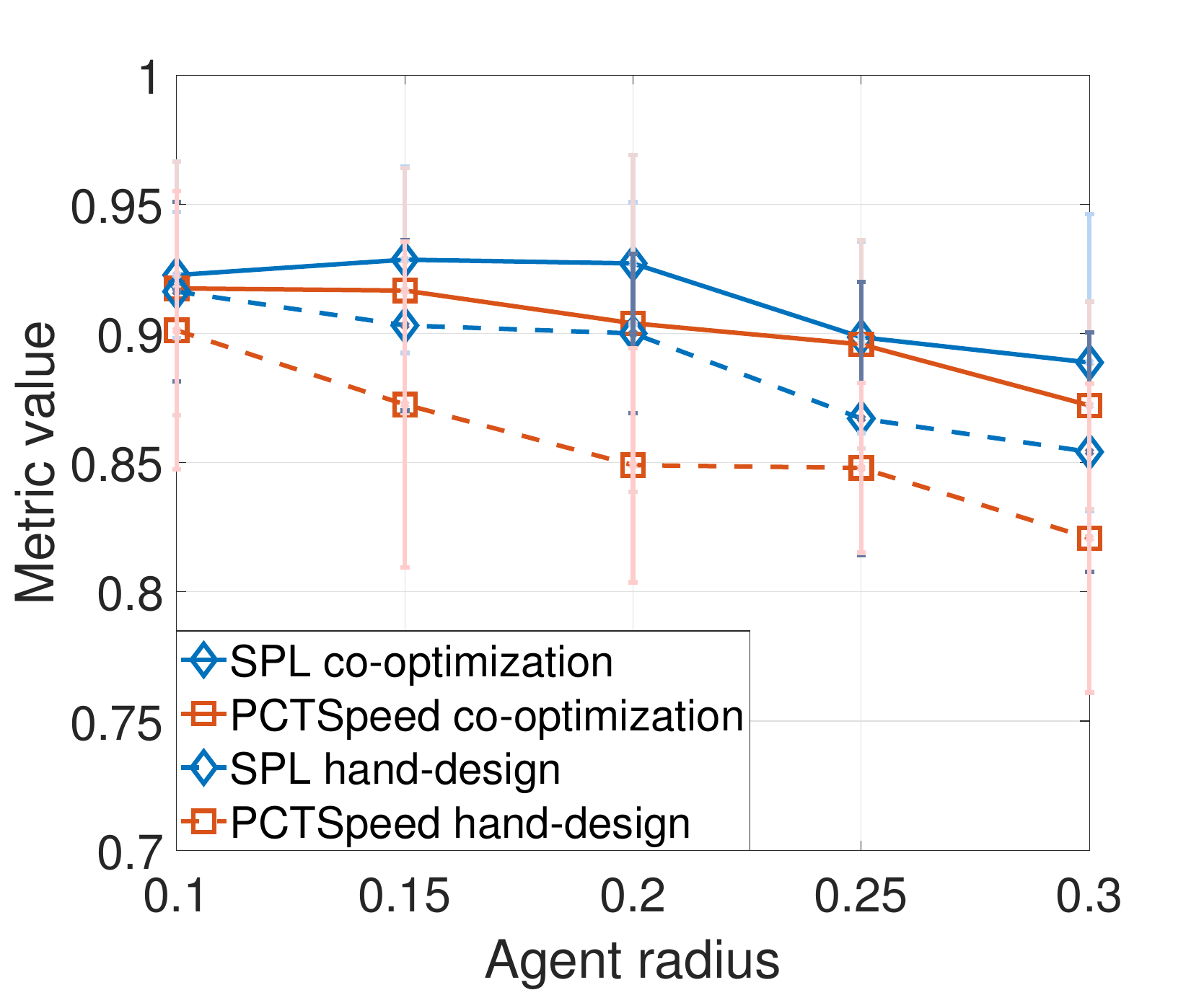}%
		\caption{}%
		\label{subfig:2-3}%
	\end{subfigure}\hfill\hfill%
	\begin{subfigure}{0.425\columnwidth}
		\includegraphics[width=1\linewidth,height = 0.85\linewidth]{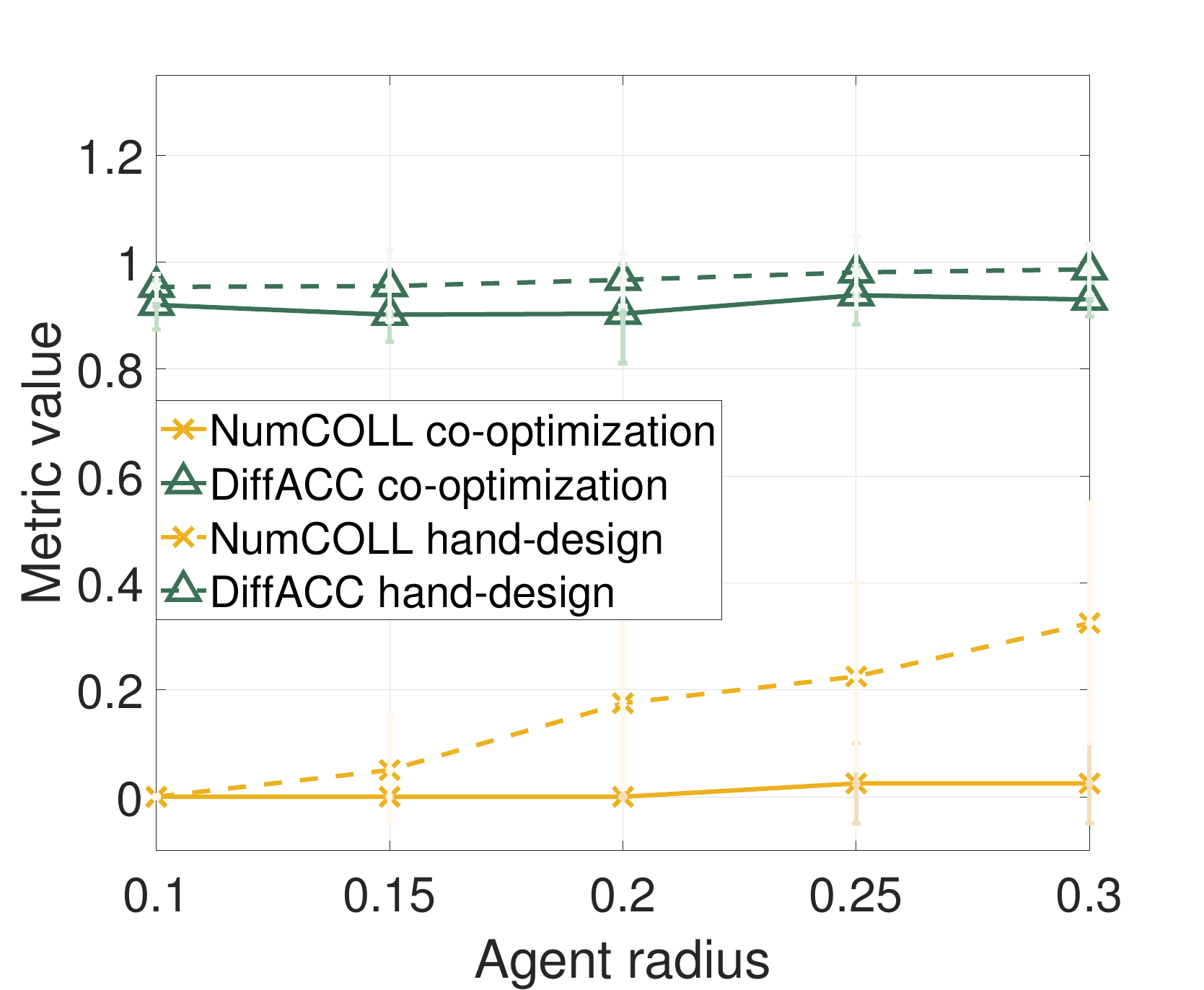}%
		\caption{}%
		\label{subfig:2-4}%
	\end{subfigure}
	\caption{Performance of agent-environment coordinated optimization (solid line) and hand-designed baseline (dashed line) with different obstacle and agent sizes. (a) SPL and PCTSpeed with different obstacle sizes, where a higher value represents a better performance. (b) NumCOLL and DiffACC with different obstacle sizes, where a lower value represents a better performance. (c) SPL and PCTSpeed with different agent sizes. (d) NumCOLL and DiffACC with different agent sizes.}\label{fig:size}\vspace{-6mm}
\end{figure*}

Fig. \ref{fig:size} shows the results with obstacle radius from $0.5$ to $2$ and agent radius from $0.1$ to $0.3$. Our method outperforms the hand-designed baseline consistently in all cases, and the performance improvement introduced by agent-environment co-optimization increases with the obstacle or agent size. This is because a larger obstacle or agent size leads to a more challenging navigation scenario and requires a more compatible environment for agents, highlighting the importance of establishing a symbiotic agent-environment co-existing system. 

\subsection{Inter-Agent Collision Penalty}

\begin{figure}%
	\centering
\captionsetup[subfigure]{justification=centering}
	\begin{subfigure}{0.425\columnwidth}
		\includegraphics[width=1\linewidth,height = 0.75\linewidth]{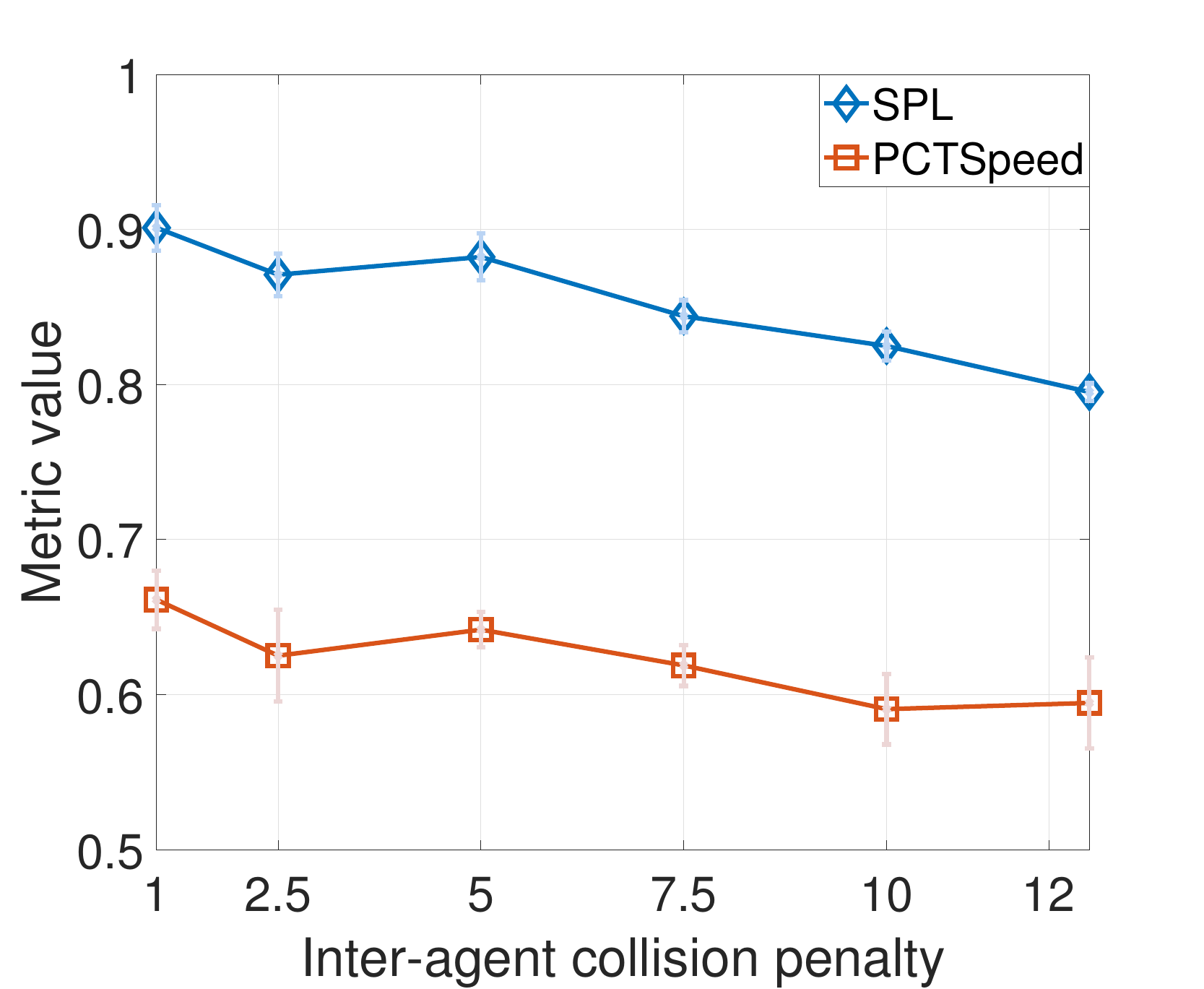}%
		\caption{}%
		\label{subfig:SPL-penalty}%
	\end{subfigure}\hfill\hfill%
	\begin{subfigure}{0.425\columnwidth}
		\includegraphics[width=1\linewidth, height = 0.75\linewidth]{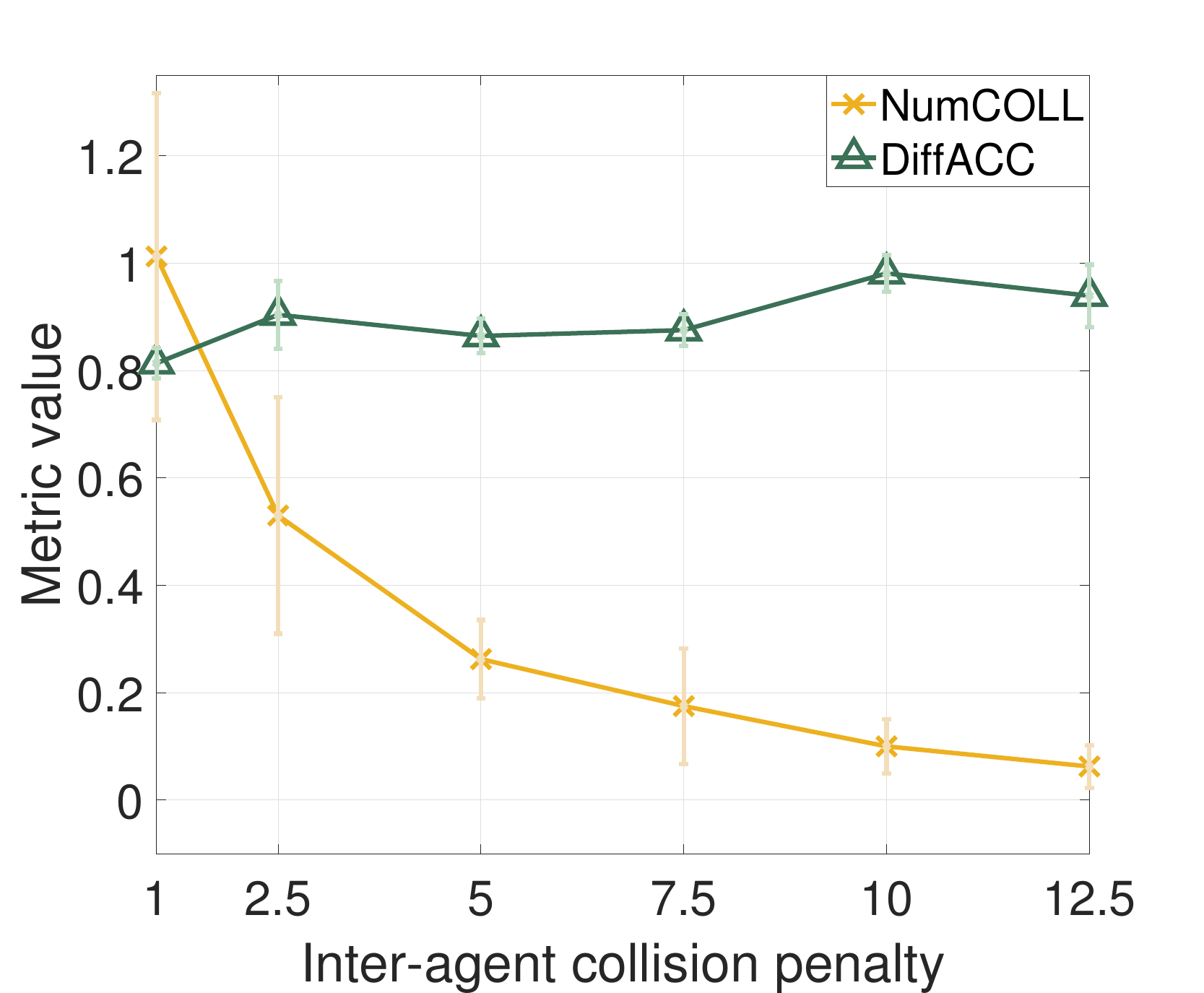}%
		\caption{}%
		\label{subfig:COLL-penalty}
	\end{subfigure}
	\caption{Navigation performance in the circular setting of $16$ agents, with no obstacle and different inter-agent collision penalties. (a) SPL and PCTSpeed. (b) NumCOLL and DiffACC.}\label{fig:penalty}\vspace{-6mm}
\end{figure}

We show in Sec. \ref{subsec:blockingorGuidance} that obstacles can have positive effects that guide agents to de-conflict for safe navigation. On the other hand, each agent can be viewed as a dynamic ``obstacle'' to the other agents and may play a similar guidance role. To investigate this factor, we consider the circular scenario with no obstacle in Sec. \ref{subsec:blockingorGuidance}, and train multi-agent navigation policies with different collision penalties between agents in the range of $[1, 12.5]$. Fig. \ref{fig:penalty} shows the results. 

NumCOLL decreases with the increasing of inter-agent collision penalty. This is expected as a larger penalty leads to more conservative behaviors and the latter facilitates agent de-confliction for safe navigation. However, this results in worse navigation performance, i.e., lower SPL and PCTSpeed, because the agents focus more on safety than goal reaching. Moreover, the co-optimization method consistently outperforms the case of the largest penalty, with a lower NumCOLL and larger SPL, PCTSpeed. These highlight the importance of obstacle guidance for safe multi-agent navigation. 

We conclude that while inter-agent collision is beneficial for de-confliction, it may not be sufficient to guide agents and involving obstacles is helpful, especially in cluttered scenarios. Moreover, there exist scenarios where obstacles have already existed and are not allowed to be removed. For example, in the warehouse setting, there must exist shelves to store packages and we cannot remove them from the space. In these cases, the proposed method can optimize the layout of the existing obstacles to guide agents and improve performance.

\section{Control Dynamics}\label{appendix:dynamics}

In this section, we provide details about control dynamics considered in our experiments. Specifically, when training multi-agent navigation policies, we assume single integrator systems with constrained velocity inputs, i.e., 
\begin{align}
    &\dot{\bbx}(t) = \bbA\bbx + \bbB\bbv(t),~\text{s.t.}~ \bbv(t) \in \ccalV, \label{continousSystem} 
\end{align}
where $\bbx(t)$ is the 2-D state vector, $\bbA$ a 2 by 2 matrix with all zeroes, $\bbB$ a 2 by 2 identity matrix, $\bbv(t)$ the velocity vector, $\ccalV$ the action space. This is a standard setting in the study of multi-agent navigation, and has been widely used in \cite{martinovic2022cooperative, amirkhani2022consensus, sun2024group}. 

When deploying multi-agent policies in real-world experiments (or Webots simulations), we have omnidirectional robots with nonlinear dynamics, and use Model Predictive Control (MPC) and inverse kinematics as the low-level controller to convert the output of navigation policies to the speed of wheels. Specifically, consider the discrete-time linear time-invariant (LTI) system $\bbx[k+1] = \tilde{\bbA}\bbx[k] + \tilde{\bbB}\bbu[k]$, where $\bbx[k]$ is the state vector at time step $k$, $\bbu[k]$ is the control input, $\tilde{\bbA}$ and $\tilde{\bbB}$ are state transition matrix and control input matrix derived from \eqref{continousSystem}. The MPC minimizes a cost function over a finite prediction horizon $M$, subject to system dynamics and control input constraints. In particular, the cost function is \cite{zhu2018new, ghandriz2024trajectory}
\begin{align}
J \!=\! &\sum_{i=0}^{M-1} \big( \bbx[k\!+\!i|k]^T \bbQ \bbx[k\!+\!i|k] \!+\! \bbu[k\!+\!i|k]^T \bbR \bbu[k+i|k] \big)\nonumber \\
&+ \bbx[k+M|k]^T \bbP \bbx[k+M|k],
\end{align}
where $\bbx[k+i|k]$ ($\bbu[k+i|k]$) is the predicted state (control input) at time step $k+i$ given the state (control input) at time step $k$, $\bbQ$ is the state weighting matrix, $\bbR$ is the control weighting matrix, and $\bbP$ is the terminal state weighting matrix. The MPC problem can then be formulated as 
\begin{align}\label{eq:MPCProblem}
    & \quad \quad \quad \quad \quad \quad \quad \quad \min_{\{\bbu[k], \dots, \bbu[k+M-1]\}}~ J \\
    &\text{s.t.}~~~~\bbx[k+i+1|k] = \tilde{\bbA}\bbx[k+i|k] + \tilde{\bbB}\bbu[k+i|k], \nonumber \\
    &~~~~~~ \quad \quad \quad\bbu[k+i|k] \in \ccalU, \quad i = 0, \dots, M-1, \nonumber
\end{align}
where $\ccalU$ is the action space of control input. We consider off-the-shelf solver CVXPY \cite{diamond2016cvxpy} that leverages Primal-Dual Interior-Point optimization method to solve problem \eqref{eq:MPCProblem}. 

Next, we convert the obtained control input $\bbu = [\bbu_\bbx, \bbu_\bby]^\top$ to the wheel speed. Since omnidirectional robots have squared base, let $l$ be the length from its center to the edge. The velocity command of wheels can be computed as \cite{siradjuddin2022general, li2017motion} 
\begin{align}
    \begin{bmatrix}
        \bbv_{w_1} \\
        \bbv_{w_2} \\
        \bbv_{w_3} \\
        \bbv_{w_4}
    \end{bmatrix} = 
    \begin{bmatrix}
        \frac{1}{\sqrt{2}} & \frac{1}{\sqrt{2}} & l \\
        -\frac{1}{\sqrt{2}} & \frac{1}{\sqrt{2}} & l \\
        -\frac{1}{\sqrt{2}} & -\frac{1}{\sqrt{2}} & l \\
        \frac{1}{\sqrt{2}} & -\frac{1}{\sqrt{2}} & l
    \end{bmatrix}
    \begin{bmatrix}
        \bbu_x \\
        \bbu_y \\
        \omega
    \end{bmatrix},
\end{align}
where $\omega = 0$ is the angular velocity in our case. This indicates that our method can handle actual nonlinear dynamics by abstracting the nonlinear dynamical system as a single integrator system to train navigation policies and leveraging the low-level controller to convert policy outputs to robot actions for real-world deployment. Such an abstraction technique has been widely used in the literature of omnidirectional robots and also extended to drone systems \cite{belta2004abstraction, leahy2016persistent, shankar2021freyja}.

Lastly, we remark that the performance of the proposed method can be further improved by establishing high-fidelity physics models for real-world systems and integrating the latter into the coordinated optimization procedure. Specifically, we can characterize the actual dynamical system through system identification methods \cite{martin2015omniad, seegmiller2016high, khan2021review}. In the phase of multi-agent navigation, we can leverage the high-fidelity model to determine the state transition function to train the navigation policy with reinforcement learning. In the phase of environment optimization, we can evaluate the multi-agent policy in the generated environment with the high-fidelity model and update the generative model with the corresponding reward. It accounts for realistic dynamics when updating multi-agent policies and environment generative models, and has potential of further improving the co-optimization performance.

\bibliographystyle{IEEEbib}
\bibliography{myIEEEabrv,reference,BiblioOp,AProrok_bib}

\end{document}